\documentclass{article}

\usepackage[main, final]{neurips_2026}

\usepackage[utf8]{inputenc} % allow utf-8 input
\usepackage[T1]{fontenc}    % use 8-bit T1 fonts
\usepackage{hyperref}       % hyperlinks
\usepackage{url}            % simple URL typesetting
\usepackage{booktabs}       % professional-quality tables
\usepackage{amsfonts}       % blackboard math symbols
\usepackage{nicefrac}       % compact symbols for 1/2, etc.
\usepackage{microtype}      % microtypography
\usepackage{xcolor}         % colors

\newcommand{\rev}[1]{#1}

\newcommand{\codeurl}{https://github.com/mhdfb/sideinfo-search-reconstruction}

\title{Inference-Time Search Using Side Information for Diffusion-Based Image Reconstruction}

\author{%
  Mahdi Farahbakhsh\thanks{Equal Contribution.}%
  \thanks{\texttt{\{mahdi.farahbakhsh,vishnukunde,dileep.kalathil,krn,chmbrlnd\}@tamu.edu}} \quad
  Vishnu Teja Kunde\footnotemark[1] \quad
  Dileep Kalathil \\
  \bfseries Krishna Narayanan \quad
  Jean-Francois Chamberland \\
  \normalfont Department of Electrical and Computer Engineering, Texas A\&M University
}

\usepackage{amsmath}
\usepackage{amssymb}
\usepackage{mathtools}
\usepackage{amsthm}
\usepackage{tikz}
\usepackage{amsmath, amssymb, bbm}
\usepackage[]{enumitem}
\usepackage{wrapfig}

\usepackage{subcaption}
\usepackage[dvipsnames]{xcolor}
\usepackage{hyperref}

\newcommand{\abs}[1]{\vert#1\vert}
\newcommand{\norm}[1]{\lVert#1\rVert}

\newcommand{\bE}{\mathbb{E}}

\newcommand{\bR}{\mathbb{R}}

\newcommand{\cD}{\mathcal{D}}

\newcommand{\cL}{\mathcal{L}}

\newcommand{\cN}{\mathcal{N}}

\newcommand{\cP}{\mathcal{P}}

\newcommand{\vg}{\mathbf{g}}

\newcommand{\vs}{\mathbf{s}}

\newcommand{\vv}{\mathbf{v}}
\newcommand{\vw}{\mathbf{w}}
\newcommand{\vx}{\mathbf{x}}
\newcommand{\vy}{\mathbf{y}}
\newcommand{\vz}{\mathbf{z}}

\newcommand{\mA}{\mathbf{A}}

\newcommand{\mI}{\mathbf{I}}
\newcommand{\mJ}{\mathbf{J}}

\newcommand{\bmu}{\boldsymbol{\mu}}

\newcommand{\btheta}{\boldsymbol{\theta}}

\usepackage[capitalize]{cleveref}

\theoremstyle{plain}
\newtheorem{theorem}{Theorem}
\newtheorem{proposition}[theorem]{Proposition}
\newtheorem{lemma}[theorem]{Lemma}

\theoremstyle{definition}

\theoremstyle{remark}
\newtheorem{remark}[theorem]{Remark}

\usepackage{array}
\usepackage{multirow}
\usepackage{graphicx}
\usepackage{subcaption}
\usepackage{algorithm}
\usepackage{algorithmic}

\begin{document}

\maketitle

\begin{abstract}
  Diffusion models have been used as priors for solving inverse problems. However, existing approaches typically overlook side information that could significantly improve reconstruction quality, especially in severely ill-posed settings. In this work, we propose a novel framework that incorporates side information into existing diffusion-based inverse problem solvers via inference-time search, in a plug-and-play, training-free manner. Through extensive experiments across a range of inverse problems, including inpainting, super-resolution, and several deblurring tasks, and across multiple diffusion-based inverse problem solvers (DPS, DAPS, and MPGD), we show that augmenting each solver with our framework consistently improves the  quality of the reconstructions over the corresponding original method. To demonstrate the generality of our approach, we consider diverse forms of side information, including reference images, textual descriptions, and anatomical MRI scans.
\rev{Code is available at this \href{\codeurl}{repository}.\footnote{\url{\codeurl}}}

\end{abstract}

\vspace{-0.2cm}
\section{Introduction}
\vspace{-0.2cm}

\label{sec:introduction}

Diffusion models \citep{ho2020ddpm, song2021scorebased}  have shown remarkable success in diverse generative tasks like text-to-image synthesis \citep{rombach2022stablediffusion}, protein generation \citep{wu2024protein}, video \citep{ho2022imagen_video}, audio \citep{kong2021diffwave}, and language modeling \citep{sahoo2024maskeddiff}. Beyond generation, these models have shown great promise in \textbf{solving inverse problems}, where the goal is to reconstruct an image from partial or noisy measurements  \citep{chung2023dps, daras2024survey}.
A key distinction is that unlike many generative tasks (e.g., text-to-image synthesis, personalized editing, style transfer), which are often judged subjectively, inverse problems have a precise objective: to recover the ground-truth signal from incomplete measurements.

When the measurement is heavily degraded, the inverse problem becomes highly ill-posed as many reconstructions can explain the measurements almost equally well\rev{, and the measurement alone cannot tell them apart}. A practical solution is to incorporate \textbf{side information}, defined as auxiliary measurements perceptually related to the target signal, to constrain the solution space and guide reconstruction. This idea is well established in classical signal processing, where structural or encoded properties are used to guide the iterative algorithms that solve the inverse problem \citep{joens2009localminimax, chun2012nonlocalmeans, oymak2013simple, ehrhardt2014jointmrirecon, mota2017cswithprior, hyder2019fourierphasewithsi}.  In medical imaging, complementary measurements or modalities, such as multiple MRI contrasts, multimodal microscopy, or RGB guidance for NIR imaging, have been shown to substantially improve quality \citep{atalik2025mriwithsideinfo,  tsiligianni2019sparselinwithsideinfo}.

\begin{figure*}[t]
    \vspace{-5pt}
    \centering
    \includegraphics[width=0.93\textwidth]{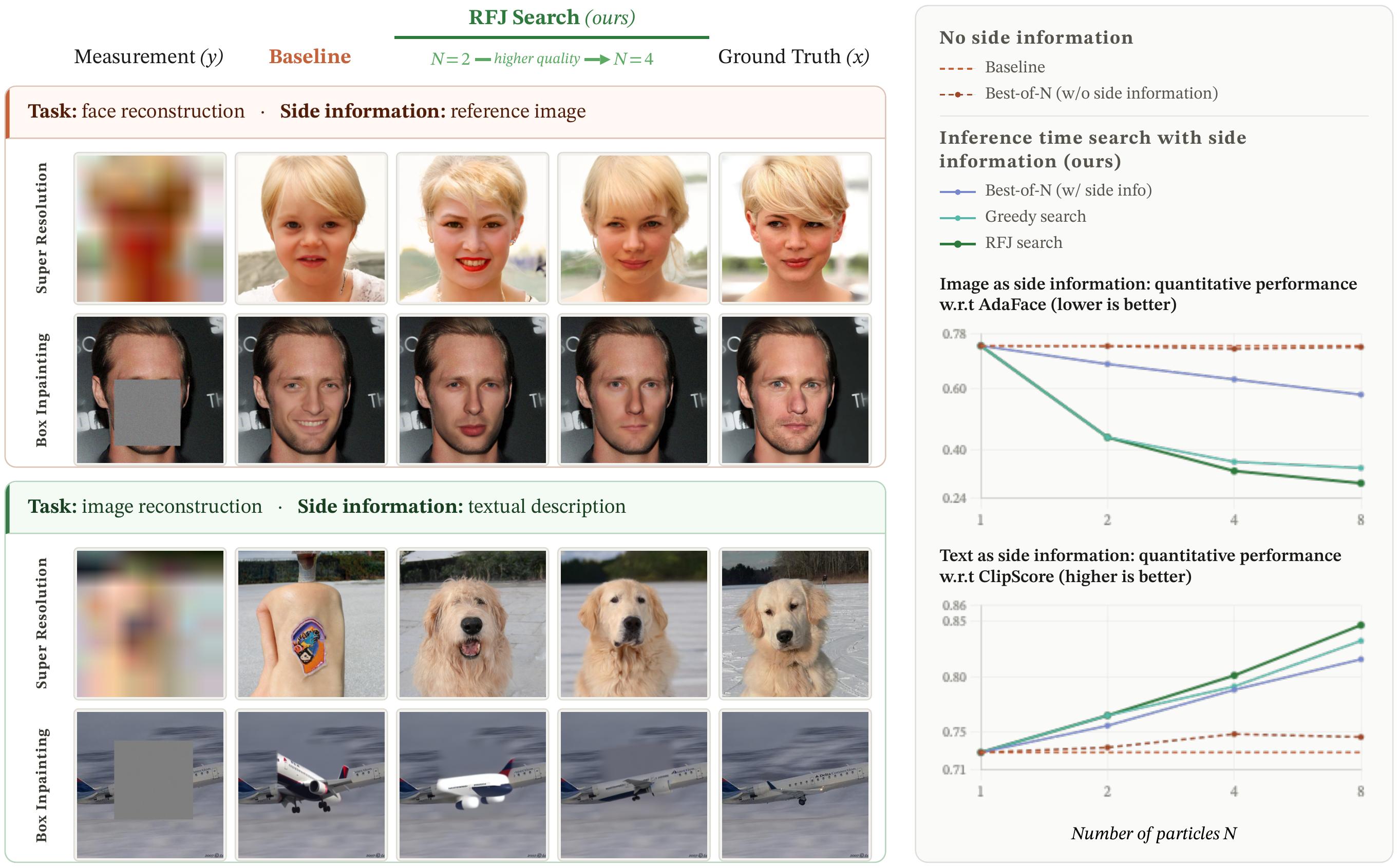}
    \caption{Incorporating side information via inference-time search improves reconstructions. (Left) Our framework augments state-of-the-art inverse problem solvers \citep{chung2023dps, chung2023blindps, zhang2024daps, he2024mpgd} with side information through inference-time search, yielding perceptually superior reconstructions with quality further improving as the number of particles N increases. (Right) All search algorithms within our side-information framework scale significantly better with additional compute than baselines without side information, demonstrating our framework's effectiveness.}
    \label{fig:teaser}
\end{figure*}

While the existing works on diffusion-based solvers have made significant progress on inverse problems,  they largely sidestep the harder and increasingly common setting where we must also exploit side information (e.g., a reference photograph of the same person, a text description, or features from another modality). A key obstacle is the challenge of learning the conditional distribution $p_{X \mid Y, S}$, where $X$ denotes the target image, $Y$ denotes the noisy measurement, and $S$ denotes the side information. While some recent works \citep{kim2025regularization, chung2025contextmri}  address the limited setting of textual side information, these approaches typically train a diffusion model to take a specific side information modality as input; this demands large paired datasets and expensive training, ties the solver to a single conditioning format, and is impractical when the test-time side information differs from what the model was trained on. This motivates us to address the following question:

\textit{How can we leverage a pre-trained (unconditional) diffusion prior to solve inverse problems with side information at inference time, without any retraining, so that the method is modality-agnostic and can use text, images, or features depending on the end-use applications?} We provide constructive solutions to these questions in our work. Our main contributions are the following.
\vspace{-0.2cm}
\begin{itemize}[leftmargin=*, itemsep=0.3em]
    \item \textbf{Modeling:} We introduce an approach that incorporates arbitrary side information via an auxiliary reward. This abstraction decouples the measurement model from the side information, is modality-agnostic (text, image, features), and requires no retraining. We derive tractable approximations with error bounds for the conditional score functions needed for sampling from pre-trained diffusion models.
    \item \textbf{Algorithm:} Motivated by inference-time search successes in LLMs \citep{snell2025scaling, setlur2025rewarding, liu2025can}, we incorporate side information via inference time search. We use an off-the-shelf reward function scoring reconstructions by consistency with the side information. We consider two search strategies: Greedy Search (GS), an SMC-based approach that periodically selects the highest reward particles, and Recursive Fork-Join Search (RFJS), which resamples within dynamic particle groups, preserving diversity and balancing exploration and exploitation. Our approach plugs into standard inverse problem solvers and supports black-box, non-differentiable reward functions. We demonstrate that inference-time search with side information yields substantial gains even with moderate compute (N=4) (Fig.~\ref{fig:teaser}, left), whereas without side information, scaling provides negligible improvement, as seen in Best-of-N without side information (Fig.~\ref{fig:teaser}, right).

    \item \textbf{Experiments:} We evaluate across linear and nonlinear inverse problems (e.g., box inpainting, super-resolution, and motion/Gaussian/nonlinear deblurring) and multiple side information types (reference images, textual descriptions, and MRI). Applying our method to several diffusion-based inverse problem solvers (DPS \citep{chung2023dps}, Blind DPS \citep{chung2023blindps}, DAPS \citep{zhang2024daps}, and MPGD \citep{he2024mpgd}) we show consistent improvements in perceptual reconstruction quality over all baselines.

    \item \textbf{Ablation:} We provide extensive ablation studies demonstrating robustness to the choice of reward function (Appendix~\ref{app:reward_ablation}), side information quality (Appendix~\ref{app: side_info_quality}), and the resampling hyperparameter $B$ across a range of values (Appendix~\ref{app:hyperparameter}). They also confirm that scaling inference-time compute \emph{without} side information fails to improve, or can even degrade, reconstruction quality (Table~\ref{tab:bon_si_ablation}), justifying the additional compute cost.

\end{itemize}

\vspace{-0.2cm}
\section{Related work}
\vspace{-0.2cm}

\textbf{Inverse problems with diffusion priors:}  Diffusion models \citep{dhariwal2021dmsbeatgans, ho2020ddpm, song2019generativemodellingbyestimatinggradients, sohl2015deepunsupervisedlearning, song2020improvedtechforscorebasedgenmodelling, song2021ddim} are powerful generative models that sample from data distributions by iteratively denoising random noise. Several works adapt diffusion priors to inverse problems via likelihood score approximations. Diffusion Posterior Sampling (DPS) \citep{chung2023dps} is a foundational method for solving inverse problems in a principled way. Its key idea is to approximate the expected conditional likelihood by evaluating the likelihood at the conditional mean. Recent work has focused on improving performance by leveraging the manifold constraints \citep{he2024mpgd} and by decoupling diffusion steps  \citep{zhang2024daps}.  Latent diffusion priors are also used: PSLD \citep{rout2023psld} adds consistency terms, ReSample \citep{song2024harddataconsistency} solves per-step optimization problems, and \citet{chung2024prompttuningforldms} tunes prompts for efficiency. \textit{None of these methods, however, leverage side information}.

\textbf{Inverse problems with side information:} Many works in signal processing \citep{mota2017cswithprior, oymak2013simple, joens2009localminimax, chun2012nonlocalmeans, ehrhardt2014jointmrirecon, hyder2019fourierphasewithsi} integrate structural correlations from auxiliary signals, often via designing appropriate optimization algorithms. Diffusion-based approaches include training with joint priors across modalities \citep{levac2023mrirw, efimov2025diffwithsideinfo}, metadata conditioning \citep{chung2025contextmri}, and text-guided regularization \citep{kim2025regularization}. In MRI, LeSITA \citep{tsiligianni2019sparselinwithsideinfo} learns coupled sparse representations, and TGVN \citep{atalik2025mriwithsideinfo} constrains ambiguous subspaces with additional contrasts using learned unrolled networks. Most approaches, however, are training-based or bound to one modality of side information associated with the trained conditional diffusion model.

\textbf{Inference-time search:} Reward-guided inference-time search has advanced LLM reasoning using Process Advantage Verifiers (PAVs) \citep{setlur2025rewarding}, compute-optimal scheduling \citep{snell2025scaling}, and reward-guided small models \citep{liu2025can}. Some recent works \citep{singhal2025fksteering, li2025dynamic} apply reward-based search in diffusion for text-to-image/protein generation, but do not consider side information or inverse problems.

\rev{\textbf{SMC methods:} Sequential Monte Carlo approaches \citep{cardoso2024monte, dou2024fpssmc, wu2023practicalexactsampling, achituve2025ldsmc} generate and resample particles according to how well they fit the measurement. These methods, however, do not leverage side information.}

\textbf{Other related work:} A detailed additional related work is provided in \cref{app: additional_related_work}.

\vspace{-0.2cm}
\section{Preliminaries and Problem Formulation}
\vspace{-0.2cm}

\subsection{Preliminaries}
\vspace{-0.1cm}

\label{ssec:prelims}

\textbf{Diffusion models:} Diffusion models \citep{ho2020ddpm, song2021scorebased} are powerful generative models in which a neural network learns to reverse a forward process that progressively corrupts samples from $p_{\mathrm{data}}$ with Gaussian noise, enabling generation by iterative denoising. The forward process is represented by the stochastic differential equation (SDE), $ \mathrm{d}{\vx}_t = f({\vx}_t, t)  \mathrm{d}t + g(t)  \mathrm{d}{\vw}_t,~ \forall t\in[0,T],$ where $\vw_t$ is a Wiener process. Common choices for $f, g$ are $f(\vx_t, t) = -({\beta(t)}/{2})\vx_t$ and $g(t)=\sqrt{\beta(t)}$ for non-negative monotonically increasing $\beta(\cdot)$. The corresponding reverse process of this SDE is described by \citep{anderson1983reversetimesde, song2021scorebased} $\mathrm{d} \vx_t = \left( f(\vx_t, t) - g^2(t)\nabla_{\vx_t} \log p_t(\vx_t) \right) \mathrm{d}t +  \rev{g(t)} \mathrm{d}\vw_t,~ \forall t \in [T, 0], $
where  $p_{t}$ is the marginal of $\vx_{t}$, $\vx_{T} \sim \mathcal{N}(0, I)$, and $\nabla_{\vx_{t}}\log p_{t}(\vx_{t})$ represents the \textit{score function}. Since $p_{t}$ is unknown, the score function is approximated by a neural network $\cD_{\btheta}(\vx_{t},t)$ via score-matching. In practice, the SDE is discretized into $T$ steps with  $\alpha_t \triangleq \prod_{s=1}^{t} (1-\beta_s)$.

\textbf{Solving inverse problems using diffusion models:} An inverse problem consists of recovering an unknown signal $\vx_0$ from noisy, partial measurements $\vy = \mA(\vx_0) + \sigma_y \vz$, where  $\mA$ is the measurement model, $\sigma_y$ is the measurement noise level, and $\vz$ is typically a Gaussian noise. Often,  $\mA$ is non-injective, making the problem ill-posed. A standard approach is Bayesian: assume a prior $p_0$ over the signal $\vx_0$ and sample from the posterior $\vx_0 \sim p_{0 \mid Y}(\cdot \mid \vy)$. Though $p_{0 \mid Y}$ is unknown, sampling can be achieved by replacing the score function in the backward SDE with the conditional score function $\nabla_{\vx_t} \log p_{t\mid Y}(\vx_t \mid \vy)$. Using Bayes' theorem, $\nabla_{\vx_t} \log p_{t\mid Y}(\vx_t \mid \vy) = \nabla_{\vx_t} \log p_t(\vx_t) + \nabla_{\vx_t} \log p_{Y \mid t}(\vy \mid \vx_t).$
While the score function network $\cD_{\btheta}$ of the pre-trained diffusion model can be used to approximate the first term, approximating the second term is significantly more challenging, and numerous approaches \citep{daras2024survey} have addressed this challenge. In particular, Diffusion Posterior Sampling (DPS) \citep{chung2023dps} proposes a simple approach to approximate $p_{Y \mid t}$ as  $p_{Y \mid t}(\vy \mid \vx_t)  = \bE_{\vx_0 \sim p_{0\mid t}(\cdot \mid \vx_t)}[p_{Y\mid 0}(\vy \mid \vx_0)] \approx p_{Y \mid 0}(\vy \mid \bE_{\vx_0 \sim p_{0\mid t}(\cdot \mid \vx_t)}[\vx_0])$, by pushing the expectation inside the nonlinear $p_{Y\mid 0}(\vy \mid \cdot)$. The remaining challenge is to compute the conditional mean $\bE_{\vx_0 \sim p_{0\mid t}(\cdot \mid \vx_t)}[\vx_0] \triangleq \hat{\vx}_{0\mid t}(\vx_t)$, which is typically tackled by using Tweedie's formula \citep{efron2011tweedie}, leveraging the fact that $\vx_t$ given $\vx_0$ is Gaussian. This results in the estimate
$\hat{\vx}_{0\mid t}(\vx_t) = ({1}/{\sqrt{\alpha_t}})(\vx_t + (1-\alpha_t) \nabla_{\vx_t} \log p_t(\vx_t)) \approx ({1}/{\sqrt{\alpha_t}})(\vx_t + (1-\alpha_t) \cD_{\btheta} (\vx_t, t)).$

\vspace{-0.1cm}
\subsection{Problem formulation: solving inverse problems with side information}
\vspace{-0.1cm}

In many applications, the measurement $\vy$ alone is insufficient to identify the latent signal $\vx_0$; auxiliary side information $\vs$ (e.g., a reference image, identity/text embedding, or physics-derived features) can dramatically reduce ambiguity. \rev{Throughout, $\vs$ is acquired independently of the measurement rather than computed from it, since anything derived from $\vy$ could add nothing that $\vy$ does not already contain.} Formally, when side information $\vs$ is available, \textit{the goal is to sample from the target conditional distribution $p_{0 \mid Y, S}(\cdot \mid \vy, \vs)$}. A seemingly direct route is to \textit{train a conditional diffusion model} that accepts $\vs$ as input, learn the conditional score function $\nabla_{\vx_t} \log p_{t \mid S}(\vx_t \mid \vs)$, and then approximate the full conditional score $ \nabla_{\vx_t} \log p_{t \mid Y, S}(\vx_t \mid \vy, \vs) = \nabla_{\vx_t} \log p_{t \mid S}(\vx_t \mid \vs) + \nabla_{\vx_t} \log p_{Y \mid t, S}(\vy \mid \vx_t, \vs)$ through a DPS-style method for the second term, to run the backward SDE. However, this training-based approach is  impractical: it demands large paired datasets $(\vx_{0}, \vs)$, which are expensive or impossible to curate; it locks the solver to the training modality of $\vs$; and general multi-modal conditioning requires prohibitive data and compute. These constraints motivate a training-free alternative that reuses strong unconditional diffusion priors and uses $\vs$ only at inference, preserving modality-agnosticism and avoiding costly data collection.

Designing such a training-free method is technically challenging. First, DPS-style derivations rely on tractable likelihoods (e.g., Gaussian $p_{Y\mid 0}$), whereas realistic $p_{S\mid 0}$ are often non-Gaussian, complicating conditional-score construction. Second, even for measurement-only guidance, DPS-style algorithms require back-propagating through the denoiser at every step. Naively extending to side information forces second-order/Hessian terms through the diffusion network. Third, purely gradient-guided diffusion is brittle: it struggles with non-differentiable or black-box rewards, amplifies early-step errors, and can drift off the data manifold. Inference-time search, which has shown remarkable gains in LLMs \citep{setlur2025rewarding, liu2025can, snell2025scaling} and text-conditioned diffusion \citep{singhal2025fksteering, kim2025testtime}, but not yet in inverse problems, offer a promising path to overcome these challenges. In this context, we address the following questions:

$(i)$ \textit{Modeling:} How can we realize $p_{0\mid Y,S}$ at inference time, without any retraining, by constructing a surrogate objective that is valid across diverse side-information modalities? \\$(ii)$ \textit{Algorithm}: How can we design a plug-and-play inference-time search module that is modality-agnostic, compute-aware, and capable of making global corrections (beyond local gradient steps)?

\vspace{-0.2cm}
\section{Modeling and Algorithm}
\label{sec:methodology}
\vspace{-0.2cm}

\subsection{Modeling side information using reward function}
\label{ssec:reward-modeling}
\vspace{-0.1cm}

Given a side-information signal $\vs$ corresponding to an unknown $\vx_0$, and two candidate reconstructions,  $\vx_0^1$ and  $\vx_0^2$, a principled way to decide which reconstruction is more truthful is to compare the (unknown) conditional probabilities $p_{0\mid S}(\vx_0^1\mid \vs)$ and $p_{0\mid S}(\vx_0^2\mid \vs)$. Directly estimating $p_{0\mid S}$ is intractable: it is typically non-Gaussian, multi-modal, and modality dependent. We therefore introduce a reward function $r:\mathbb{R}^d\times \mathcal{S}\to\mathbb{R}$ that orders reconstructions given $\vs$: if $r(\vx_0^1,\vs)>r(\vx_0^2,\vs)$, then $\vx_0^1$ is deemed more compatible with $\vx_0$ than $\vx_0^2$.  This abstraction aligns with many real-world applications (as shown in our experiments):  when $\vs$ is a text description of the target image $\vx_{0}$, we can use a pre-trained text-image model to score text-image alignment.  When $\vs$ is a reference image of the same entity (e.g., the same person under different poses/lighting), we can use a pre-trained network to score identity similarity. Such rewards are typically available and serve as practical surrogates for $p_{0\mid S}$ without requiring an explicit conditional density model. \rev{Although $\vs$ is written as a single signal, the formulation generalizes to $K$ reference signals $\vs^1,\dots,\vs^K$ by replacing $r(\vx_0; \vs)$ with an aggregate such as $\frac{1}{K}\sum_{k=1}^{K} r(\vx_0; \vs^k)$, which rewards equal alignment to all references, or $\max_{1 \le k \le K} r(\vx_0; \vs^k)$, which rewards alignment to the closest reference (see Table~\ref{tab:multi_reference} for a quantitative comparison).}

Our key modeling choice is to use $r(\cdot, \vs)$ to implicitly characterize  $p_{0\mid S}(\cdot\mid \vs)$ by tilting the unconditional prior $p_0$ toward higher-reward regions. Our approach is inspired by LLM alignment \citep{rlhf-ouyang,rafailov2023dpo}, where generation is steered toward high-reward samples without deviating too much from the pre-trained prior $p_{0}$. This is typically formalized as a KL-regularized reward maximization problem, $\max_{p \in \cP} \big( \bE_{\vx \sim p}[r(\vx)] - \tau D_{\mathrm{KL}}(p \Vert p_0) \big)$, where  $\tau > 0$ offers the trade-off between the deviation from the prior and reward maximization.  This optimization problem admits a closed-form solution,  $p^*(\vx) \propto p_0(\vx)\exp(r(\vx)/\tau)$ \citep{rafailov2023dpo}. Based on this, we \textit{define} our tractable surrogate for
$p_{0 \mid S}$ as
\begin{align}
\label{eq:proxy-posterior}
p_{0 \mid S}(\vx_0 \mid \vs) \propto p_0(\vx_0)\exp\left(\tfrac{r(\vx_0;\vs)}{\tau}\right),
\end{align}
\rev{Comparing \cref{eq:proxy-posterior} with Bayes' rule,
$p_{0 \mid S}(\vx_0 \mid \vs) = p_{S \mid 0}(\vs \mid \vx_0) p_0(\vx_0) / p_S(\vs)$, the above surrogate becomes the true posterior
precisely when, for every $\vx_0$}
\begin{align}
\label{eq:exactness}
\rev{\frac{r(\vx_0; \vs)}{\tau} = \log p_{S \mid 0}(\vs \mid \vx_0) + c(\vs)}
\end{align}
\rev{for some $c$ that does not depend on $\vx_0$; that is, when the reward is an affine function
of the log-likelihood with slope $1/\tau$. For each reward we use, a capable pretrained network
models $\log p_{S \mid 0}(\vs \mid \vx_0)$ approximately and makes it computable in closed form from the network's own
training objective, so \cref{eq:exactness} can be evaluated directly. \Cref{app:exactness} derives
this and \cref{app:reward_quality} measures it to show that the rewards used here meet the condition to a good
approximation, which justifies \cref{eq:proxy-posterior} as a model of $p_{0 \mid S}$.}
This assumption: $(i)$ preserves the powerful unconditional diffusion prior $p_0$, $(ii)$ injects modality-agnostic side information via a reward, and $(iii)$ produces a tractable objective that we can combine with the measurement model to target $p_{0\mid Y,S}$ at inference time using a pre-trained diffusion model. We do not claim optimality of \cref{eq:proxy-posterior}; rather, we show it leads to a practical, training-free algorithm that consistently improves reconstructions over strong baselines while keeping compute comparable. We now leverage \cref{eq:proxy-posterior} to compute the conditional posteriors for the reverse diffusion.

\begin{proposition} \label{prop:seq-diff-sampling}
    Let $p_{t\mid t+1, Y, S}$ denote the conditional posterior distribution for the reverse diffusion process. Then using \cref{eq:proxy-posterior} we have
    \begin{align} \label{eq:seq-diff-sampling}
        &p_{t\mid t+1, Y, S}(\vx_{t} \mid \vx_{t+1}, \vy, \vs) \propto  p_{t \mid t+1, Y}(\vx_t \mid \vx_{t+1}, \vy) \exp(V_t^\tau(\vx_t; \vs, \vy)), \\
        \label{eq:marginal-dist}
        &p_{t\mid Y, S}(\vx_{t} \mid \vy, \vs) \propto  p_{t \mid Y}(\vx_t \mid \vy) \exp(V_t^\tau(\vx_t; \vs, \vy)),
    \end{align}
    where the value function is given by $V_t^\tau(\vx_t; \vs, \vy) \triangleq \log \bE_{\vx_0 \sim p_{0\mid t, Y}(\cdot \mid \vx_t, \vy)}[\exp(r(\vx_0; \vs)/\tau)]$.
\end{proposition}
The proof is provided in Appendix \ref{proof:seq-diff-sampling}. Using \cref{eq:marginal-dist}, we can get the conditional score function as,
\begin{equation} \label{eq:ConditionalTimeDependentScoreFunctionSideInfo2}
\begin{split}
\nabla_{\vx_t} \log p_{t \mid Y, S}(\vx_t \mid \vy, \vs)
= \nabla_{\vx_t} \log p_{t}(\vx_t) + \nabla_{\vx_t} \log p_{Y\mid t}(\vy \mid \vx_t) + \nabla_{\vx_t} V_t^{\tau}(\vx_t; \vs, \vy).
\end{split}
\end{equation}

The computation of $V_t^\tau$ is not straightforward. So, we use a DPS-style approximation  as  $V_t^\tau(\vx_t; \vs, \vy) = \log \bE_{\vx_0 \sim p_{0\mid t, Y}(\cdot \mid \vx_t, \vy)}[\exp(r(\vx_0; \vs)/\tau)]
\approx r(\bE_{\vx_0 \sim p_{0\mid t, Y}(\cdot \mid \vx_t, \vy)}[\vx_0]; \vs)/\tau = r(\hat{\vx}_{0\mid t, Y}(\vx_t, \vy);\vs)/\tau  $. Using some approximation and the fact that $p_{Y\mid 0}$ is Gaussian, we can get
\begin{align}
         &\hat{\vx}_{0\mid t, Y}(\vx_t, \vy) \approx \hat{\vx}_{0\mid t}(\vx_t)
         \label{eq:dps-approximation-for-conditional-mean} - \rev{\left( \frac{1-\alpha_t}{\sqrt{\alpha_t}} \right)} \eta \nabla_{\vx_t} \norm{\vy-\mA \hat{\vx}_{0\mid t}(\vx_t)}_2^2, \\
    \label{eq:approx-value-function}
    &V_t^\tau(\vx_t; \vs, \vy) \approx \hat{V}_t^\tau(\vx_t; \vs, \vy) \triangleq r\left(\hat{\vx}_{0\mid t, Y}(\vx_t, \vy);\vs \right)/\tau.
\end{align}

Appendix \ref{proof:approx-error-bound} details the steps leading to \cref{eq:dps-approximation-for-conditional-mean}--\cref{eq:approx-value-function}, and provides a bound on the approximation error $\abs{V_t^\tau(\vx_t; \vs, \vy)- \hat{V}_t^\tau(\vx_t; \vs, \vy)}$ in \cref{prop:approx-error-bound}.

We can now get  $\nabla_{\vx_t} \log p_{t \mid Y, S}(\vx_t \mid \vy, \vs)$ given in \cref{eq:ConditionalTimeDependentScoreFunctionSideInfo2} by replacing $\nabla_{\vx_t} V_t^{\tau}(\vx_t; \vs, \vy)$   with $\nabla_{\vx_t} \hat{V}^\tau_t(\vx_t; \vy, \vs)$. However, running a backward diffusion using  $\nabla_{\vx_t} \hat{V}_t^{\tau}(\vx_t; \vs, \vy)$ is computationally infeasible because it involves computing second-order derivatives through the denoiser network (Eqs. \eqref{eq:dps-approximation-for-conditional-mean}--\eqref{eq:approx-value-function}). This issue, however,  can be circumvented by making a further approximation, by setting $\eta = 0$ in \cref{eq:dps-approximation-for-conditional-mean} to get $\hat{\vx}_{0\mid t, Y}(\vx_t, \vy) \approx \hat{\vx}_{0\mid t}(\vx_t)$, which leads to the approximation $\nabla_{\vx_t} V_t^{\tau}(\vx_t; \vs, \vy) \approx  \nabla_{\vx_t} r(\hat{\vx}_{0\mid t}(\vx_t);\vs)\rev{/\tau}$. We show that the approximation error remains small when $t$ is small even when $\eta = 0$, see  Appendix \ref{proof:approx-error-bound} and  \cref{app: discussion_on_eta}. This approach then reduces to the  \textbf{reward gradient guidance (RGG)} approach used for the inference-time alignment of diffusion models \citep{bansal2024universal, kim2025testtime, yu2023freedom, he2024mpgd}, with the critical difference being that the guidance is from both $\vs$ and $\vy$.

The RGG approach,  however, is limited only to differentiable rewards, and calculating a gradient through the denoiser network at each step of the backward diffusion is computationally intensive. Moreover, RGG is highly sensitive to the hyperparameter, leading to limited performance improvements and undesirable artifacts in the reconstructed images, as illustrated in~\cref{app:effect_of_grad_scale}. This motivates a gradient-free approach for leveraging the  side information in inverse problems.

\vspace{-0.1cm}
\subsection{Inference-Time Search Algorithms for Inverse Problems}
\vspace{-0.1cm}

\label{subsec:rkj-search}

Inference-time search have recently gained popularity in LLMs \citep{snell2025scaling, setlur2025rewarding, liu2025can}. Search algorithms solve a multi-step decision-making problem, balancing exploration and exploitation. While  Monte Carlo Tree Search (MCTS)  \citep{kocsis2006mcts} was successful in large-scale reinforcement learning systems like AlphaGo \citep{silver2016alphago}, they are infeasible for diffusion models: estimating the expected reward of a noisy state $\vx_t$ requires repeated rollouts. Using a value function to guide the search is an alternative, but this requires significant additional training for each modality of side information. These limitations motivate training-free, computationally tractable inference-time methods.

Particle-based procedures offer one such approach, using the distribution given by Eq.~\eqref{eq:seq-diff-sampling}, where the value function is replaced by the approximation in Eq.~\eqref{eq:approx-value-function}. At a given step, suppose we have $N$ samples $\vx_{t+1}[1], \dots, \vx_{t+1}[N] \sim p_{t+1 \mid Y, S}$. One way to generate samples from $p_{t \mid Y, S}$ is to $(i)$ propose candidates $\tilde{\vx}_{t}[i] \sim p_{t \mid t+1, Y}(\cdot \mid \vx_{t+1}[i], \vy)$, $(ii)$ compute rewards  $r[i] = r(\hat{\vx}_{0\mid t, Y}(\tilde{\vx}_{t}[i], \vy); \vs)$ (approximate value) $(iii)$ assign weights $w[i] \propto \exp(r[i]/\tau)$ and resample indices with replacement $I[i] \sim \mathrm{Cat}(w[1:N])$, and $(iv)$ retain $\vx_{t}[I[i]]$ for the next step. In theory, such particle methods converge to the target distribution as $N \to \infty$ ~\citep{wu2023practicalexactsampling, dou2024fpssmc}. In practice, frequent resampling collapses diversity by repeatedly duplicating high-reward particles, leading to under-exploration of the solution space. This motivates a more balanced resampling strategy.

To address this, we introduce \textbf{grouped resampling}: we partition the $N$ particles into groups of size $g_t$ and resample independently within each group, so particles only compete locally. This preserves between-group diversity while still exploiting high-reward particles within each group. Based on how we form these groups, we introduce two search algorithms.

\begin{wrapfigure}{r}{0.69\textwidth}
    \centering
    \vspace{-0.5cm}
    \caption{Illustration of inference-time search methods. Unlike GS, RFJS preserves diversity via dynamic grouping and within-group resampling, thus balancing exploration and exploitation.}
    \vspace{-0.1cm}
    \includegraphics[width=\linewidth]{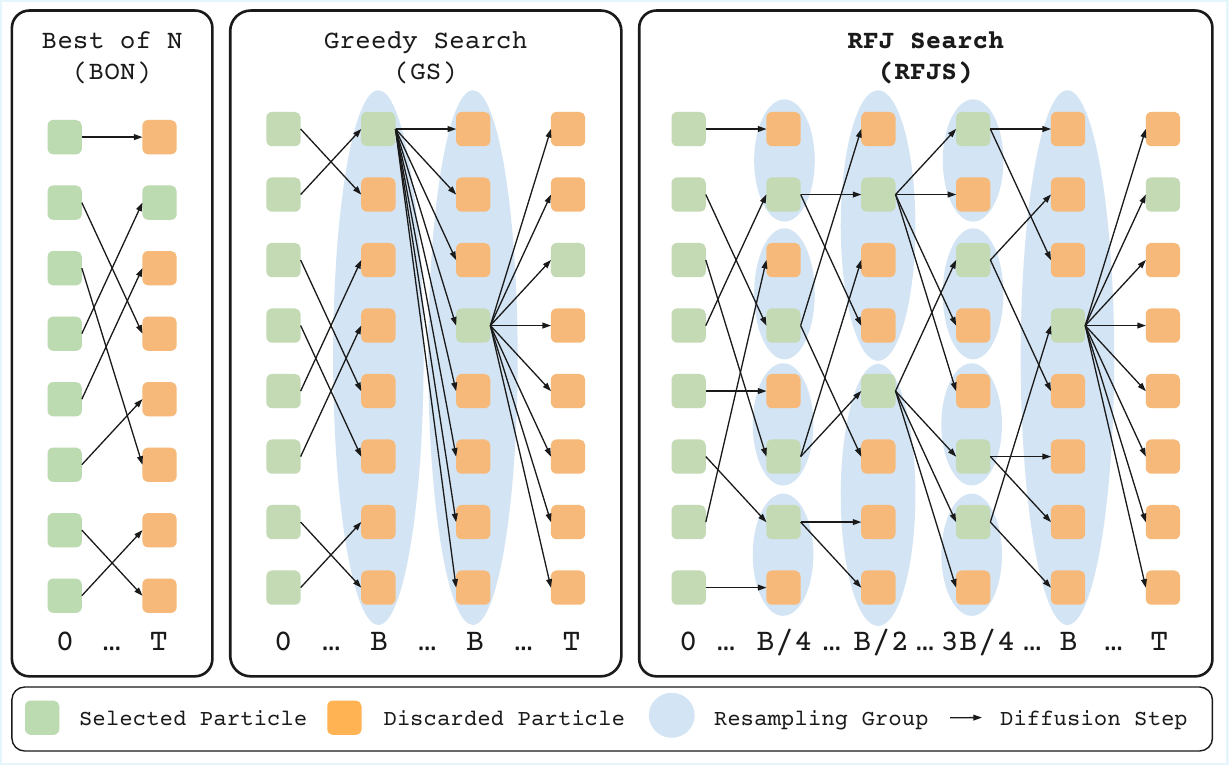}
    \label{fig:group_search}
    \vspace{-0.3cm}
\end{wrapfigure}

\textbf{Greedy Search (GS):} Here, we use a fixed resampling period $B$ and select  $g_t = N$ whenever $t \bmod B = 0$, and $g_t = 1$ otherwise.  Greedy Search reduces to the \textbf{Best-of-N} (BON) strategy when $B \geq T$, since in that case $g_t = 1$ for all $t$.
Smaller values of $B$ emphasize short-term reward exploitation, while larger values promote exploration.
An illustration of Greedy Search, with resampling interval $B$, is shown in Figure~\ref{fig:group_search}, where the particles evolve independently between resampling events and only interact at steps that are multiples of $B$.

\textbf{Recursive Fork-Join Search (RFJS):} To balance exploration and exploitation, we propose a hierarchical resampling schedule inspired by how researchers collaborate. As a PhD student, you meet your immediate collaborators frequently, your research group weekly, and the entire AI community once a year at conferences, and when a larger meeting is scheduled, the smaller one is canceled. We apply the same principle to particles: you have $N$ particles in total. Every $B$ steps, you resample across the entire group. Every $B/2$ steps, you resample within groups of size $N/2$, and so on. Formally, the group size at step $t$ is $g_t = N/2^{j^*}$, where $j^* = \min \{\, i \geq 0 :\; t \bmod (B/2^i) = 0 \,\}$.

Smaller groups are resampled more often while larger groups are resampled less frequently, resembling the higher frequency of small group meetings compared to large conferences. Particles within each group search the solution space locally, while keeping distinct groups independent from each other ensures a global search over the space, naturally balancing exploration and exploitation. Figure~\ref{fig:group_search} illustrates this strategy, where resampling frequency decreases as group size increases. We have summarized this inference-time search framework in Algorithm~\ref{alg:inference_search} in  Appendix~\ref{app:algorithm}.

Our framework is modular: regardless of the choice of search algorithm, it can be used to incorporate side information into the inference process. While any search algorithm can be paired with our framework, we find that RFJS scales better in our setting. Furthermore, RFJS simplifies hyperparameter tuning: because it naturally balances exploration and exploitation, $\tau$ (used for assigning the weights $w[i]$ for particles) can be set to a sufficiently small value \rev{($\tau \to 0$)} such that selection becomes deterministic, always retaining the particle with the highest reward, without sacrificing performance.  \rev{Every experiment in this paper uses this setting, in which only the relative \emph{ordering} of the rewards among the candidates matters and not their absolute values. For comparison, \cref{tab:stochastic_resampling} reports the sampler at the positive temperatures $\tau = 1/10$ and $\tau = 1/3$, where the side-information metric improves as $\tau$ decreases, justifying deterministic selection as our default.}

\vspace{-0.4cm}
\section{Experiments}
\label{sec:experiment_setup}

\vspace{-0.2cm}
\subsection{Experimental Setup}
\vspace{-0.2cm}

\begin{wraptable}{r}{0.5\textwidth}
\vspace{-0.4cm}
\caption{\textbf{Image as side information:} Quantitative results using DPS as the baseline solver. Our algorithms significantly improve perceptual quality as measured by FS, which is more aligned with human perception, while keeping classical metrics (PSNR, SSIM, LPIPS) in a similar range and slightly improving them.}
\label{tab:quantitative_results_dps}
\centering
\setlength{\tabcolsep}{1.5pt}
\resizebox{\linewidth}{!}{%
\begin{tabular}{l|cccc|cccc}
\toprule
& \multicolumn{4}{c|}{Box Inpainting} & \multicolumn{4}{c}{Super Resolution ($\times$4)} \\
\midrule
Algorithm & FS ($\downarrow$) & PSNR ($\uparrow$) & LPIPS ($\downarrow$) & SSIM ($\uparrow$) & FS ($\downarrow$) & PSNR ($\uparrow$) & LPIPS ($\downarrow$) & SSIM ($\uparrow$) \\
\midrule
RFJS (ours) & $\mathbf{0.308}$    & $\mathbf{28.29}$    & $\mathbf{0.136}$    & $\mathbf{0.855}$    & $\mathbf{0.380}$    & $\mathbf{25.26}$    & $\mathbf{0.225}$    & $\underline{0.695}$ \\
GS (ours)   & $\underline{0.349}$ & $\underline{28.22}$ & $\underline{0.137}$ & $\underline{0.855}$ & $\underline{0.460}$ & $\underline{25.24}$ & $\underline{0.225}$ & $\mathbf{0.696}$ \\
RGG         & $0.475$             & $27.96$             & $0.138$             & $0.851$             & $0.573$             & $25.13$             & $0.228$             & $0.690$ \\
BON (w/ side)        & $0.584$             & $28.20$             & $0.137$             & $0.854$             & $0.915$             & $25.14$             & $0.229$             & $0.694$ \\
DPS         & $0.739$             & $27.93$             & $0.139$             & $0.852$             & $1.042$             & $25.13$             & $0.229$             & $0.693$ \\
\midrule
& \multicolumn{4}{c|}{Non-linear Deblur} & \multicolumn{4}{c}{Motion Deblur} \\
\midrule
Algorithm & FS ($\downarrow$) & PSNR ($\uparrow$) & LPIPS ($\downarrow$) & SSIM ($\uparrow$) & FS ($\downarrow$) & PSNR ($\uparrow$) & LPIPS ($\downarrow$) & SSIM ($\uparrow$) \\
\midrule
RFJS (ours) & $\mathbf{0.394}$    & $\underline{23.89}$ & $\mathbf{0.229}$    & $\underline{0.668}$ & $\mathbf{0.326}$    & $\mathbf{26.64}$    & $\mathbf{0.193}$    & $\mathbf{0.736}$ \\
GS (ours)   & $\underline{0.467}$ & $\mathbf{23.92}$    & $0.232$             & $\mathbf{0.669}$    & $\underline{0.392}$ & $\underline{26.58}$ & $\underline{0.193}$ & $\underline{0.735}$ \\
RGG         & $0.654$             & $23.89$             & $\underline{0.231}$ & $0.665$             & $0.497$             & $26.55$             & $0.193$             & $0.733$ \\
BON (w/ side)        & $0.881$             & $23.89$             & $0.233$             & $0.667$             & $0.671$             & $26.57$             & $0.194$             & $0.735$ \\
DPS         & $1.008$             & $23.87$             & $0.232$             & $\underline{0.666}$ & $0.815$             & $26.54$             & $0.194$             & $0.734$ \\
\midrule
& \multicolumn{4}{c|}{Gaussian Deblur} & \multicolumn{4}{c}{Blind Deblur} \\
\midrule
Algorithm & FS ($\downarrow$) & PSNR ($\uparrow$) & LPIPS ($\downarrow$) & SSIM ($\uparrow$) & FS ($\downarrow$) & PSNR ($\uparrow$) & LPIPS ($\downarrow$) & SSIM ($\uparrow$) \\
\midrule
RFJS (ours) & $\mathbf{0.330}$    & $\mathbf{26.20}$    & $\mathbf{0.196}$    & $\mathbf{0.712}$    & $\mathbf{0.341}$    & $\underline{25.04}$ & $\mathbf{0.209}$    & $\underline{0.707}$ \\
GS (ours)   & $\underline{0.385}$ & $\underline{26.16}$ & $\underline{0.198}$ & $\underline{0.711}$ & $\underline{0.417}$ & $25.04$             & $0.211$             & $0.706$ \\
RGG         & $0.495$             & $26.15$             & $0.200$             & $0.709$             & $0.473$             & $24.97$             & $0.211$             & $0.701$ \\
BON (w/ side)        & $0.667$             & $26.18$             & $0.201$             & $0.711$             & $0.642$             & $\mathbf{25.15}$    & $\underline{0.210}$ & $\mathbf{0.708}$ \\
DPS         & $0.807$             & $26.15$             & $0.200$             & $0.711$             & $0.779$             & $24.98$             & $0.213$             & $0.704$ \\
\bottomrule
\end{tabular}%
}
\vspace{-0.4cm}
\end{wraptable}

We evaluate our inference-time search framework for solving inverse problems with side information by instantiating two search algorithms we proposed: \textbf{Greedy Search (GS)}  and \textbf{Recursive Fork Join Search (RFJS)}. We consider two types of side information: (i) \textbf{image side information}, a reference image of the same entity (e.g., the same person under different poses or lighting), and (ii) \textbf{text side information}, a textual description of the target image.

We demonstrate the plug-and-play nature of our algorithms by considering four different baseline inverse problem solvers: $(i)$ \textbf{DPS}  \citep{chung2023dps},  $(ii)$ \textbf{BlindDPS} \citep{chung2023blindps}, $(iii)$ \textbf{DAPS} \citep{zhang2024daps}, and $(iv)$ \textbf{MPGD} \citep{he2024mpgd} (reported in Appendix \ref{app:other_samplers}).

\textbf{Tasks:} We consider six inverse problems, covering both linear: $(i)$ box inpainting, $(ii)$ super resolution,  $(iii)$ motion deblur, $(iv)$ Gaussian deblur, and nonlinear: $(v)$ nonlinear deblur, $(vi)$ blind deblur. See Appendix \ref{app:exp_settings} for details.

\textbf{Baselines:} We compare the performance of GS and RFJS against the following baselines: $(i)$ \textit{Baseline solvers (DPS, BlindDPS, MPGD, DAPS)}, $(ii)$ \textit{Best-of-N (BoN)}, which generates $N$ independent samples and selects the one with the best reward at the end, $(iii)$ \textit{Reward Gradient Guidance (RGG)}, which solves the inverse problem by running the backward diffusion according to \cref{eq:ConditionalTimeDependentScoreFunctionSideInfo2}, but with the approximation $\nabla_{\vx_t} V_t^{\tau}(\vx_t; \vs, \vy) \approx  \nabla_{\vx_t} r(\hat{\vx}_{0\mid t}(\vx_t);\vs)\rev{/\tau}$. Experiments are run on NVIDIA A100 GPUs.

\setlength{\tabcolsep}{1pt}
\renewcommand{\arraystretch}{1.0}
\newcommand{\mf}[1]{\textcolor{blue}{[MF: #1]}}

\begin{figure}[t]
  \centering
  \caption{\textbf{Image as side information:} Qualitative results using DPS and BlindDPS baseline solvers. In severely ill-posed inverse problems, baseline solvers can fail in identity preservation, while RFJS remains effective; in milder settings, RFJS recovers finer details.}
  \label{fig:quantitative_results}
  \setlength{\tabcolsep}{2pt}
  \makebox[\linewidth]{
  \scalebox{0.6}{
  \begin{tabular}{c c}
    \begin{tabular}{>{\centering\arraybackslash}m{0.5cm} *{5}{>{\centering\arraybackslash}m{2.0cm}}}
      & \small Side & \small Meas. & \small DPS & \small RFJS (Ours) & \small GT \\
      {\centering\rotatebox{90}{\footnotesize Box Inpainting }} &
      \includegraphics[width=\linewidth]{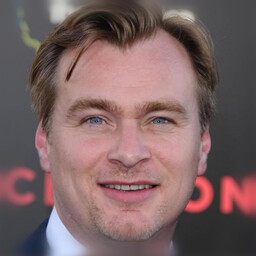} &
      \includegraphics[width=\linewidth]{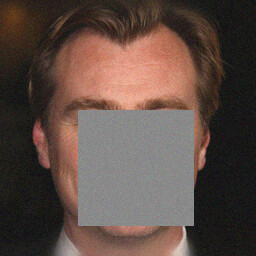} &
      \includegraphics[width=\linewidth]{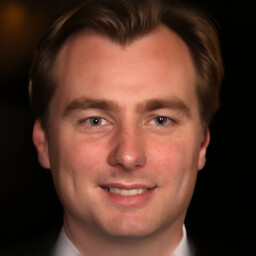} &
      \includegraphics[width=\linewidth]{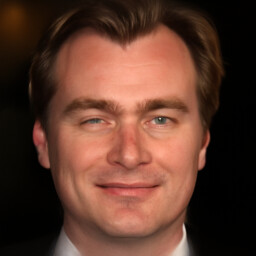} &
      \includegraphics[width=\linewidth]{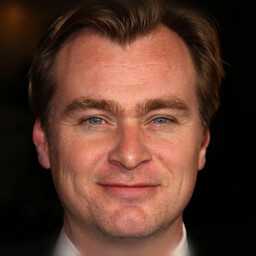} \\ [-2pt]
      {\centering\rotatebox{90}{\footnotesize Super Res. }} &
      \includegraphics[width=\linewidth]{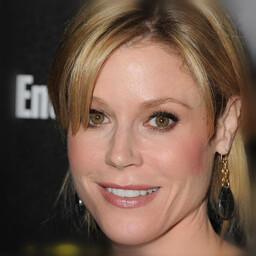} &
      \includegraphics[width=\linewidth]{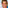} &
      \includegraphics[width=\linewidth]{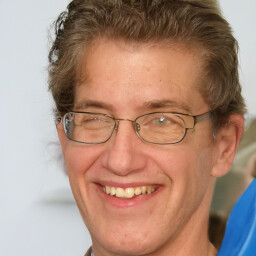} &
      \includegraphics[width=\linewidth]{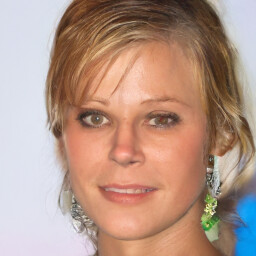} &
      \includegraphics[width=\linewidth]{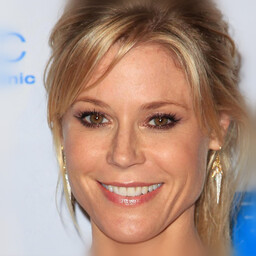} \\ [-2pt]
      {\centering\rotatebox{90}{\footnotesize Motion Deblur}} &
      \includegraphics[width=\linewidth]{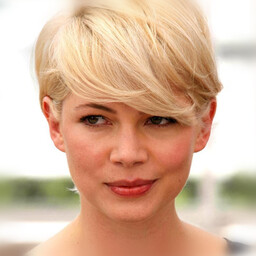} &
      \includegraphics[width=\linewidth]{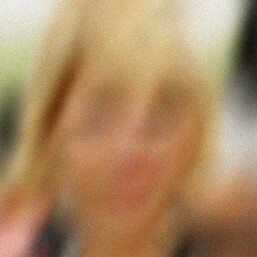} &
      \includegraphics[width=\linewidth]{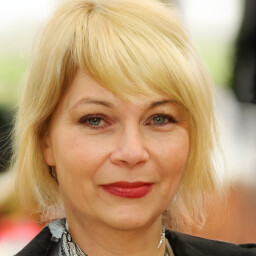} &
      \includegraphics[width=\linewidth]{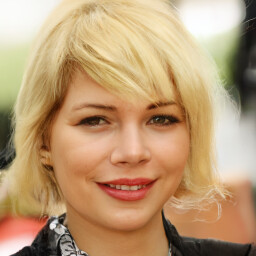} &
      \includegraphics[width=\linewidth]{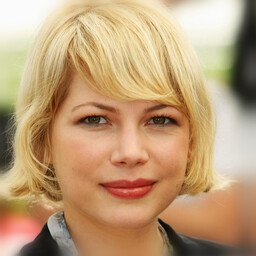} \\ [-2pt]
    \end{tabular}
    &
    \begin{tabular}{>{\centering\arraybackslash}m{0.5cm} *{5}{>{\centering\arraybackslash}m{2.0cm}}}
      & \small Side & \small Meas. & \small DPS & \small RFJS (Ours) & \small GT \\
      {\centering\rotatebox{90}{\footnotesize Gauss. Deblur }} &
      \includegraphics[width=\linewidth]{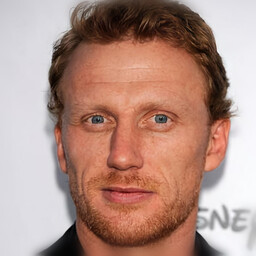} &
      \includegraphics[width=\linewidth]{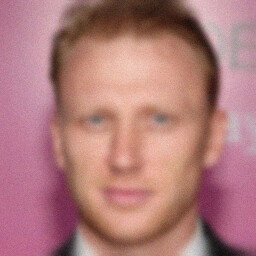} &
      \includegraphics[width=\linewidth]{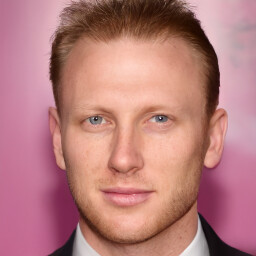} &
      \includegraphics[width=\linewidth]{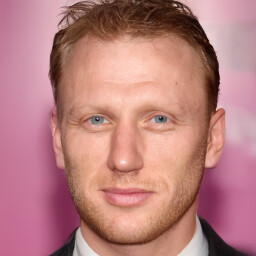} &
      \includegraphics[width=\linewidth]{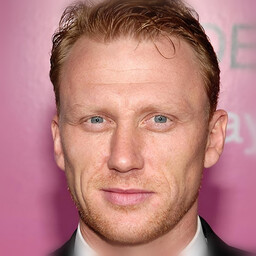} \\ [-2pt]
      {\centering\rotatebox{90}{\footnotesize Nonlin. Deblur }} &
      \includegraphics[width=\linewidth]{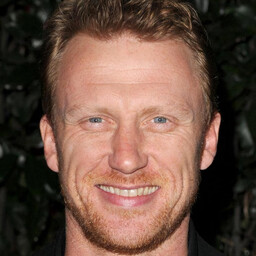} &
      \includegraphics[width=\linewidth]{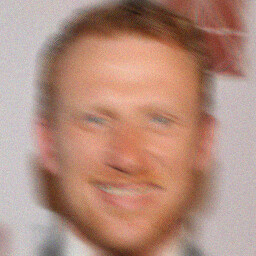} &
      \includegraphics[width=\linewidth]{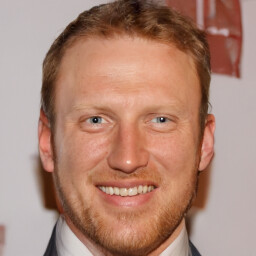} &
      \includegraphics[width=\linewidth]{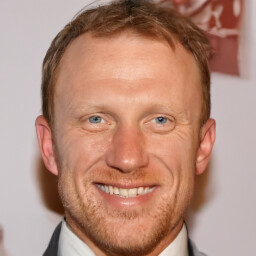} &
      \includegraphics[width=\linewidth]{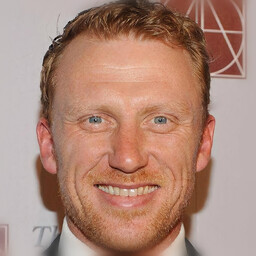} \\ [-2pt]
      {\centering\rotatebox{90}{\footnotesize Blind Deblur }} &
      \includegraphics[width=\linewidth]{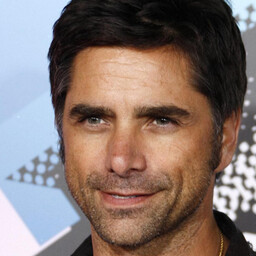} &
      \includegraphics[width=\linewidth]{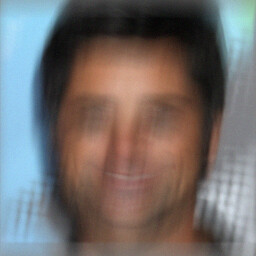} &
      \includegraphics[width=\linewidth]{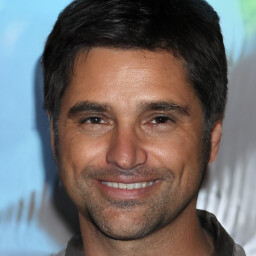} &
      \includegraphics[width=\linewidth]{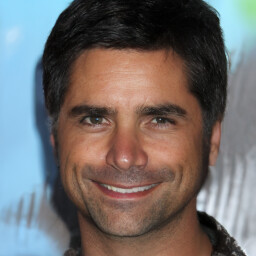} &
      \includegraphics[width=\linewidth]{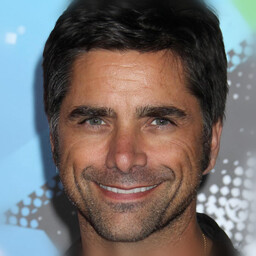} \\ [-2pt]
    \end{tabular}
  \end{tabular}
  }}
\end{figure}

\vspace{-0.3cm}
\subsection{Main Results}
\vspace{-0.2cm}

\begin{figure}[t]
  \centering
  \begin{minipage}[t]{0.48\textwidth}
    \centering
    \captionof{figure}{\textbf{Image as side information:} Qualitative results with DAPS as baseline solver.}
    \label{fig:DAPS_Hard}
    \vspace{4pt}
    \setlength{\tabcolsep}{2pt}
    \scalebox{0.55}{
    \begin{tabular}[t]{>{\centering\arraybackslash}m{0.6cm} *{5}{>{\centering\arraybackslash}m{2.0cm}}}
      & \small Side & \small Meas. & \small DAPS & \small RFJS (Ours) & \small GT \\
      {\centering\rotatebox{90}{\footnotesize Box Inpainting }} &
      \includegraphics[width=\linewidth]{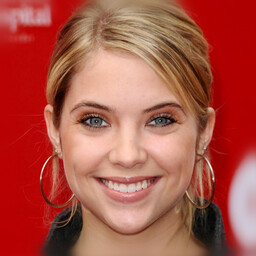} &
      \includegraphics[width=\linewidth]{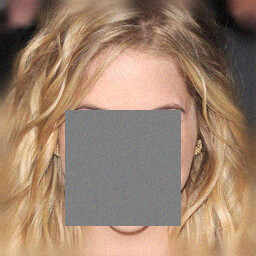} &
      \includegraphics[width=\linewidth]{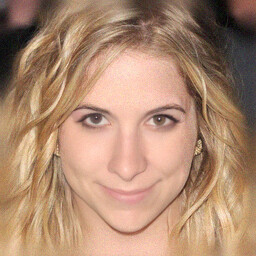} &
      \includegraphics[width=\linewidth]{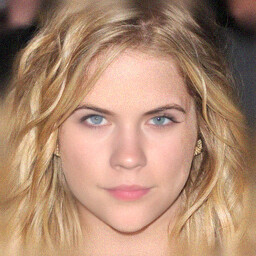} &
      \includegraphics[width=\linewidth]{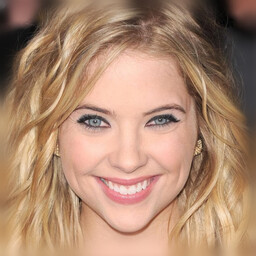} \\ [-2pt]
      {\centering\rotatebox{90}{\footnotesize Super Res. }} &
      \includegraphics[width=\linewidth]{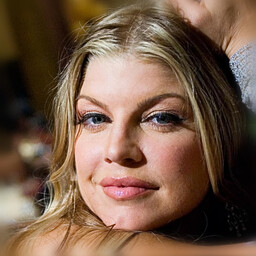} &
      \includegraphics[width=\linewidth]{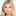} &
      \includegraphics[width=\linewidth]{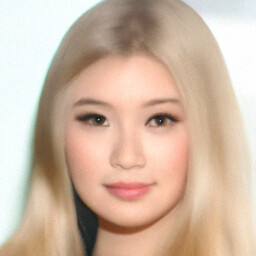} &
      \includegraphics[width=\linewidth]{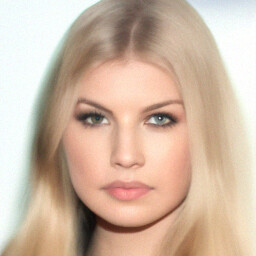} &
      \includegraphics[width=\linewidth]{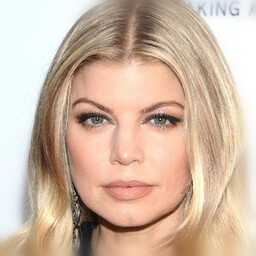} \\ [-2pt]
    \end{tabular}
    }
  \end{minipage}
  \hfill
  \begin{minipage}[t]{0.5\textwidth}
    \centering
    \captionof{table}{\textbf{Image as side information:} Quantitative results with DAPS as the baseline solver.}
    \label{tab:quantitative_results_daps1}
    \vspace{4pt}
    \setlength{\tabcolsep}{1.5pt}
    \resizebox{\linewidth}{!}{%
    \begin{tabular}{l|cccc|cccc}
    \toprule
    & \multicolumn{4}{c|}{Box Inpainting} & \multicolumn{4}{c}{Super Resolution ($\times 10$)} \\
    \midrule
    Algorithm & FS ($\downarrow$) & PSNR ($\uparrow$) & LPIPS ($\downarrow$) & SSIM ($\uparrow$) & FS ($\downarrow$) & PSNR ($\uparrow$) & LPIPS ($\downarrow$) & SSIM ($\uparrow$) \\
    \midrule
    RFJS (ours)   & $\textbf{0.423}$    & $\textbf{28.720}$    & $\textbf{0.140}$    & $\textbf{0.788}$    & $\underline{0.654}$ & $\underline{25.228}$ & $\textbf{0.282}$    & $\underline{0.661}$ \\
    GS (ours)     & $0.511$             & $28.640$             & $\underline{0.140}$ & $\underline{0.787}$ & $0.760$             & $\textbf{25.271}$    & $0.285$             & $\textbf{0.662}$ \\
    RGG                     & $\underline{0.436}$ & $28.410$             & $0.141$             & $0.784$             & $\textbf{0.579}$    & $25.210$             & $\underline{0.282}$ & $0.659$ \\
    BON (w/ side) & $0.611$             & $\underline{28.660}$ & $0.141$             & $0.787$             & $0.909$             & $25.220$             & $0.285$             & $0.660$ \\
    DAPS                    & $0.739$             & $28.290$             & $0.142$             & $0.784$             & $1.020$             & $25.170$             & $0.285$             & $0.659$ \\
    \bottomrule
    \end{tabular}%
    }
  \end{minipage}
\end{figure}

\textbf{Image as side information:}  The goal is to reconstruct a face image from a noisy measurement when another image of the same identity is available (Fig.~\ref{fig:quantitative_results}, \cref{fig:DAPS_Hard}, \cref{tab:quantitative_results_dps}, \cref{tab:quantitative_results_daps1}). Using CelebA-HQ \citep{karras2018progressive} as an out-of-distribution set and a diffusion model pretrained on FFHQ \citep{chung2023dps}, we sample two random images per identity for target and side information. We compute the reward as follows: first, detect the face using MTCNN \citep{zhang2016mtccn} and then extract identity features with AdaFace \citep{kim2022adaface}. Then, we measure the reward as the negative of \rev{FS}, computed as the distance between the identity embeddings of the reconstructed and side-information faces, extracted by pretrained AdaFace network.

Since standard metrics (PSNR, SSIM, LPIPS) often fail to capture identity similarity, we additionally use \rev{the face-similarity distance FS ($\downarrow$), the distance between the AdaFace identity embeddings of two faces, so that smaller values mean greater similarity. Reported as a metric, FS compares the reconstruction with the ground truth; the search reward is the negative of the same distance taken against the side information instead. We use the name FS throughout.} Table~\ref{tab:quantitative_results_dps} and Table~\ref{tab:quantitative_results_daps1} show that both proposed inference-time search methods, GS and RFJS, outperform baselines (DPS and DAPS respectively), with RFJS achieving the best overall scores.  Qualitative results in Fig~\ref{fig:quantitative_results} show sharper facial details and preserve identity. We used  $N = 8$ particles. \rev{The resampling base $B$ is fixed once for the base sampler and then used unchanged for every task, with no per-task tuning: in the face experiments it is $B = 16$ with DPS, $4$ with DAPS and $8$ with MPGD. \Cref{app:hyperparameter} reports the sensitivity to this choice, and \cref{app:exp_settings} lists the value used in each experiment.}

\begin{figure}[t]
  \centering
  \begin{minipage}[t]{0.46\textwidth}
    \centering
    \captionof{figure}{\textbf{Text as side information:} Qualitative results using textual side information under severe degradation (e.g., $32\times$ super-resolution with the prompt \textit{``golden retriever sitting on a snowy frozen lake, facing forward''}). RFJS recovers perceptually meaningful content while DPS fails.}
    \label{fig:qualitative_results_text_reward}
    \vspace{4pt}
    \scalebox{0.45}{
    \begin{tabular}{>{\centering\arraybackslash}m{0.5cm} *{4}{>{\centering\arraybackslash}m{3cm}}}
      & \small Meas. & \small DPS & \small RFJS (Ours) & \small GT \\
      {\centering\rotatebox{90}{\footnotesize Box Inpainting }} &
      \includegraphics[width=\linewidth]{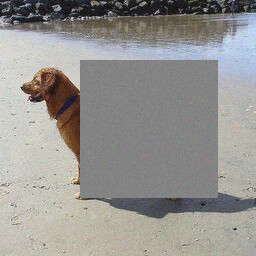} &
      \includegraphics[width=\linewidth]{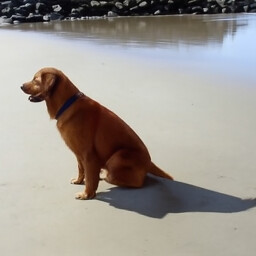} &
      \includegraphics[width=\linewidth]{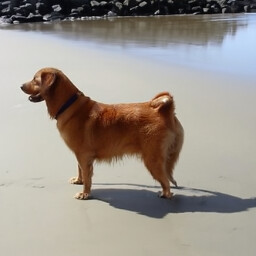} &
      \includegraphics[width=\linewidth]{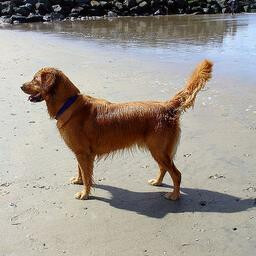} \\ [-2pt]
      {\centering\rotatebox{90}{\footnotesize Super Res. }} &
      \includegraphics[width=\linewidth]{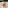} &
      \includegraphics[width=\linewidth]{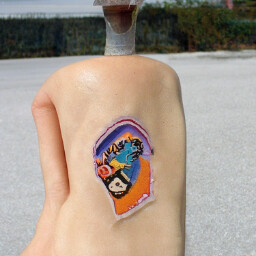} &
      \includegraphics[width=\linewidth]{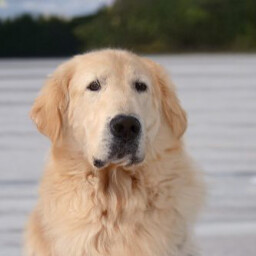} &
      \includegraphics[width=\linewidth]{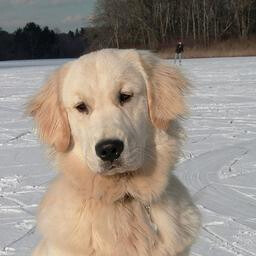} \\ [-2pt]
      {\centering\rotatebox{90}{\footnotesize Motion Deblur}} &
      \includegraphics[width=\linewidth]{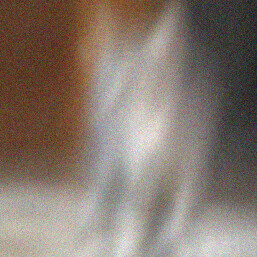} &
      \includegraphics[width=\linewidth]{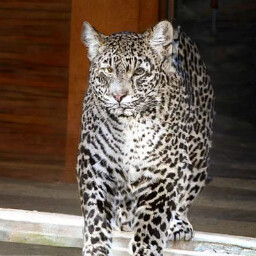} &
      \includegraphics[width=\linewidth]{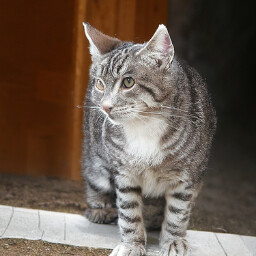} &
      \includegraphics[width=\linewidth]{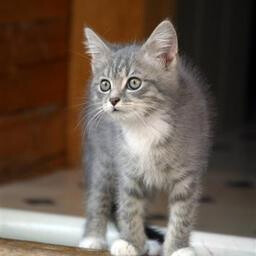} \\ [-2pt]
      {\centering\rotatebox{90}{\footnotesize Gauss. Deblur }} &
      \includegraphics[width=\linewidth]{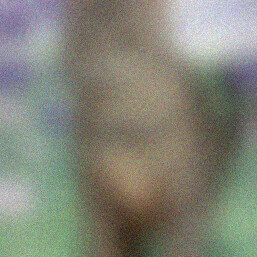} &
      \includegraphics[width=\linewidth]{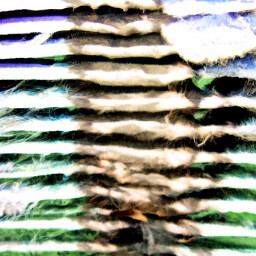} &
      \includegraphics[width=\linewidth]{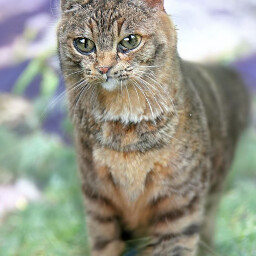} &
      \includegraphics[width=\linewidth]{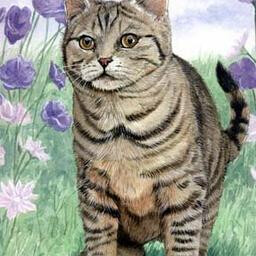} \\ [-2pt]
      {\centering\rotatebox{90}{\footnotesize Nonlin. Deblur }} &
      \includegraphics[width=\linewidth]{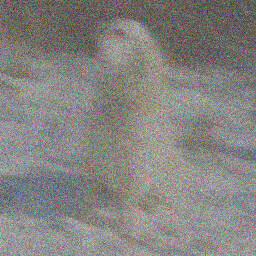} &
      \includegraphics[width=\linewidth]{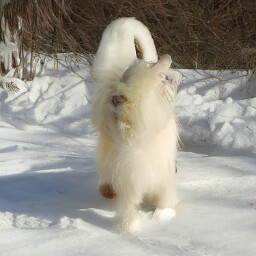} &
      \includegraphics[width=\linewidth]{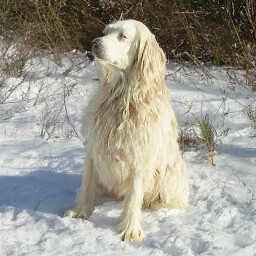} &
      \includegraphics[width=\linewidth]{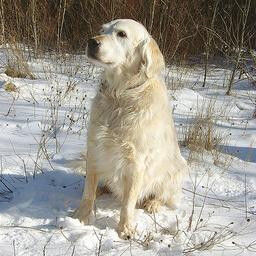} \\ [-2pt]
      {\centering\rotatebox{90}{\footnotesize Blind Deblur }} &
      \includegraphics[width=\linewidth]{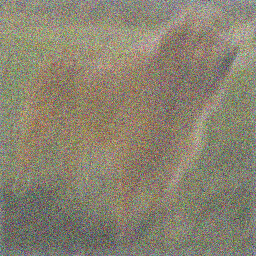} &
      \includegraphics[width=\linewidth]{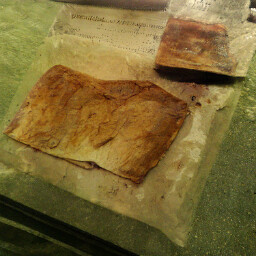} &
      \includegraphics[width=\linewidth]{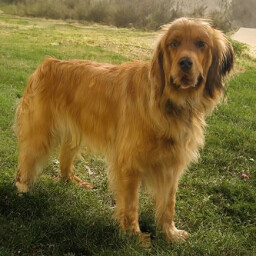} &
      \includegraphics[width=\linewidth]{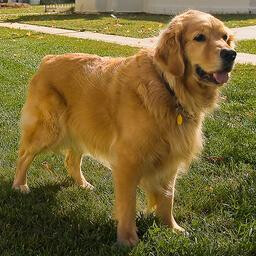} \\ [-2pt]
    \end{tabular}
    }
  \end{minipage}
  \hfill
  \begin{minipage}[t]{0.5\textwidth}
    \centering
    \captionof{table}{\textbf{Text as side information:} Quantitative comparison of our algorithms with the baseline. RFJS and GS attain the best CS on every task and the best value in $19$ of the $24$ cells, with Best-of-$N$ leading in the remaining five.}
    \label{tab:quantitative_results_text_reward}
    \vspace{4pt}
    \setlength{\tabcolsep}{2pt}
    \resizebox{\linewidth}{!}{%
    \begin{tabular}{l|cccc|cccc}
    \toprule
    & \multicolumn{4}{c|}{Box Inpainting} & \multicolumn{4}{c}{Super Resolution ($\times$32)} \\
    \midrule
    Algorithm & CS ($\uparrow$) & PSNR ($\uparrow$) & SSIM ($\uparrow$) & LPIPS ($\downarrow$) & CS ($\uparrow$) & PSNR ($\uparrow$) & SSIM ($\uparrow$) & LPIPS ($\downarrow$) \\
    \midrule
    RFJS (ours) & $\mathbf{0.901}$    & $\mathbf{20.75}$    & $\mathbf{0.678}$    & $\mathbf{0.294}$    & $\mathbf{0.801}$    & $17.13$             & $\mathbf{0.352}$    & $\mathbf{0.493}$ \\
    GS (ours)   & $\underline{0.894}$ & $19.76$             & $\underline{0.676}$ & $\underline{0.305}$ & $\underline{0.791}$ & $\underline{17.20}$ & $\underline{0.351}$ & $0.509$ \\
    BON (w/ side)        & $0.882$             & $\underline{19.99}$ & $0.672$             & $0.308$             & $0.788$             & $\mathbf{17.21}$    & $0.350$             & $\underline{0.500}$ \\
    DPS         & $0.871$             & $19.86$             & $0.672$             & $0.312$             & $0.731$             & $16.90$             & $0.330$             & $0.522$ \\
    \midrule
    & \multicolumn{4}{c|}{Non-linear Deblur} & \multicolumn{4}{c}{Motion Deblur} \\
    \midrule
    Algorithm & CS ($\uparrow$) & PSNR ($\uparrow$) & SSIM ($\uparrow$) & LPIPS ($\downarrow$) & CS ($\uparrow$) & PSNR ($\uparrow$) & SSIM ($\uparrow$) & LPIPS ($\downarrow$) \\
    \midrule
    RFJS (ours) & $\underline{0.863}$ & $\mathbf{20.58}$ & $\mathbf{0.473}$ & $\mathbf{0.405}$ & $\mathbf{0.858}$    & $\underline{18.61}$ & $\underline{0.402}$ & $\mathbf{0.424}$ \\
    GS (ours)   & $\mathbf{0.865}$    & $20.32$          & $0.456$          & $0.405$          & $0.835$             & $17.83$             & $0.369$             & $0.453$ \\
    BON (w/ side)        & $0.855$             & $20.52$          & $0.464$          & $0.406$          & $\underline{0.848}$ & $\mathbf{19.24}$    & $\mathbf{0.415}$    & $\underline{0.427}$ \\
    DPS         & $0.839$             & $20.55$          & $0.469$          & $0.409$          & $0.794$             & $18.16$             & $0.384$             & $0.458$ \\
    \midrule
    & \multicolumn{4}{c|}{Gaussian Deblur} & \multicolumn{4}{c}{Blind Deblur} \\
    \midrule
    Algorithm & CS ($\uparrow$) & PSNR ($\uparrow$) & SSIM ($\uparrow$) & LPIPS ($\downarrow$) & CS ($\uparrow$) & PSNR ($\uparrow$) & SSIM ($\uparrow$) & LPIPS ($\downarrow$) \\
    \midrule
    RFJS (ours) & $\mathbf{0.843}$    & $\mathbf{18.10}$    & $\underline{0.358}$ & $\underline{0.457}$ & $\mathbf{0.851}$    & $\underline{18.84}$ & $\underline{0.412}$ & $\mathbf{0.433}$ \\
    GS (ours)   & $\underline{0.835}$ & $17.96$             & $0.356$             & $0.457$             & $\underline{0.835}$ & $\mathbf{18.93}$    & $\mathbf{0.414}$    & $\underline{0.438}$ \\
    BON (w/ side)        & $0.831$             & $\underline{17.99}$ & $\mathbf{0.365}$    & $\mathbf{0.452}$    & $0.831$             & $18.78$             & $0.410$             & $0.443$ \\
    DPS         & $0.778$             & $16.79$             & $0.329$             & $0.487$             & $0.793$             & $18.82$             & $0.409$             & $0.459$ \\
    \bottomrule
    \end{tabular}%
    }

    \vspace{10pt}

    \captionof{figure}{\textbf{Contrast Image as Side Information:} Qualitative MRI reconstruction with RFJS vs.\ ContextMRI. The shapes and line edges are well preserved in our reconstruction.}
    \label{fig:mri-qualitative}
    \vspace{4pt}
    \resizebox{\linewidth}{!}{%
    \begin{tabular}{>{\centering\arraybackslash}m{0.5cm} *{5}{>{\centering\arraybackslash}m{2.2cm}}}
      & \small Side & \small Meas. & \small ContextMRI & \small RFJS (Ours) & \small GT \\
      {\centering\rotatebox{90}{\footnotesize PDFS with PD}} &
      \includegraphics[width=\linewidth]{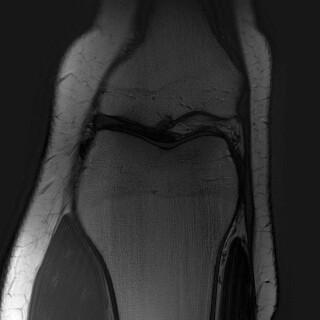} &
      \includegraphics[width=\linewidth]{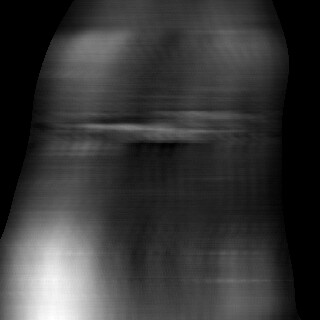} &
      \includegraphics[width=\linewidth]{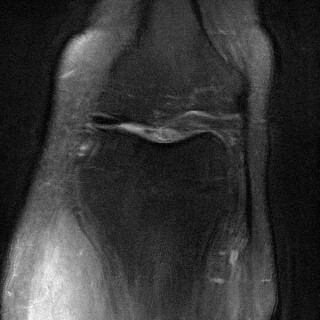} &
      \includegraphics[width=\linewidth]{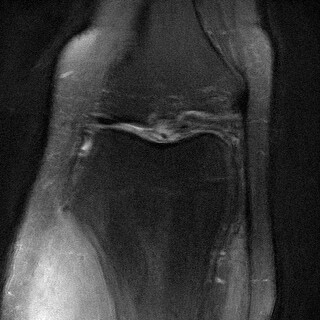} &
      \includegraphics[width=\linewidth]{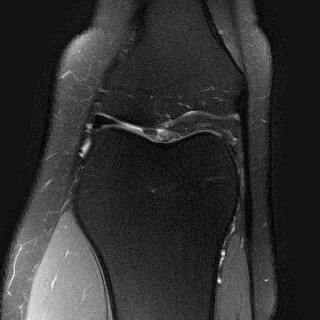} \\ [-2pt]
      {\centering\rotatebox{90}{\footnotesize PD with PDFS}} &
      \includegraphics[width=\linewidth]{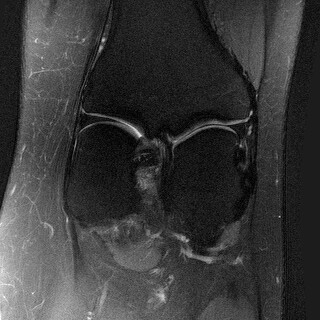} &
      \includegraphics[width=\linewidth]{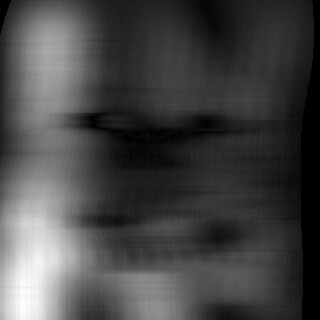} &
      \includegraphics[width=\linewidth]{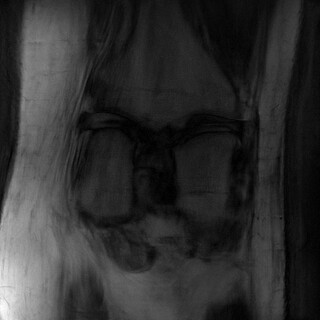} &
      \includegraphics[width=\linewidth]{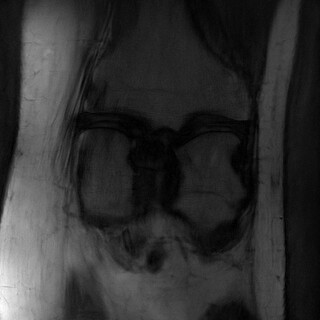} &
      \includegraphics[width=\linewidth]{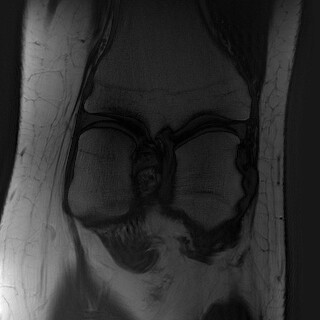} \\
    \end{tabular}%
    }
  \end{minipage}
\end{figure}

\textbf{Text as side information:}  The goal is to reconstruct an image from its noisy measurement, with a text description of the image available as side information. We use a pre-trained diffusion model trained on the ImageNet data \citep{dhariwal2021dmsbeatgans}. We use 25 images from the ImageNet validation set to evaluate the algorithms and generated a short one-sentence textual description for each image using ChatGPT as the side information. We use ImageReward \citep{xu2023imagereward}, a pre-trained network that measures text-to-image similarity, as the reward function. We consider some inverse problem tasks that are significantly challenging, including $32\times$ super resolution, and strong blur with larger kernels. Experiments use $N{=}4$ and $B{=}100$, and we report the standard metrics and CLIPScore \citep{radford2021clip}. CLIPScore measures the cosine similarity between CLIP image embeddings of the ground truth and reconstruction, providing a semantically informed metric that reflects both visual and textual alignment. The qualitative reconstructions in Fig.~\ref{fig:qualitative_results_text_reward} closely match the textual descriptions, and the quantitative metrics in Table~\ref{tab:quantitative_results_text_reward} show that both GS and RFJS outperform competing baselines. \rev{\Cref{app:text75} repeats this evaluation on $75$ images.}

\textbf{MRI with multi-contrast side information:}
\rev{We include this modality to show that the framework carries over to scientific imaging, where the side information is a second acquisition of the same anatomy rather than a photograph or a caption (\cref{app: mri_experiments} gives the setup in full).}
We use fastMRI knee dataset \citep{zbontar2018fastmri} with the ContextMRI model \citep{chung2025contextmri}. We pair PD and PDFS contrasts, reconstructing one from the other under  $16\times$ undersampling with 2\% ACS. We use normalized mutual information (NMI) as reward, which is robust to contrast changes. Fig.~\ref{fig:mri-qualitative} highlights sharper edges and more faithful structure. Table~\ref{tab:mri-quantitative-results} in Appendix \ref{app: mri_experiments} shows our methods consistently improve the baseline.

\begin{wraptable}{r}{0.5\textwidth}
\vspace{-0.4cm}
\caption{Without side information, Best-of-N selection by $\|\vy - \mA\vx_0^i\|$ fails to improve over the baseline in noisy settings. With side information, selection by our reward $r(\vx_0;\vs)$ consistently improves all metrics.}
\label{tab:bon_si_ablation_short}
\centering
\setlength{\tabcolsep}{2pt}
\resizebox{\linewidth}{!}{%
\begin{tabular}{l|cccc|cccc}
\toprule
& \multicolumn{4}{c|}{Box Inpainting} & \multicolumn{4}{c}{Super Resolution} \\
\midrule
Algorithm & PSNR ($\uparrow$) & LPIPS ($\downarrow$) & SSIM ($\uparrow$) & CS ($\uparrow$) & PSNR ($\uparrow$) & LPIPS ($\downarrow$) & SSIM ($\uparrow$) & CS ($\uparrow$) \\
\midrule
DAPS              & $18.21$          & $0.274$          & $0.742$          & $0.855$          & $18.44$          & $0.583$          & $0.391$          & $0.727$ \\
BON (w/o SI)      & $18.02$          & $0.274$          & $0.738$          & $0.846$          & $18.13$          & $0.590$          & $0.368$          & $0.729$ \\
BON (w/ SI, ours) & $\mathbf{19.51}$ & $\mathbf{0.262}$ & $\mathbf{0.750}$ & $\mathbf{0.893}$ & $\mathbf{18.65}$ & $\mathbf{0.579}$ & $\mathbf{0.398}$ & $\mathbf{0.767}$ \\
\bottomrule
\end{tabular}%
}
\vspace{-0.3cm}
\end{wraptable}

\textbf{Role of side information in enabling inference-time scaling:}
A central contribution of our framework is the principled use of side information through a reward that measures consistency between the reconstruction and the side information at inference time. To see why this is essential, consider the natural alternative: scaling compute via Best-of-N using only the measurement residual $\|\vy - \mA \vx_0^i\|$ as the selection criterion (BoN w/o SI). When $\vy$ is noisy or the inverse problem is ill-posed, this criterion is actively misleading as it favors particles that explain the noise rather than the underlying signal, and as Table~\ref{tab:bon_si_ablation_short} shows, performance \emph{can degrade} relative to the baseline. Incorporating side information fundamentally changes this picture: BoN with our reward (BoN w/ SI) draws samples that are more faithful to the conditional posterior $p_{X \mid Y, S}$, resolving ambiguities that the measurement alone cannot (see Table~\ref{tab:bon_si_ablation} for more tasks).

\textbf{Perceptual Quality vs.\ Classical Metrics:} Recent works \citep{lin2025pixelwisemetricsreliablesparseview, su2025rethinkingimageevaluationsuperresolution} have shown that improvements in perceptual quality for inverse problems do not necessarily translate into better values of classical metrics such as PSNR, SSIM, and LPIPS, and they advocate evaluating with task-specific metrics. In our setting, we use FS for face reconstruction and CLIPScore for ImageNet as task-specific metrics, which are also more aligned with human perceptual judgments for these tasks. \rev{Because FS and the AdaFace reward share a model, we additionally report three
other perceptual metrics not derived from the same model: DreamSim \citep{fu2023dreamsim}, a perceptual distance fit to
human similarity judgements; the CLIP distance between the reconstruction and the ground truth;
and the normalised landmark error NME, the mean distance between the $68$ FAN landmarks
\citep{bulat2017far} of the reconstruction and of the ground truth over the latter's
inter-ocular distance, which measures facial geometry rather than identity. All
three improve alongside FS (\cref{tab:perceptual_metrics_main}, and all twelve tasks in
\cref{tab:perceptual_metrics}) indicating better (perceptual) reconstruction rather than simply target reward optimization.}

\begin{wrapfigure}{r}{0.4\textwidth}
    \centering
    \vspace{-0.4cm}
    \caption{Classical metrics fail to capture perceptual fidelity: three BlindDPS reconstructions achieve higher PSNR values (23.5, 23.6, 23.9) than RFJS (23.2), yet RFJS exhibits clearly stronger perceptual similarity to the ground truth. The same pattern holds for LPIPS/SSIM.}
    \label{fig:identity_preservation_main}
    {\scriptsize
    \begin{tabular}{ccc}
        \includegraphics[width=0.3\linewidth]{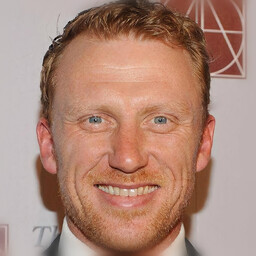} &
        \includegraphics[width=0.3\linewidth]{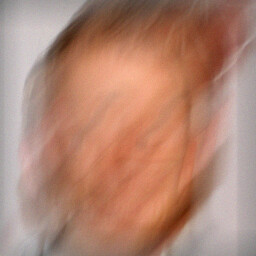} &
        \includegraphics[width=0.3\linewidth]{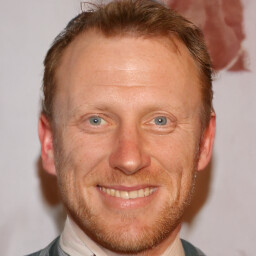} \\
        Ground truth & Input & RFJS (Ours) \\[6pt]
        \includegraphics[width=0.3\linewidth]{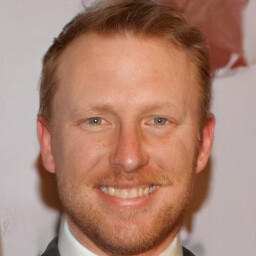} &
        \includegraphics[width=0.3\linewidth]{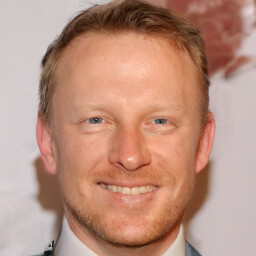} &
        \includegraphics[width=0.3\linewidth]{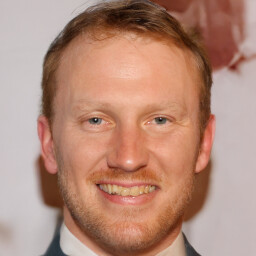} \\
        DPS (1) & DPS (2) & DPS (3) \\
    \end{tabular}
    }
    \vspace{-0.4cm}
\end{wrapfigure}

Fig.~\ref{fig:identity_preservation_main} provides a clear qualitative example of this phenomenon. Although our RFJS reconstruction is more faithful to the ground truth in terms of identity preservation, all three BlindDPS reconstructions achieve better PSNR, SSIM, and LPIPS values. This illustrates that classical metrics can fail to capture meaningful semantic improvements, even when perceptual quality is clearly superior. We provide additional examples and quantitative evidence in Appendix~\ref{app:importance_of_target_metric}, demonstrating large perceptual gains are not reflected in PSNR, SSIM, or LPIPS.

\begin{table}[t]
\centering
\footnotesize
\setlength{\tabcolsep}{5pt}
\resizebox{0.72\textwidth}{!}{%
\rev{%
\begin{tabular}{l cc cc cc}
\toprule
 & \multicolumn{2}{c}{Box Inpainting} & \multicolumn{2}{c}{Super Resolution $\times 16$} & \multicolumn{2}{c}{Motion Deblur $K{=}256$} \\
\cmidrule(lr){2-3}\cmidrule(lr){4-5}\cmidrule(lr){6-7}
Metric & DPS & RFJS (ours) & DPS & RFJS (ours) & DPS & RFJS (ours) \\
\midrule
DreamSim ($\downarrow$) & $0.078$  & $\mathbf{0.054}\;{\scriptstyle (+30.8\%)}$  & $0.178$  & $\mathbf{0.127}\;{\scriptstyle (+29.0\%)}$  & $0.120$  & $\mathbf{0.088}\;{\scriptstyle (+27.2\%)}$ \\
CLIP ($\downarrow$)     & $0.088$  & $\mathbf{0.061}\;{\scriptstyle (+30.1\%)}$  & $0.202$  & $\mathbf{0.136}\;{\scriptstyle (+32.7\%)}$  & $0.123$  & $\mathbf{0.085}\;{\scriptstyle (+30.4\%)}$ \\
NME ($\downarrow$)      & $0.0275$ & $\mathbf{0.0248}\;{\scriptstyle (+9.7\%)}$  & $0.0331$ & $\mathbf{0.0318}\;{\scriptstyle (+4.1\%)}$  & $0.0292$ & $\mathbf{0.0271}\;{\scriptstyle (+7.0\%)}$ \\
\midrule
FS ($\downarrow$)       & $0.740$  & $\mathbf{0.298}\;{\scriptstyle (+59.8\%)}$  & $1.310$  & $\mathbf{0.481}\;{\scriptstyle (+63.3\%)}$  & $1.212$  & $\mathbf{0.439}\;{\scriptstyle (+63.8\%)}$ \\
PSNR ($\uparrow$)       & $27.21$  & $\mathbf{27.78}$  & $20.98$  & $\mathbf{21.04}$  & $22.18$  & $\mathbf{22.20}$ \\
LPIPS ($\downarrow$)    & $0.141$  & $\mathbf{0.137}$  & $0.296$  & $\mathbf{0.293}$  & $0.263$  & $\mathbf{0.261}$ \\
SSIM ($\uparrow$)       & $0.846$  & $\mathbf{0.852}$  & $0.563$  & $\mathbf{0.567}$  & $0.604$  & $\mathbf{0.607}$ \\
\bottomrule
\end{tabular}}%
}
\caption{\rev{Search with side information improves other perceptual metrics which are not directly derived from the reward modeling network. The last two
tasks use the severe-degradation setting. See \cref{tab:perceptual_metrics} for the full set of tasks.}}
\label{tab:perceptual_metrics_main}
\end{table}

\subsection{Additional Experiments (Appendix)}
The results of our framework for other samplers like MPGD, along with additional DPS results, are provided in Appendix~\ref{app:other_samplers}. We further include ablation/sensitivity analyses that address \textbf{compute, runtime, robustness, and sensitivity}: $(i)$ \textbf{Compute--performance trade-off:} effect of the \textbf{number of particles} (\cref{app:num_particles_effect}).
$(ii)$  \textbf{Runtime:} wall-clock comparison across methods/settings (\cref{app:runtimes}).
$(iii)$ \textbf{Robustness:} degradation of \textbf{measurement quality} (harder inverse problems) (\cref{app: measurement_quality}) and \textbf{side-information quality} (\cref{app: side_info_quality}).
$(iv)$ \textbf{Sensitivity of Gradient Guidance:} motivating our search-based alternative (\cref{app:effect_of_grad_scale}).
$(v)$ \textbf{Toy examples:} illustrating the benefits of side information (\cref{app:simulation}).

\vspace{-0.2cm}
\section{Conclusion}
\vspace{-0.2cm}

We proposed a lightweight, modular inference-time search algorithm that integrates side information into diffusion-based image reconstruction, in a principled way. By adaptively guiding the generative process, our method delivers substantial quality gains, especially in ill-posed settings, while requiring only minimal changes to existing pipelines. Extensive experiments across standard reconstruction tasks show consistent improvements in both visual fidelity and quantitative metrics, and our approach surpasses gradient-based alternatives. These results highlight the power of leveraging side information at inference time to make diffusion-based solvers more reliable and accurate.

\begin{ack}
\rev{We thank the department of Electrical and Computer Engineering (ECEN) at Texas A\&M
University for providing access to the Olympus computing cluster. Portions of this research were
conducted with the advanced computing resources provided by Texas A\&M High Performance Research
Computing. This work was supported, in part, by the U.S. Army Combat Capabilities Development
Command (DEVCOM) under Grant Number W911NF2520046. This work was also supported in part by the
National Science Foundation grant NSF-CNS 2148354, federal agencies, and industry partners as
specified in the Resilient \& Intelligent NextG Systems (RINGS) program. Any opinions, findings,
and conclusions or recommendations expressed in this material are those of the authors and do not
necessarily reflect the views of the sponsoring agencies.}

\rev{The authors declare no competing interests in relation to this work.}
\end{ack}

\bibliography{contents/ref}
\bibliographystyle{plainnat}

%%%%%%%%%%%%%%%%%%%%%%%%%%%%%%%%%%%%%%%%%%%%%%%%%%%%%%%%%%%%%%%%%%%%%%%%%%%%%%%
% APPENDIX
%%%%%%%%%%%%%%%%%%%%%%%%%%%%%%%%%%%%%%%%%%%%%%%%%%%%%%%%%%%%%%%%%%%%%%%%%%%%%%%
\newpage
\appendix

\section*{Appendices}

\section{Additional Related Work}
\label{app: additional_related_work}

\textbf{Reward-gradient guidance:} LGD \citep{song2023lossguideddmpnp} refines DPS via Monte Carlo estimates, while UGD \citep{bansal2024universal}, FreeDoM \citep{yu2023freedom}, and RB-Modulation \citep{rout2025rbmodulation} propose to guide the diffusion with a gradient of the reward function. \rev{More recently, ABMS \citep{ren2026abms} reduces the estimation error of this gradient by inserting an additional backward denoising step and averaging over $M$ Monte Carlo samples of $\vx_0$ before differentiating. It maintains no particle weights and performs no resampling, and is therefore best understood as a lower-bias variant of the RGG baseline we compare against (\cref{sec:experiment_setup}) rather than as a particle-based method.} In addition to being gradient-based approaches, they are typically used for semantic generation tasks rather than inverse problems\rev{, and they all require the reward to be differentiable}.

\textbf{SMC methods:} Sequential Monte Carlo approaches \citep{cardoso2024monte, dou2024fpssmc, wu2023practicalexactsampling} generate and resample particles under tilted distributions, offering gradient-free alternatives. DAS \citep{kim2025testtime} combines resampling with gradients for text-to-image tasks. These methods rely only on the measurement to guide the unconditional sampler and do not exploit side information. \rev{LD-SMC \citep{achituve2025ldsmc} extends this construction to the latent space of a diffusion model.}

\rev{An SMC method is specified by two choices: how particles are weighted, and how they are resampled. The methods above weight particles by the measurement $\vy$ and resample over the full particle set. Our algorithms differ in both respects. The weights come from the side information, through the reward $r(\vx_0; \vs)$, which carries information that $\vy$ does not when the inverse problem is severely ill-posed. Resampling is local: particles in RFJS are resampled within dynamically formed groups rather than globally. Greedy Search is the special case of a single global group, and is in this sense an SMC algorithm with side-information weights, while RFJS generalizes it along the resampling axis. \Cref{app:resampling_strategies} reports comparisons with other weighting and resampling rules, namely adaptive and annealed SMC, stratified and multinomial resampling, adaptive group sizes and reward-guided beam search, together with stochastic resampling at a positive temperature. \Cref{prop:seq-diff-sampling} holds for every $\tau > 0$, and the deterministic selection used in our main experiments is its $\tau \to 0$ limit.}

\rev{\textbf{Reference-based restoration with trained conditional models:} A complementary line of work trains the generative model itself to consume the side information. ReF-LDM \citep{hsiao2024refldm} is a latent diffusion model for reference-based face restoration: it conditions on one or several reference images of the same identity through a cached key--value mechanism and is trained with a timestep-scaled identity loss on a paired dataset of degraded images and references. Because the references are available during training, such a model can aggregate several of them jointly, and restoration quality improves as more references are supplied \citep{hsiao2024refldm}. What this requires is a paired dataset of targets and side information, a training run, and a fixed choice of degradation process and side-information modality. \Cref{app:additional_baselines} compares the two approaches on super-resolution and on motion deblurring. Multiple references enter our framework at the level of the reward, by aggregating per-reference rewards (\cref{app: side_info_quality}).}

\rev{\textbf{Text-conditioned solvers and training-free wrappers:} TReg \citep{kim2025regularization} uses a textual description as side information by regularizing the reverse diffusion of a text-conditioned latent model through that model's own conditioning interface. Since our search runs on top of an arbitrary sampler, TReg can also serve as a base solver for it, and we report this combination in \cref{app:additional_baselines}. ReGuidance \citep{karan2025reguidance} is a training-free wrapper that acts on the initialization rather than the guidance: it inverts a candidate reconstruction by running the unconditional probability-flow ODE in reverse and uses the resulting latent to initialize DPS, which improves realism and measurement consistency on hard inverse problems. Like our framework it reuses an existing solver as a black box, but it does not use side information.}

\textbf{Guidance and side information in diffusion models:} Several recent works have explored conditioning, guidance, and optimization strategies for diffusion models, but they differ substantially in whether and how \emph{side information} is incorporated.

CLAY \citep{zhang2024clay} is a conditional generative model for 3D asset synthesis that incorporates text, image views, and simple 3D cues via trained, modality-specific conditioning modules (e.g., cross-attention adapters) within a native 3D diffusion model. While this enables controllable generation, the conditioning signals are integrated through learned modules during training, and the method is not formulated as an inverse problem under a measurement model.

The term side information is also used by \citet{fei2023generative}, but in a different sense: their setting considers multiple measurements of the same underlying ground truth, where the relationship between measurements and the clean image is either known or can be estimated parametrically. As a result, the additional information can be incorporated through an explicit forward-model-based formulation. In contrast, we consider side information that may come from heterogeneous modalities (e.g., text, reference images, alternative MRI contrasts) with unknown and nonlinear relationships to the target image. Since such relationships are not expressible as forward models, we incorporate side information implicitly via a reward that guides inference-time search, without retraining.

Other inference-time optimization approaches for diffusion models do not incorporate side information. For example, \citet{chung2024prompttuningforldms} optimizes latent variables and text-prompt embeddings during inference to better exploit the learned prior, improving alignment with the prompt but relying solely on internal model variables. Similarly, \citet{janati2024divideandconquer} uses distribution tilting aligned with the measurement via Langevin dynamics, \emph{without incorporating external side information}, and does not perform inference-time search.

Finally, \citet{pandey2025variational} proposes a variational framework that optimizes a reward defined with respect to the measurement likelihood and assumes differentiable reward functions. By contrast, we model the relationship between the target image and side information using a reward-tilted posterior and support non-differentiable rewards, enabling flexible integration of arbitrary side information at inference time.

\section{Proofs}

\subsection{Proof of Proposition~\ref{prop:seq-diff-sampling}}
\label{proof:seq-diff-sampling}

To begin, recall that $p_{0 \mid S}(\vx_0 | \vs) = \frac{1}{Z} p_0(\vx_0) e^{\frac{r(\vx_0; \vs)}{\tau}}$.
Using Bayes' rule, we can rewrite this expression as
\begin{align*}
\frac{1}{Z} p_0(\vx_0) \exp\left(\frac{r(\vx_0; \vs)}{\tau}\right)
&= \frac{p_{0, S}(\vx_0, \vs)}{p_{S} (\vs)}
= \frac{p_{S \mid 0}(\vs \mid \vx_0) p_0(\vx_0)}{{p_{S} (\vs)}}
\end{align*}
and we gather that, for $\vs$ fixed, $p_{S \mid 0}(\vs | \vx_0) \propto e^{\frac{r(\vx_0; \vs)}{\tau}}$ where the proportionality is up to constants on independent of $\vx_0$.
Starting from the LHS of \eqref{eq:seq-diff-sampling}, we first apply Bayes' rule, reverse the conditioning, and introduce a marginalized $\vx_0$ to leverage conditional independence.
This sequence leads to
\begin{align*}
&p_{t \mid t+1, Y, S}(\vx_{t} \mid \vx_{t+1}, \vy, \vs)
= \frac{ p_{t, S \mid t+1, Y}(\vx_{t}, \vs \mid \vx_{t+1}, \vy) }{ p_{S \mid t+1, Y}(\vs \mid \vx_{t+1}, \vy ) } \\
&= \frac{ p_{t \mid t+1, Y}(\vx_t \mid \vx_{t+1}, \vy) p_{S \mid t, t+1, Y}(\vs \mid \vx_{t}, \vx_{t+1}, \vy) }
{ p_{S \mid t+1, Y}(\vs \mid \vx_{t+1}, \vy ) } \\
&\propto p_{t \mid t+1, Y}(\vx_t \mid \vx_{t+1}, \vy)
\int_{\vx_0} p_{S \mid t, t+1, Y}(\vs \mid \vx_0, \vx_{t}, \vx_{t+1}, \vy) p_{0 \mid t, t+1, Y}(\vx_0 \mid \vx_{t}, \vx_{t+1}, \vy) d\vx_0 \\
&= p_{t \mid t+1, Y}(\vx_t \mid \vx_{t+1}, \vy)
\int_{\vx_0} p_{S \mid 0, t, t+1, Y}(\vs \mid \vx_0, \vx_{t}, \vx_{t+1}, \vy) p_{0 \mid t, Y}(\vx_0 \mid \vx_{t}, \vy) d\vx_0 \\
&\propto p_{t \mid t+1, Y}(\vx_t \mid \vx_{t+1}, \vy)
\int_{\vx_0} p_{S \mid 0}(\vs \mid \vx_0) p_{0 \mid t, Y}(\vx_0 \mid \vx_{t}, \vy) d\vx_0 \\
&= p_{t \mid t+1, Y}(\vx_t \mid \vx_{t+1}, \vy) \bE_{\vx_0 \sim p_{0\mid t, Y}(\cdot \mid \vx_t, \vy)}[\exp(r(\vx_0; \vs)/\tau)] \\
&\propto p_{t \mid t+1, Y}(\vx_t \mid \vx_{t+1}, \vy) \exp(V_t^\tau(\vx_t; \vs, \vy)) .
\end{align*}
The penultimate step follows from the discussion at the onset of the proof.
The last step captures the definition for $V_t^\tau(\vx_t; \vs, \vy)$ found in Proposition~\ref{prop:seq-diff-sampling}. The proof of \eqref{eq:marginal-dist} is similar without conditioning on $t+1$.

\textit{Proof of Value-titled KL}. Given a distribution $p_0$ over $\bR^d$, a reward function $r: \bR^d \to \bR$, and $\tau > 0$, we are interested in sampling from the distribution $p^*$ given by
\begin{align*}
    p^* &= \arg\max_{p} \bE_{\vx\sim p}[r(\vx)]-\tau D_{\rm KL}(p \Vert p_0) \\
    &= \arg\max_{p} \bE_{\vx\sim p}\left[r(\vx)-\tau \log \frac{p(\vx)}{p_0(\vx)}\right] \\
    &= \arg\min_{p} \bE_{\vx\sim p}\left[\log \frac{p(\vx)}{p_0(x)}-\frac{r(\vx)}{\tau}\right] \\
    &= \arg\min_{p} \bE_{x\sim p}\left[\log \frac{p(\vx)}{p_0(\vx)e^{r(\vx)/\tau}}\right] \triangleq \arg\min_p \cL(p).
\end{align*}
Let $q(\vx) \triangleq \frac{1}{Z} p_0(\vx)e^{r(\vx)/\tau}$, where $Z$ is chosen such that $\int q(\vx) d\vx = 1$. Then
\begin{align*}
    \cL(p) = \bE_{\vx\sim p}\left[\log \frac{p(\vx)}{p_0(\vx)e^{r(\vx)/\tau}}\right] = \bE_{\vx\sim p}\left[\log \frac{p(\vx)}{Zq(\vx)}\right] = D_{\rm KL}(p \Vert q) -\log Z.
\end{align*}
By the non-negativity of KL-divergence, $\cL(p) \ge \cL(q)$ for any distribution $p$, and so $p^* = q$, or $p(\vx) \propto p_0(\vx) e^{r(\vx)/\tau}$.

\subsection{\rev{Exactness Conditions for the Reward-Tilted Target}}
\label{app:exactness}

\rev{\Cref{eq:exactness} records what would have to be true of the reward for the tilted
distribution of \cref{eq:proxy-posterior} to be the conditional law of $\vx_0$ given $\vs$: the
reward must be an affine function of $\log p_{S \mid 0}(\vs \mid \vx_0)$ with slope $1/\tau$, up to
a term that does not depend on $\vx_0$. Each of the two constructions below pairs a reward used in
this paper with a particular temperature rather than leaving $\tau$ free.}

\rev{Both rewards used in this paper are obtained by training a network, and in each case the
training objective determines, at its population optimum, a quantity from which
$\log p_{S \mid 0}(\vs \mid \vx_0)$ can be recovered \emph{exactly}. The two statements below make
this precise. Neither requires a model of how $\vs$ was generated; both require only that the
network has the capacity to attain its own optimum.}

\paragraph{\rev{Face identity.}}\rev{ Here $\vs$ is a photograph of the same person as $\vx_0$. Let
$y \in \{1, \dots, M\}$ be the identity, and let $\pi(\vx) \in \Delta^{M-1}$ denote the classifier's
softmax, $\pi_j(\vx) = \mathrm{softmax}_j \big( \kappa \langle \Phi(\vx), \vw_j \rangle \big)$, which is
what AdaFace \citep{kim2022adaface} is trained to produce: a normalised softmax over identities with
$\ell_2$-normalised embeddings $\Phi$ and class weights $\vw_j$, at logit scale $\kappa$. Assume}
\begin{itemize}[leftmargin=2em, itemsep=0.1em]
  \item \rev{\textbf{(A1)} \emph{capacity}: the classifier attains the population minimum of its
    cross-entropy objective, so $\pi_j(\vx) = \Pr(y = j \mid \vx)$ for almost every $\vx$;}
  \item \rev{\textbf{(A2)} \emph{conditional independence}: $\vs \perp \vx_0 \mid y$, two
    photographs of one person being independent given the identity;}
  \item \rev{\textbf{(A3)} \emph{uniform identity prior}: $\Pr(y = j) = 1/M$.}
\end{itemize}

\rev{\textbf{Proposition.} Under \textup{(A1)--(A3)}, for every $\vx_0$ and $\vs$,}
\begin{align}
\label{eq:adaface-exact}
\rev{\log \frac{p_{S \mid 0}(\vs \mid \vx_0)}{p_S(\vs)} = \log M + \log \big\langle \pi(\vx_0), \pi(\vs) \big\rangle .}
\end{align}
\begin{proof}
\rev{Marginalising the identity and using (A2),
$p_{S \mid 0}(\vs \mid \vx_0) = \sum_{j} p(\vs \mid y = j) \Pr(y = j \mid \vx_0)$. Bayes' rule gives
$p(\vs \mid y = j) = \Pr(y = j \mid \vs) \, p_S(\vs) / \Pr(y = j)$, and (A3) makes the denominator
$1/M$, so $p_{S \mid 0}(\vs \mid \vx_0) = M \, p_S(\vs) \sum_j \Pr(y = j \mid \vs) \Pr(y = j \mid \vx_0)$.
By (A1) the two posteriors are $\pi(\vs)$ and $\pi(\vx_0)$. Divide by $p_S(\vs)$ and take logarithms.}
\end{proof}

\rev{The right-hand side is computable from the embeddings alone,}
\begin{align*}
\rev{\big\langle \pi(\vx_0), \pi(\vs) \big\rangle = \sum_{j=1}^{M}
\mathrm{softmax}_j \big( \kappa \langle \Phi(\vx_0), \vw_j \rangle \big) \,
\mathrm{softmax}_j \big( \kappa \langle \Phi(\vs), \vw_j \rangle \big) ,}
\end{align*}
\rev{while $\log M$ and $\log p_S(\vs)$ do not depend on $\vx_0$ and are therefore absorbed into
$c(\vs)$. The quantity against which \cref{eq:exactness} measures a reward is thus available for
any pair $(\vx_0, \vs)$, which is what turns that condition from a statement about an unknown
density into one that can be checked.}

\rev{AdaFace is trained at $\kappa = 64$, a value suited to its own objective of deciding among
the $M$ identities: at that scale the softmax is close to one-hot, so
$\langle \pi(\vx_0), \pi(\vs) \rangle$ is close to the indicator that $\vx_0$ and $\vs$ select the
same identity. Our objective asks for the opposite. A reward must separate candidate
reconstructions that already share an identity, which a binary verdict cannot do, so we evaluate
the same expression at the smaller $\kappa = 8$. This keeps the form of \cref{eq:adaface-exact} and
yields the soft, graded tilt the search needs.}

\rev{The
reward that we use based on the embedding distance, $r = -\lVert \Phi(\vx_0) - \Phi(\vs) \rVert_2$, varies continuously with
$\langle \Phi(\vx_0), \Phi(\vs) \rangle$ and has the same soft distinction while being much simpler to interpret. Besides it has no scale to choose and no
dependence on which $M$ identities the network was trained to separate.}

\paragraph{\rev{Textual descriptions.}}\rev{ Here $\vs = T$ is a description of $\vx_0$ and the
reward is the score $f(T, \vx_0)$ of ImageReward \citep{xu2023imagereward}, trained with the
pairwise logistic (Bradley--Terry) loss \citep{bradley1952paired}
$-\log \sigma \big( f(T, \vx^1) - f(T, \vx^2) \big)$ over annotator-preferred pairs
$\vx^1 \succ \vx^2$. Assume}
\begin{itemize}[leftmargin=2em, itemsep=0.1em]
  \item \rev{\textbf{(B1)} \emph{capacity}: $f$ attains the population minimum of that loss;}
  \item \rev{\textbf{(B2)} \emph{annotators compare by likelihood}:
    $\Pr(\vx^1 \succ \vx^2 \mid T) = p_{S \mid 0}(T \mid \vx^1) / \big( p_{S \mid 0}(T \mid \vx^1) + p_{S \mid 0}(T \mid \vx^2) \big)$.}
\end{itemize}

\rev{\textbf{Proposition.} Under \textup{(B1)--(B2)} there is a function $a$ of the prompt alone with}
\begin{align}
\label{eq:imagereward-exact}
\rev{f(T, \vx_0) = \log p_{S \mid 0}(T \mid \vx_0) + a(T),}
\end{align}
\rev{so \cref{eq:exactness} holds exactly for $r = f$ at $\tau = 1$ and $c(\vs) = a(T)$.}

\begin{proof}
\rev{By (B1) the optimum satisfies
$\sigma \big( f(T, \vx^1) - f(T, \vx^2) \big) = \Pr(\vx^1 \succ \vx^2 \mid T)$ for every pair, and by
(B2) the right-hand side equals
$\sigma \big( \log p_{S \mid 0}(T \mid \vx^1) - \log p_{S \mid 0}(T \mid \vx^2) \big)$. As $\sigma$ is
injective, $f(T, \cdot) - \log p_{S \mid 0}(T \mid \cdot)$ takes the same value at $\vx^1$ and
$\vx^2$ for every pair, hence is a function of $T$ alone.}
\end{proof}

\rev{Note that the logistic link has unit slope, so (B1) identifies the \emph{scale} of $f$ and not
only its ordering; the single remaining degree of freedom is the additive $a(T)$, which
\cref{eq:exactness} permits.}

\rev{Both rewards used in this paper therefore satisfy \cref{eq:exactness} approximately. For
textual side information it holds directly, since \cref{eq:imagereward-exact} makes the
ImageReward score itself the required affine function at $\tau = 1$. For face identity the reward
we run is the embedding distance rather than $\log \langle \pi(\vx_0), \pi(\vs) \rangle$, and
\cref{app:reward_quality} settles the difference between the two by fitting one against the other,
finding the embedding distance to be an approximately affine image of the log-likelihood ratio
that \cref{eq:adaface-exact} supplies.}

\paragraph{\rev{Ordering in the small-$\tau$ limit.}}\rev{ The experiments reported in this paper use the small-$\tau$ setting described in \cref{subsec:rkj-search}, in which resampling retains the highest-reward particle in each group. In that regime the weights $w[i] \propto \exp(r[i]/\tau)$ enter only through which of them is largest, so any strictly increasing transformation of $r$ traces out the same trajectory while defining a different distribution through \cref{eq:proxy-posterior}. Exactness at a particular $\tau$ is consequently not what the reported numbers rest on; what they require is that the reward \emph{order} candidates as the posterior would, which is a weaker condition and the one measured in \cref{app:reward_quality}. The conditions above matter for the positive-$\tau$ sampler, where the magnitudes of the weights, and not only their order, affect the outcome.}

\subsection{Value Approximation Bound}
\label{proof:approx-error-bound}

In the following, we provide the steps that lead to Eq.~\ref{eq:approx-value-function}, and subsequently bound the approximation error. We begin with the following lemma.

\begin{lemma}
    \label{lem:conditional-mean}
    The conditional mean of $X_0$ given $X_t = \vx_t$ and $Y = \vy$ is given by
    \begin{align}
    \label{eq:exact-conditional-mean-xy}
        \hat{\vx}_{0\mid t, Y}(\vx_t, \vy) = \hat{\vx}_{0\mid t}(\vx_t) + \left(\frac{1-\alpha_t}{\sqrt{\alpha_t}} \right) \nabla_{\vx_t} \log p_{Y\mid t}(\vy \mid \vx_t).
    \end{align}
\end{lemma}

\textit{Proof.} For any distribution over $X_0$ since $p_{t\mid 0}(\vx_t \mid \vx_0) = \cN( \vx_t \mid \sqrt{\alpha_t} \vx_0, (1-\alpha_t) \mI)$, we can use the Tweedie's formula \citep{efron2011tweedie} to $p_{0\mid Y}(\vx_0 \mid \vy)$ and $p_{0}(\vx_0)$ to get
\begin{align*}
    \sqrt{\alpha_t} \hat{\vx}_{0\mid t, Y}(\vx_t, \vy) &= \vx_t + (1-\alpha_t) \nabla_{\vx_t} \log p_{t \mid Y}(\vx_t \mid \vy) \\
    \sqrt{\alpha_t} \hat{\vx}_{0\mid t}(\vx_t) &= \vx_t + (1-\alpha_t)\nabla_{\vx_t} \log p_{t}(\vx_t).
\end{align*}

Since by Bayes theorem, $ \nabla_{\vx_t} \log p_{t \mid Y}(\vx_t \mid \vy) = \nabla_{\vx_t} \log p_{t}(\vx_t) + \log p_{Y \mid t}(\vy \mid \vx_t)$, the results follows by simple algebra.

Since $p_{Y\mid 0}$ is Gaussian, using the DPS approximation on the second term in Eq.~\eqref{eq:exact-conditional-mean-xy}, we get
\begin{align*}
    \hat{\vx}_{0\mid t, Y}(\vx_t, \vy) \approx \hat{\vx}_{0\mid t}(\vx_t) - \left(\frac{1-\alpha_t}{\sqrt{\alpha_t}} \right) \frac{1}{2\sigma_y^2}\nabla_{\vx_t} \norm{\vy - \mA(\hat{\vx}_{0\mid t}(\vx_t))}_2^2.
\end{align*}
Replacing $1/2\sigma_y^2$ by $\eta$ to control the approximation error gives Eq.~\eqref{eq:dps-approximation-for-conditional-mean}. In the following, we denote this approximation as
\begin{align*}
    \hat{\vx}_{0\mid t, Y}(\vx_t, \vy) \approx \hat{\vx}_{0\mid t}(\vx_t) - \left(\frac{1-\alpha_t}{\sqrt{\alpha_t}} \right) \eta \nabla_{\vx_t} \norm{\vy - \mA(\hat{\vx}_{0\mid t}(\vx_t))}_2^2\triangleq \tilde{\vx}_{0\mid t}^\eta(\vx_t, \vy).
\end{align*}

\begin{proposition}
\label{prop:approx-error-bound}
Assume that $r$ is a Lipschitz function that takes values in $[0, 1]$. For any $\vx_t, \vy, \vs$, the error in the value approximation $\hat{V}_t^\tau(\vx_t; \vs, \vy) = r(\tilde{\vx}_{0\mid t}^\eta(\vx_t, \vy); \vs)/\tau$ with the true value $V_t^\tau(\vx_t; \vs, \vy)$ is bounded as
    \begin{align*}
        \abs{V_t^\tau(\vx_t; \vs, \vy)- \hat{V}_t^\tau(\vx_t; \vs, \vy)} \le \rev{\min \left\{ c_\tau c_1(t), \frac{M_r(t)}{\tau} \right\}} + \frac{L_r}{\tau} \left( \sqrt{c_2(t)} + \sqrt{c_3(t) c_4^\eta(t)} \right),
    \end{align*}
where $c_1(t) = \mathrm{Var}\left(r(X_0; \vs) \mid \vx_t, \vy \right)$, $c_2(t) = \mathrm{Var}(X_0 \mid \vx_t, \vy)$, $c_3(t) = \mathrm{Var}(X_0 \mid \vx_t)$,
\begin{align*}
    c_4^\eta(t) &=  1 + \mathrm{CV}^2(t) + \eta^2 \norm{\mA^T(\vy-\mA\hat{\vx}_{0\mid t}(\vx_t))}_{\Sigma_{0\mid t}(\vx_t)}^2 \\
    &\hspace{40pt} - 2 \eta\langle \mA(\hat{\vx}_{0\mid t, Y}(\vx_t, \vy) - \hat{\vx}_{0\mid t}(\vx_t)), \vy-\mA\hat{\vx}_{0\mid t}(\vx_t) \rangle,
\end{align*}
where $\mathrm{CV}(t) \triangleq \frac{\sqrt{\mathrm{Var}(p_{Y\mid 0}(\vy \mid X_0) \mid \vx_t)}}{\bE[p_{Y\mid 0}(\vy \mid X_0) \mid \vx_t]}$ is the coefficient of variation of the likelihood function $p_{Y\mid 0}(\vy \mid X_0)$ given $\vx_t$ and $c_\tau = e^{1/\tau} - 1 - 1/\tau$ is a positive constant. \rev{The quantity}
\begin{align*}
    \rev{M_r(t) \triangleq \operatorname*{ess\,sup}_{p_{0 \mid t, Y}(\cdot \mid \vx_t, \vy)} r(X_0; \vs) - \bE[r(X_0; \vs) \mid \vx_t, \vy] \in [0, 1]}
\end{align*}
\rev{is the gap between the largest attainable reward and the mean reward under $p_{0\mid t, Y}(\cdot \mid \vx_t, \vy)$.}
\end{proposition}

\begin{remark}
In Proposition~\ref{prop:approx-error-bound}, the term $c_1(t)$ denotes the conditional variance of the reward given $\vx_t, \vy$, $c_2(t)$ denotes the conditional variance of $X_0$ given $\vx_t, \vy$, and $c_3(t)$ denotes the conditional variance of $X_0$ given only $\vx_t$. Since the variance of the reverse distribution $p_{0\mid t}(\cdot \mid \vx_t)$ decreases as $t$ becomes smaller, we have that all the terms $c_1(t), c_2(t), c_3(t)$ are small when $t$ is small. Therefore, the approximation error is small when $t$ is small.
\end{remark}

\begin{remark}
\rev{The three constants are nested, so the whole bound is controlled by $c_3(t) = \mathrm{Var}(X_0 \mid \vx_t)$ alone. Since a variance is the smallest mean squared deviation from any constant and $r$ is $L_r$-Lipschitz,}
\begin{align*}
    \rev{c_1(t) = \mathrm{Var}(r(X_0; \vs) \mid \vx_t, \vy) \le \bE \big[ \big( r(X_0; \vs) - r(\hat{\vx}_{0 \mid t, Y}(\vx_t, \vy); \vs) \big)^2 \,\big|\, \vx_t, \vy \big] \le L_r^2 \, c_2(t) .}
\end{align*}
\rev{For the second link, write $\kappa_y \triangleq \sup_{\vx_0} p_{Y\mid 0}(\vy \mid \vx_0) / p_{Y \mid t}(\vy \mid \vx_t)$, which is the largest value of the density ratio $f$ used in the proof above. Changing measure with $f$ gives}
\begin{align*}
    \rev{c_2(t) \le \bE_{p_{0\mid t, Y}}[\norm{X_0 - \hat{\vx}_{0\mid t}(\vx_t)}_2^2] = \bE_{p_{0 \mid t}}[f(X_0) \norm{X_0 - \hat{\vx}_{0\mid t}(\vx_t)}_2^2] \le \kappa_y \, c_3(t) .}
\end{align*}
\rev{Hence $c_1(t) \le L_r^2 \kappa_y c_3(t)$ and $c_2(t) \le \kappa_y c_3(t)$, and the bound of \cref{prop:approx-error-bound} becomes}
\begin{align*}
    \rev{\abs{V_t^\tau - \hat{V}_t^\tau} \le \frac{L_r}{\tau} \sqrt{c_3(t)} \left( \sqrt{\kappa_y} + \sqrt{c_4^\eta(t)} \right) + \min \left\{ c_\tau L_r^2 \kappa_y \, c_3(t), \, \frac{M_r(t)}{\tau} \right\} ,}
\end{align*}
\rev{so at fixed $\tau$ the leading behaviour at small $t$ is $O(\sqrt{c_3(t)}/\tau)$, the remaining term being $O(c_3(t))$. The reward is evaluated on $\hat{\vx}_{0\mid t}$, so this is the regime the bound describes.}
\end{remark}

\begin{remark}
\rev{The quantity that governs the bound is moreover computable from the pre-trained denoiser, without drawing samples from $p_{0\mid t}$. The identity established in the proof above,  $\big( \frac{1-\alpha_t}{\sqrt{\alpha_t}} \big) \nabla_{\vx_t} \hat{\vx}_{0\mid t}(\vx_t) = \Sigma_{0\mid t}(\vx_t)$, gives}
\begin{align*}
    \rev{c_3(t) = \mathrm{tr} \, \Sigma_{0 \mid t}(\vx_t) = \left( \frac{1-\alpha_t}{\sqrt{\alpha_t}} \right) \mathrm{tr} \big( \nabla_{\vx_t} \hat{\vx}_{0 \mid t}(\vx_t) \big) .}
\end{align*}
\rev{The trace on the right can be estimated with Hutchinson's identity \citep{hutchinson1989trace}, $\mathrm{tr}(\mJ) = \bE_{\vv}[\vv^T \mJ \vv]$ for $\vv$ with independent Rademacher entries, and each sample $\mJ \vv$ is a single Jacobian-vector product through the denoiser network. A handful of such products therefore evaluates $c_3(t)$ along a trajectory, so the bound can be measured on a given run.}
\end{remark}

\textit{Proof.} Since $r$ is a bounded random variable, assuming finite variance, we can use Bennett's inequality for the log moment-generating function
\begin{align*}
    V_t^\tau(\vx_t; \vs, \vy)  = \log \bE[\exp(r(X_0; \vs) / \tau)] \le \frac{1}{\tau} \bE[r(X_0; \vs)] +  c_\tau c_1(t).
\end{align*}
\rev{The same quantity admits a second bound. Bounding the integrand by its essential supremum gives $\log \bE[\exp(r(X_0;\vs)/\tau)] \le \frac{1}{\tau} \operatorname*{ess\,sup} r(X_0; \vs)$, and therefore}
\begin{align*}
    \rev{V_t^\tau(\vx_t; \vs, \vy) \le \frac{1}{\tau} \bE[r(X_0; \vs)] + \frac{M_r(t)}{\tau} .}
\end{align*}
\rev{The statement retains the smaller of the two. With $c_\tau = e^{1/\tau} - 1 - 1/\tau$ and $M_r(t)/\tau \le 1/\tau$, the minimum is $O(1/\tau)$ as $\tau \to 0$, the same order in $\tau$ as the remaining terms.}
Then, we have
\begin{align*}
    \abs{\log \bE[\exp(r(X_0; \vs) / \tau)] - \frac{1}{\tau} r(\tilde{\vx}^\eta_{0\mid t,Y}(\vx_t, \vy))} \le \frac{1}{\tau}\abs{\bE[r(X_0; \vs)]-r(\tilde{\vx}^\eta_{0\mid t,Y}(\vx_t, \vy))} +  c_\tau c_1(t).
\end{align*}

Now, let us simplify the first term,
\begin{align*}
    \abs{\bE[r(X_0; \vs)]-&r(\tilde{\vx}^\eta_{0\mid t,Y}(\vx_t, \vy); \vs)}
    \le \bE[\abs{r(X_0; \vs)-r(\tilde{\vx}^\eta_{0\mid t,Y}(\vx_t, \vy); \vs)}] \\
    &\le L_r \bE[\norm{X_0 - \tilde{\vx}^\eta_{0\mid t,Y}(\vx_t, \vy)}_2] \\
    &\le L_r (\bE[\norm{X_0 - \hat{\vx}_{0\mid t,Y}(\vx_t, \vy)}_2] + \norm{\hat{\vx}_{0\mid t,Y}(\vx_t, \vy)- \tilde{\vx}^\eta_{0\mid t,Y}(\vx_t, \vy)}_2).
\end{align*}
The first term can be bounded by $\sqrt{c_2(t)}$ using Cauchy-Schwarz inequality in $L^2$-probability space. For the second term, first we simplify
\begin{align*}
    \tilde{\vx}^\eta_{0\mid t,Y}(\vx_t, \vy) = \hat{\vx}_{0\mid t}(\vx_t) - \left( \frac{1-\alpha_t}{\sqrt{\alpha_t}} \right)\eta \nabla_{\vx_t} \norm{\vy-\mA \hat{\vx}_{0\mid t}(\vx_t)}_2^2,
\end{align*}
and then $\nabla_{\vx_t} \norm{\vy-\mA \hat{\vx}_{0\mid t}(\vx_t)}_2^2 =  -(\nabla_{\vx_t}\hat{\vx}_{0\mid t}(\vx_t)) \mA^T(\vy - \mA\hat{\vx}_{0\mid t}(\vx_t))$. Now,

\begin{align*}
    \nabla_{\vx_t}\hat{\vx}_{0\mid t}(\vx_t) &= \nabla_{\vx_t} \int \vx_0^T p_{0\mid t}(\vx_0 \mid \vx_t) \mathrm{d}\vx_0 = \int  \nabla_{\vx_t} p_{0\mid t}(\vx_0 \mid \vx_t) \vx_0^T \mathrm{d}\vx_0 \\
    &=  \int \nabla_{\vx_t} \log p_{0\mid t}(\vx_0 \mid \vx_t) \vx_0^T p_{0\mid t}(\vx_0 \mid \vx_t)  \mathrm{d}\vx_0.
\end{align*}

Now, we shall compute
\begin{align*}
    \nabla_{\vx_t} \log p_{0\mid t}(\vx_0 \mid \vx_t) = \nabla_{\vx_t} \log p_{t\mid 0}(\vx_t \mid \vx_0) - \nabla_{\vx_t} \log p_t(\vx_t).
\end{align*}
But since $\sqrt{\alpha_t} \hat{\vx}_{0\mid t}(\vx_t) = \vx_t + (1-\alpha_t) \nabla_{\vx_t} \log p_t(\vx_t)$, and $\nabla_{\vx_t} \log p_{t\mid 0}(\vx_t \mid \vx_0) = \frac{1}{1-\alpha_t}(\sqrt{\alpha_t}\vx_0-\vx_t)$, which gives
\begin{align*}
    \left(\frac{1-\alpha_t}{\sqrt{\alpha_t}} \right) \nabla_{\vx_t} \log p_{0\mid t}(\vx_0 \mid \vx_t) = \vx_0 - \hat{\vx}_{0\mid t}(\vx_t),
\end{align*}
which gives
\begin{align*}
    \left(\frac{1-\alpha_t}{\sqrt{\alpha_t}} \right) \nabla_{\vx_t}\hat{\vx}_{0\mid t}(\vx_t) &= \bE_{X_0 \sim p_{0\mid t}(\vx_t)}[(X_0-\hat{\vx}_{0\mid t}(\vx_t)) X_0^T] \\
    &= \bE_{p_{0\mid t}(\vx_t)}[X_0X_0^T]-\bE_{p_{0\mid t}(\vx_t)}[X_0]\bE_{p_{0\mid t}(\vx_t)}[X_0^T] \\
    &= \bE_{p_{0\mid t}(\vx_t)}[(X_0-\hat{\vx}_{0\mid t}(\vx_t))(X_0-\hat{\vx}_{0\mid t}(\vx_t))^T]\triangleq \Sigma_{0\mid t}(\vx_t),
\end{align*}
which is precisely the covariance matrix of $X_0$ given $\vx_t$.

Thus, we finally get,
\begin{align*}
    \tilde{\vx}^\eta_{0\mid t, Y}(\vx_t, \vy) &= \hat{\vx}_{0\mid t}(\vx_t) + \eta \Sigma_{0\mid t}(\vx_t) \mA^T(\vy-\mA \hat{\vx}_{0\mid t}(\vx_t))
\end{align*}

Next, note that since $\bE_{p_{0\mid t}(\vx_t)}[p_{Y\mid 0}(\vy\mid X_0)] = p_{Y\mid t}(\vy \mid \vx_t)$, we can define $f(X_0) =  \frac{p_{Y\mid 0}(\vy \mid X_0)}{\bE_{p_{0\mid t}(\vx_t)}[p_{Y\mid 0}(\vy\mid X_0)]}$, whose expectation is $\bE_{p_{0\mid t}(\vx_t)}[f(X_0)] = 1$. Further, it is easy to see that $\hat{\vx}_{0\mid t, Y}(\vx_t, \vy) = \bE_{p_{0\mid t}(\vx_t)}[X_0 f(X_0)]$. Now, we are ready to bound the final term as follows
\begin{align*}
    \norm{\hat{\vx}_{0\mid t, Y}&(\vx_t, \vy)-\tilde{\vx}^\eta_{0\mid t, Y}(\vx_t, \vy)}_2
    \\
    &= \norm{\bE_{p_{0\mid t}(\vx_t)}[X_0 f(X_0)]-\hat{\vx}_{0\mid t}(\vx_t) -\eta \Sigma_{0\mid t}(\vx_t) \mA^T(\vy-\mA \hat{\vx}_{0\mid t}(\vx_t))}_2 \\
    &=\norm{\bE_{p_{0\mid t}(\vx_t)}[X_0 f(X_0)-\hat{\vx}_{0\mid t}(\vx_t)f(X_0)] \\
    &\hspace{40pt} - \eta \bE_{p_{0\mid t}(\vx_t)}[(X_0-\hat{\vx}_{0\mid t}(\vx_t))(X_0-\hat{\vx}_{0\mid t}(\vx_t))^T] \mA^T(\vy-\mA \hat{\vx}_{0\mid t}(\vx_t))}_2 \\
    &= \norm{\bE_{p_{0\mid t}(\vx_t)}[(X_0-\hat{\vx}_{0\mid t}(\vx_t))(f(X_0)-\eta(X_0-\hat{\vx}_{0\mid t}(\vx_t))^T\mA^T(\vy-\mA\hat{\vx}_{0\mid t}(\vx_t))))]}_2 \\
    &\le \bE_{p_{0\mid t}(\vx_t)}[\norm{(X_0-\hat{\vx}_{0\mid t}(\vx_t))(f(X_0)-\eta(X_0-\hat{\vx}_{0\mid t}(\vx_t))^T\mA^T(\vy-\mA\hat{\vx}_{0\mid t}(\vx_t))))}_2] \\
    &= \bE_{p_{0\mid t}(\vx_t)}[\norm{X_0-\hat{\vx}_{0\mid t}(\vx_t)}_2 \abs{f(X_0)-\eta(X_0-\hat{\vx}_{0\mid t}(\vx_t))^T\mA^T(\vy-\mA\hat{\vx}_{0\mid t}(\vx_t))}] \\
    &\le \sqrt{c_3(t)} \sqrt{c_4^\eta(t)},
\end{align*}
where the last step follows by Cauchy-Schwarz inequality in $L^2$ probability space, where
\begin{align*}
    c_3(t) \triangleq \bE_{p_{0\mid t}(\vx_t)}[\norm{X_0-\bE_{p_{0\mid t}(\vx_t)[X_0]}}_2^2] = \mathrm{Var}(X_0 \mid \vx_t).
\end{align*}
and at last, we have
\begin{align*}
    c_4^\eta(t) &\triangleq \bE_{p_{0\mid t}(\vx_t)}[(f(X_0) - \eta (X_0-\hat{\vx}_{0\mid t}(\vx_t))^T\mA^T(\vy-\mA\hat{\vx}_{0\mid t}(\vx_t)))^2] \\
    &= \bE_{p_{0\mid t}(\vx_t)}[f(X_0)^2] + \eta^2 \bE[((X_0-\hat{\vx}_{0\mid t}(\vx_t))^T\mA^T(\vy-\mA\hat{\vx}_{0\mid t}(\vx_t)))^2] \\
    & \hspace{40pt} - 2\eta \bE_{p_{0\mid t}(\vx_t)}[(X_0f(X_0)-f(X_0)\hat{\vx}_{0\mid t}(t))^T] \mA^T(\vy-\mA\hat{\vx}_{0\mid t}(\vx_t)) \\
    &= \bE_{p_{0\mid t}(\vx_t)}[f(X_0)^2] + \eta^2 \norm{\mA^T(\vy-\mA\hat{\vx}_{0\mid t}(\vx_t))}_{\Sigma_{0\mid t}(\vx_t)}^2 \\
    &\hspace{40pt} - 2 \eta\langle \mA(\hat{\vx}_{0\mid t, Y}(\vx_t, \vy) - \hat{\vx}_{0\mid t}(\vx_t)), \vy-\mA\hat{\vx}_{0\mid t}(\vx_t) \rangle.
\end{align*}
Since $\bE_{p_{0\mid t}(\vx_t)}[f(X_0)^2] = \frac{\bE_{p_{0\mid t}(\vx_t)}[(p_{Y\mid 0}(\vy \mid X_0))^2]}{(\bE_{p_{0\mid t}(\vx_t)}[p_{Y\mid 0}(\vy \mid X_0)])^2} = 1 + \mathrm{CV}^2(t)$, where $\mathrm{CV}(t) = \frac{\sqrt{\mathrm{Var}(p_{Y\mid 0}(\vy \mid X_0) \mid \vx_t)}}{\bE[p_{Y\mid 0}(\vy \mid X_0) \mid \vx_t]}$ is the coefficient of variation of the likelihood function given $\vx_t$.

\subsubsection{Discussion on \texorpdfstring{$\eta$}{eta}}
\label{app: discussion_on_eta}

\cref{prop:approx-error-bound} states that the approximation error is small when $t$ is small. In fact, under the (standard) assumptions, the error goes to $0$ irrespective of the value of $\eta$. Thus, one can set $\eta = 0$ and still obtain consistent estimate of the conditional mean $\hat{\vx}_{0\mid t, Y}(\vx_t, \vy) \approx \hat{\vx}_{0\mid t}(\vx_t)$ when $t$ is small. \rev{Recalling that $\eta$ stands in for $1/2\sigma_y^2$, the choice $\eta = 0$ is the $\sigma_y \to \infty$ limit, in which the measurement is treated as uninformative about $\vx_0$ beyond what $\vx_t$ already carries; the intuition below makes precise why this is harmless when $t$ is small.}

To get an intuitive justification from a simpler theoretical perspective, notice that the measurement $\vy = \mA \vx_0 + \sigma_y \vz$ is a noisy observation of $\vx_0$ and $\vx_t$ is a corruption of $\vx_0$. When $t$ is small, $\vx_t$ retains high-correlation with $\vx_0$ and therefore contains more information about $\vx_0$ than the noisy measurement $\vy$. In this regime, additional conditioning by $\vy$ contributes negligibly, and thus the approximation  $\hat{\vx}_{0\mid t, Y}(\vx_t, \vy) \approx \hat{\vx}_{0\mid t}(\vx_t)$ holds. Formalizing the above notion, we notice that
\begin{align*}
    \hat{\vx}_{0\mid t, Y}(\vx_t, \vy) \approx \hat{\vx}_{0\mid t}(\vx_t) - \left(\frac{1-\alpha_t}{\sqrt{\alpha_t}} \right) \eta \rev{\nabla_{\vx_t}} \norm{\vy - \mA \hat{\vx}_{0\mid t}(\vx_t)}_2^2,
\end{align*}
and $\alpha_t \to 1$ as $t \to 0$, we have that
\begin{align*}
    \hat{\vx}_{0\mid t, Y}(\vx_t, \vy) \approx \hat{\vx}_{0\mid t}(\vx_t) + o_t(1).
\end{align*}

Empirically, we compared $\eta = 1$ and $\eta = 0$ in inference-time search algorithm using DPS as the gradient in the second term is computed by DPS, and found that $\eta > 0$ does not provide additional performance gains. Since the other base samplers (MPGD, DAPS) do not provide the term involving gradient, we use $\eta = 0$ for these methods.

Finally, we remark that since the base sampler guides towards $\vy$ explicitly from the likelihood term $\nabla_{\vx_t} \log p_{t \mid Y}(\vx_t)$, the generative process is gradually guided towards $\vy$ and therefore, we eventually sample from $p_{0\mid Y, S}$ (as suggested by \cref{prop:seq-diff-sampling}) and \emph{not just} $p_{0\mid S}$.

\subsection{Algorithm}
\label{app:algorithm}

We summarize our framework in Algorithm~\ref{alg:inference_search}. The \textsc{GroupSize} step in line $5$ computes the group-size at time $t$ and can be changed to obtain various search strategies: Best-of-N, Greedy Search, and RFJ Search, and the \textsc{Resample} step in line $6$ samples the indices within the groups as described in the main paper. Line $3$ is specific to the diffusion samplers and how they implement it.

Here, we roughly explain how the entire algorithm is implemented in all the three diffusion samplers.

\textbf{DPS:}
Compute the denoised mean and the clean data estimate $\bmu_{t}[i], \hat{\vx}_{0\mid t}[i]$ from $\vx_{t+1}[i]$. Compute $\vg_t[i] = \nabla_{\vx_t}\norm{\vy - \mA \hat{\vx}_{0\mid t}(\vx_{t+1}[i])}_2^2$ and use it to update $\hat{\vx}_{0\mid t, Y}[i] \gets \hat{\vx}_{0\mid t}[i] - \frac{1-\alpha_t}{\sqrt{\alpha_t}} \eta \vg_t[i]$, and $\bmu_{t}[i] \gets \bmu_{t}[i] - \zeta \vg_t[i]$. Then, compute rewards based on $\hat{\vx}_{0\mid t, Y}[i]$ to resample promising indices. Take the reverse diffusion step on the resampled $\bmu_t[I[i]]$ to obtain $\vx_t[i]$.

\textbf{DAPS:}
Compute the clean data estimate $\hat{\vx}_{0\mid t}[i]$ from $\vx_{t+1}[i]$. Compute the rewards based on $\hat{\vx}_{0\mid t}[i]$, resample, and then take MCMC steps, starting from the resampled particles, to perform a local Langevin sampling \citep{zhang2024daps}. In the end, we obtain $\hat{\vx}_{t,Y}[i]$, from which we sample $\vx_{t}[i]$ by adding appropriate decoupled noise \citep{zhang2024daps}.

\textbf{MPGD:}
Compute the clean data estimate $\hat{\vx}_{0\mid t}[i]$ from $\vx_{t+1}[i]$. Compute gradient $\vg_t = \nabla_{\vx_{0\mid t}}\norm{\vy - \mA \hat{\vx}_{0\mid t}[i]}_2^2$. Take $\hat{\vx}_{0\mid t, Y}[i] \gets \hat{\vx}_{0\mid t}[i] - \frac{1-\alpha_t}{\sqrt{\alpha_t}} \eta \vg_t[i]$, and $\hat{\vx}_{0\mid t}[i] \gets \hat{\vx}_{0\mid t}[i] - \zeta \vg_t$. Resample indices based on rewards computed from $\hat{\vx}_{0\mid t, Y}[i]$. Then, using the particles corresponding to the sampled indices $\hat{\vx}_{0\mid t, Y}[I[i]]$, take a reverse DDIM step \citep{he2024mpgd, song2021ddim}.

\begin{algorithm}[t]
\caption{Inference-Time Search with Side Information for Inverse Problems}
\label{alg:inference_search}
\begin{algorithmic}[1]
\REQUIRE Side information $\vs$, measurement $\vy$, reward function $r$, resampling parameter $B$, number of particles $N$, temperature $\tau > 0$
\STATE Initialize $N$ particles: $\vx_T[i] \sim \mathcal{N}(0, \mI)$ for $1 \le i \le N$
\FOR{$t = T-1$ to $0$}
    \STATE Sample $\vx_{t}[i] \sim p_{t\mid t+1, Y}(\cdot \mid \vx_{t+1}[i], \vy),$ \COMMENT{Sample candidate particles}
    \STATE $r[i] \gets r(\hat{\vx}_{0 \mid t, Y}[i]; \vs)$ \COMMENT{Compute reward using side information}
    \STATE $g_t \gets \textsc{GroupSize}(N, B, t)$ \COMMENT{Compute the group size at step $t$ for resampling}
    \STATE $I \gets \textsc{Resample}(r, g_t, \tau)$  \COMMENT{Resample indices with replacement among the groups of size $g_t$}
    \STATE $\vx_{t}[i] \gets \vx_{t}[I[i]]$ \COMMENT{Retain the particles in the resampled indices}
\ENDFOR
\STATE Select $\vx^*_0$ from $\hat{\vx}_{0 \mid 0}[1:N]$ (e.g., via reward maximization)
and return $\vx^*_0$
\end{algorithmic}
\end{algorithm}

\section{Additional Experiments}
\label{app:other_samplers}

{

\subsection{Perceptual Improvements Do Not Guarantee Better PSNR/LPIPS/SSIM}

\label{app:importance_of_target_metric}

\begin{figure}[H]
    \centering
    {\scriptsize
    \begin{tabular}{cccccc}
        \includegraphics[width=0.15\linewidth]{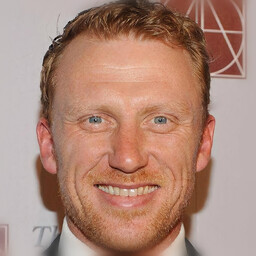} &
        \includegraphics[width=0.15\linewidth]{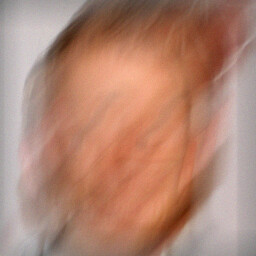} &
        \includegraphics[width=0.15\linewidth]{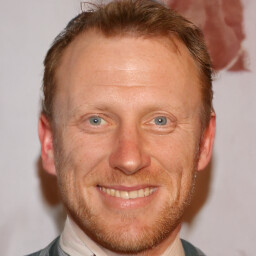} &
        \hspace{8pt} % extra spacing between groups
        \includegraphics[width=0.15\linewidth]{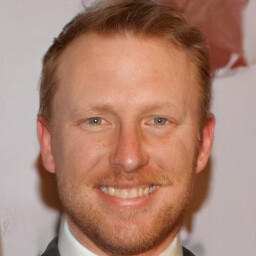} &
        \includegraphics[width=0.15\linewidth]{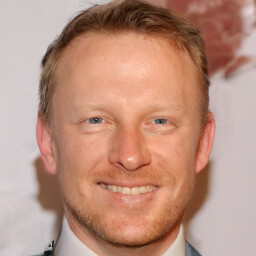} &
        \includegraphics[width=0.15\linewidth]{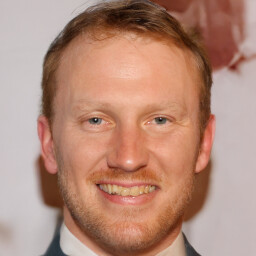} \\

        Ground truth & Input & RFJS (Ours) & DPS (1) & DPS (2) & DPS (3) \\
    \end{tabular}
    }
    \caption{Qualitative illustration of the relevance of the FS metric and the superior performance of RFJS in identity preservation. RFJS reconstruction is clearly more faithful to the ground truth, yet PSNR, SSIM, and LPIPS values slightly favor the BlindDPS outputs.}
    \label{fig:identity_preservation}
\end{figure}

Perceptual improvements do not necessarily lead to better values in classical metrics such as PSNR, LPIPS, and SSIM. Figure~\ref{fig:identity_preservation} provides a clear qualitative example: it compares three samples generated by BlindDPS with our RFJS reconstruction. Visually, our method preserves identity much better, and this is reflected in a lower FS. However, all three BlindDPS reconstructions achieve better PSNR, LPIPS, and SSIM than our result, showing that these traditional metrics can fail to capture semantic improvements.

We provide further evidence in Figure~\ref{fig:dps_hard_tasks_sr_32x}, where our reconstructions are perceptually closer to the ground truth than those of DPS, especially in challenging 32$\times$ super-resolution settings. For these examples, Table~\ref{tab:side_by_side_std_metrics} reports the standard metrics for our method and for reconstructions under DPS. Once again, the perceptual gains visible in the images are not fully reflected in PSNR, LPIPS, or SSIM. This is important because the main goal in inverse problems is to recover the underlying \emph{semantics} of the images, while PSNR, LPIPS, and SSIM are only proxies for reconstruction quality. When a metric better aligns with perceptual quality, it should be preferred. In our case, for face reconstruction this is captured by the FS metric, and for ImageNet settings by CLIPScore. In both cases, our method yields significant improvements over the baselines in these perceptual rewards.

This discrepancy is also intuitive from how these metrics are defined. PSNR is a purely pixel-level measure: it rewards putting the “right” colors in the “right” locations, even if the high-level content is not faithfully preserved. LPIPS and SSIM operate on features extracted by neural networks or on low-level structural patterns, and are more sensitive to capturing the correct class and coarse structure than to fine-grained semantics. For example, when reconstructing a human face, these metrics are not strongly incentivized to verify the correct identity; they mainly check that the output still looks like a plausible face. This explains why they may remain  in the same range even when perceptual quality clearly improves.

We emphasize that marginal differences in PSNR, SSIM, and
LPIPS between our method and the baselines should not be interpreted
as a weakness of our approach. As demonstrated above, these metrics
are fundamentally misaligned with the goal of semantic reconstruction
in ill-posed inverse problems, a conclusion supported by concurrent
work~\citep{lin2025pixelwisemetricsreliablesparseview, su2025rethinkingimageevaluationsuperresolution}. The relevant comparison
is on task-specific perceptual metrics (FS and CLIPScore),
where our method yields large and consistent improvements across all
tasks and base samplers.
\rev{We state the claim about the classical metrics in the form the evidence supports: search driven by a perceptual reward moves PSNR, SSIM and LPIPS within the same range, in either direction, rather than improving them systematically. The additional comparisons bear this out, and in both directions. Reward-guided beam search attains a higher PSNR than RFJS while trailing it on FS (\cref{tab:smc_comparison}), and ReF-LDM, a model trained for this task, is ahead on LPIPS and SSIM (\cref{tab:refldm_baseline}). \Cref{app:reward_quality} gives the particle-level reason: the reward orders candidates by identity and is close to uninformative about the distortion metrics, so there is no mechanism by which optimising it would move them in a fixed direction.}

\paragraph{\rev{Other perceptual metrics.}}\rev{ \Cref{tab:perceptual_metrics} extends
\cref{tab:perceptual_metrics_main} from three tasks to all twelve. DreamSim, the CLIP distance and
the landmark error improve on every one of them, so the FS improvement is not an artefact of
optimising an AdaFace score. They improve by less than FS does, which is what one should expect
when the reward is an identity model, and PSNR, LPIPS and SSIM improve on most tasks but not on
all of them, as this section would predict.}

\begin{table}[ht!]
\centering
\footnotesize
\setlength{\tabcolsep}{4pt}
\resizebox{\textwidth}{!}{%
\rev{%
\begin{tabular}{llccc|cccc}
\toprule
 & & \multicolumn{3}{c|}{\emph{General perceptual}} & \multicolumn{4}{c}{\emph{Task-specific and classical}} \\
Task & Algorithm & DreamSim ($\downarrow$) & CLIP ($\downarrow$) & NME ($\downarrow$) & FS ($\downarrow$) & PSNR ($\uparrow$) & LPIPS ($\downarrow$) & SSIM ($\uparrow$) \\
\midrule
\multicolumn{9}{l}{\emph{Default degradation}} \\
\midrule
\multirow{2}{*}{Box Inpainting} & DPS & $0.078$ & $0.088$ & $0.0275$ & $0.740$ & $27.21$ & $0.141$ & $0.846$ \\
 & RFJS (ours) & $\mathbf{0.054}\;{\scriptstyle (+30.8\%)}$ & $\mathbf{0.061}\;{\scriptstyle (+30.1\%)}$ & $\mathbf{0.0248}\;{\scriptstyle (+9.7\%)}$ & $\mathbf{0.298}\;{\scriptstyle (+59.8\%)}$ & $\mathbf{27.78}$ & $\mathbf{0.137}$ & $\mathbf{0.852}$ \\
\multirow{2}{*}{Super Resolution} & DPS & $0.102$ & $0.111$ & $0.0235$ & $1.054$ & $24.47$ & $0.232$ & $0.684$ \\
 & RFJS (ours) & $\mathbf{0.082}\;{\scriptstyle (+19.5\%)}$ & $\mathbf{0.090}\;{\scriptstyle (+19.0\%)}$ & $\mathbf{0.0232}\;{\scriptstyle (+1.2\%)}$ & $\mathbf{0.365}\;{\scriptstyle (+65.3\%)}$ & $\mathbf{24.57}$ & $\mathbf{0.227}$ & $\mathbf{0.686}$ \\
\multirow{2}{*}{Gaussian Deblur} & DPS & $0.066$ & $0.058$ & $0.0222$ & $0.808$ & $25.55$ & $0.202$ & $0.703$ \\
 & RFJS (ours) & $\mathbf{0.054}\;{\scriptstyle (+17.6\%)}$ & $\mathbf{0.047}\;{\scriptstyle (+18.6\%)}$ & $\mathbf{0.0212}\;{\scriptstyle (+4.7\%)}$ & $\mathbf{0.323}\;{\scriptstyle (+60.1\%)}$ & $\mathbf{25.61}$ & $\mathbf{0.198}$ & $\mathbf{0.706}$ \\
\multirow{2}{*}{Motion Deblur} & DPS & $0.066$ & $0.059$ & $0.0223$ & $0.819$ & $25.85$ & $0.196$ & $0.726$ \\
 & RFJS (ours) & $\mathbf{0.054}\;{\scriptstyle (+17.6\%)}$ & $\mathbf{0.044}\;{\scriptstyle (+24.8\%)}$ & $\mathbf{0.0214}\;{\scriptstyle (+4.0\%)}$ & $\mathbf{0.322}\;{\scriptstyle (+60.6\%)}$ & $25.84$ & $\mathbf{0.192}$ & $\mathbf{0.728}$ \\
\multirow{2}{*}{Non-linear Deblur} & DPS & $0.079$ & $0.076$ & $0.0236$ & $1.007$ & $23.49$ & $0.234$ & $0.659$ \\
 & RFJS (ours) & $\mathbf{0.065}\;{\scriptstyle (+16.9\%)}$ & $\mathbf{0.058}\;{\scriptstyle (+24.4\%)}$ & $\mathbf{0.0236}\;{\scriptstyle (+0.2\%)}$ & $\mathbf{0.374}\;{\scriptstyle (+62.8\%)}$ & $23.33$ & $0.236$ & $0.654$ \\
\multirow{2}{*}{Blind Deblur} & BlindDPS & $0.069$ & $0.061$ & $0.0241$ & $0.788$ & $24.41$ & $0.214$ & $0.696$ \\
 & RFJS (ours) & $\mathbf{0.058}\;{\scriptstyle (+15.5\%)}$ & $\mathbf{0.050}\;{\scriptstyle (+18.0\%)}$ & $\mathbf{0.0230}\;{\scriptstyle (+4.6\%)}$ & $\mathbf{0.339}\;{\scriptstyle (+57.0\%)}$ & $\mathbf{24.50}$ & $\mathbf{0.211}$ & $\mathbf{0.700}$ \\
\midrule
\multicolumn{9}{l}{\emph{Severe degradation (${\rm dps~scale} = 0.8$)}} \\
\midrule
\multirow{2}{*}{Box Inpainting $116^2$} & DPS & $0.077$ & $0.085$ & $0.0292$ & $0.697$ & $23.40$ & $0.143$ & $0.835$ \\
 & RFJS (ours) & $\mathbf{0.054}\;{\scriptstyle (+29.1\%)}$ & $\mathbf{0.056}\;{\scriptstyle (+34.1\%)}$ & $\mathbf{0.0267}\;{\scriptstyle (+8.8\%)}$ & $\mathbf{0.292}\;{\scriptstyle (+58.1\%)}$ & $\mathbf{24.08}$ & $\mathbf{0.139}$ & $\mathbf{0.841}$ \\
\multirow{2}{*}{Box Inpainting $135^2$} & DPS & $0.085$ & $0.095$ & $0.0307$ & $0.815$ & $21.54$ & $0.164$ & $0.800$ \\
 & RFJS (ours) & $\mathbf{0.061}\;{\scriptstyle (+28.7\%)}$ & $\mathbf{0.066}\;{\scriptstyle (+30.5\%)}$ & $\mathbf{0.0281}\;{\scriptstyle (+8.2\%)}$ & $\mathbf{0.324}\;{\scriptstyle (+60.3\%)}$ & $\mathbf{21.72}$ & $\mathbf{0.159}$ & $\mathbf{0.805}$ \\
\multirow{2}{*}{Super Resolution $\times 12$} & DPS & $0.137$ & $0.151$ & $0.0288$ & $1.239$ & $22.31$ & $0.267$ & $0.603$ \\
 & RFJS (ours) & $\mathbf{0.101}\;{\scriptstyle (+26.3\%)}$ & $\mathbf{0.105}\;{\scriptstyle (+30.4\%)}$ & $\mathbf{0.0286}\;{\scriptstyle (+0.9\%)}$ & $\mathbf{0.443}\;{\scriptstyle (+64.2\%)}$ & $\mathbf{22.46}$ & $\mathbf{0.264}$ & $\mathbf{0.607}$ \\
\multirow{2}{*}{Super Resolution $\times 16$} & DPS & $0.178$ & $0.202$ & $0.0331$ & $1.310$ & $20.98$ & $0.296$ & $0.563$ \\
 & RFJS (ours) & $\mathbf{0.127}\;{\scriptstyle (+29.0\%)}$ & $\mathbf{0.136}\;{\scriptstyle (+32.7\%)}$ & $\mathbf{0.0318}\;{\scriptstyle (+4.1\%)}$ & $\mathbf{0.481}\;{\scriptstyle (+63.3\%)}$ & $\mathbf{21.04}$ & $\mathbf{0.293}$ & $\mathbf{0.567}$ \\
\multirow{2}{*}{Super Resolution $\times 32$} & DPS & $0.291$ & $0.330$ & $0.0501$ & $1.386$ & $18.00$ & $0.367$ & $0.475$ \\
 & RFJS (ours) & $\mathbf{0.210}\;{\scriptstyle (+27.8\%)}$ & $\mathbf{0.228}\;{\scriptstyle (+30.9\%)}$ & $\mathbf{0.0477}\;{\scriptstyle (+4.9\%)}$ & $\mathbf{0.558}\;{\scriptstyle (+59.7\%)}$ & $\mathbf{18.04}$ & $\mathbf{0.362}$ & $\mathbf{0.481}$ \\
\multirow{2}{*}{Motion Deblur $K{=}256$} & DPS & $0.120$ & $0.123$ & $0.0292$ & $1.212$ & $22.18$ & $0.263$ & $0.604$ \\
 & RFJS (ours) & $\mathbf{0.088}\;{\scriptstyle (+27.2\%)}$ & $\mathbf{0.085}\;{\scriptstyle (+30.4\%)}$ & $\mathbf{0.0271}\;{\scriptstyle (+7.0\%)}$ & $\mathbf{0.439}\;{\scriptstyle (+63.8\%)}$ & $\mathbf{22.20}$ & $\mathbf{0.261}$ & $\mathbf{0.607}$ \\
\bottomrule
\end{tabular}}%
}
\vspace{5pt}
\caption{\rev{Search with side information improves other perceptual metrics which are not
directly derived from the reward modeling network, here over all twelve tasks: the full version of
\cref{tab:perceptual_metrics_main}. RFJS attains the better value in every DreamSim, CLIP, NME and
FS cell, and in most PSNR, LPIPS and SSIM cells. Percentages give the relative improvement of RFJS
over DPS. The two degradation groups are not comparable across the group heading. Throughout,
$64$ images with $N = 8$ particles and a reference image as side information, with selection by
the side-information reward alone; FAN detected a face in every ground truth and every
reconstruction, so no sample is dropped.}}
\label{tab:perceptual_metrics}
\end{table}

\begin{figure}[H]
  \centering
  \makebox[\textwidth]{
  \scalebox{0.75}{
  \begin{tabular}{c}
  \begin{tabular}{>{\centering\arraybackslash}m{0.5cm} *{6}{>{\centering\arraybackslash}m{2.2cm}}}
    & \small Measurement & \small Ground Truth & \small RFJS(Ours) & \small DPS(1) & \small DPS(2)  & \small DPS(3) \\

    {\centering\rotatebox{90}{\small SR(32x)}} &
    \includegraphics[width=\linewidth]{pictures/hard_problems/DPS/super32/21inp.jpg} &
    \includegraphics[width=\linewidth]{pictures/hard_problems/DPS/super32/21gt.jpg} &
    \includegraphics[width=\linewidth]{pictures/hard_problems/DPS/super32/21s.jpg} &
    \includegraphics[width=\linewidth]{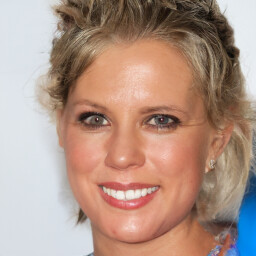} &
    \includegraphics[width=\linewidth]{pictures/hard_problems/DPS/super32/21d1.jpg} &
    \includegraphics[width=\linewidth]{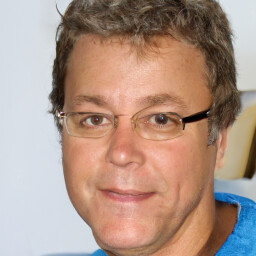}\\ [-2pt]

    {\centering\rotatebox{90}{\small SR(32x)}} &
    \includegraphics[width=\linewidth]{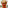} &
    \includegraphics[width=\linewidth]{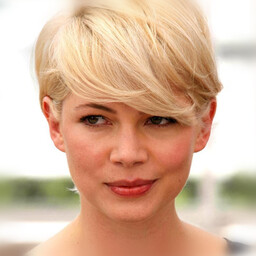} &
    \includegraphics[width=\linewidth]{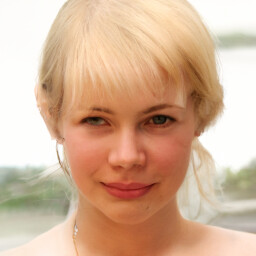} &
    \includegraphics[width=\linewidth]{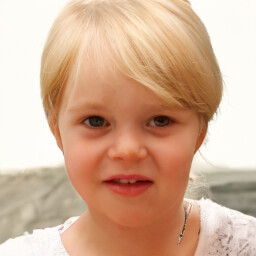} &
    \includegraphics[width=\linewidth]{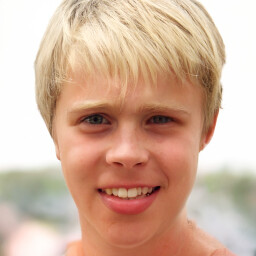} &
    \includegraphics[width=\linewidth]{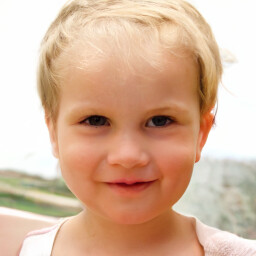}\\ [-2pt]

    {\centering\rotatebox{90}{\small SR(32x)}} &
    \includegraphics[width=\linewidth]{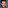} &
    \includegraphics[width=\linewidth]{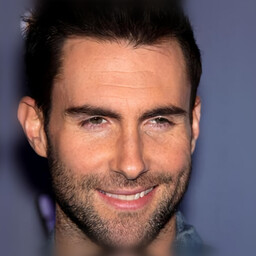} &
    \includegraphics[width=\linewidth]{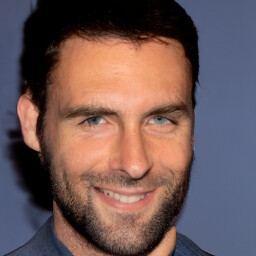} &
    \includegraphics[width=\linewidth]{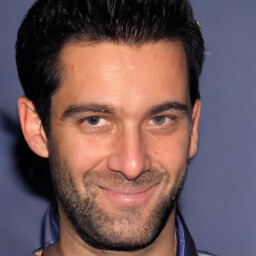} &
    \includegraphics[width=\linewidth]{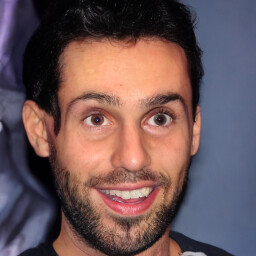} &
    \includegraphics[width=\linewidth]{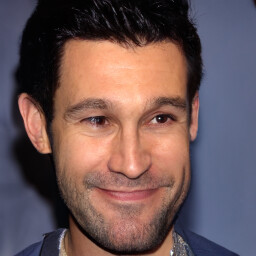}\\ [-2pt]

    {\centering\rotatebox{90}{\small SR(32x)}} &
    \includegraphics[width=\linewidth]{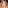} &
    \includegraphics[width=\linewidth]{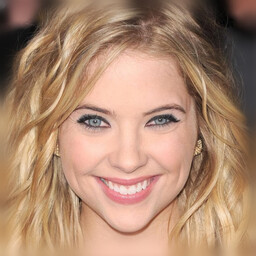} &
    \includegraphics[width=\linewidth]{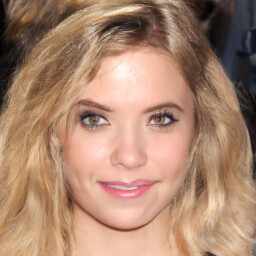} &
    \includegraphics[width=\linewidth]{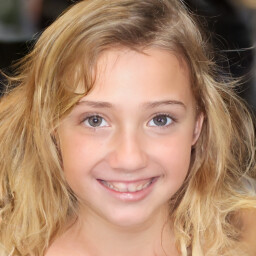} &
    \includegraphics[width=\linewidth]{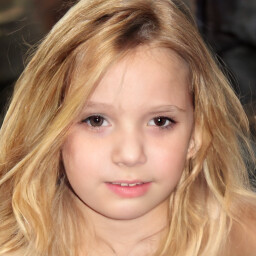} &
    \includegraphics[width=\linewidth]{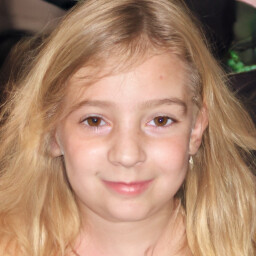}\\ [-2pt]

  \end{tabular}
  \end{tabular}
  }} % END \makebox and \scalebox
  \caption{Our RFJS reconstructions preserve identity and perceptual details far better than DPS, as clearly visible from the images. This strong perceptual improvement is not reflected by classical metrics, which are reported in the table below.}
  \label{fig:dps_hard_tasks_sr_32x}
\end{figure}

{\setlength{\tabcolsep}{6pt}
\begin{table}[H]
\caption{PSNR, LPIPS, and SSIM for the reconstructions shown in the figure above. Classical metrics fail to reflect the significant improvement.}
\label{tab:side_by_side_std_metrics}
\centering
\begin{tabular}{| c | c c | c c | c c |}
\toprule
\multirow{2}{*}{Image ID} & \multicolumn{2}{c|}{\textbf{PSNR}} & \multicolumn{2}{c|}{\textbf{LPIPS}}  & \multicolumn{2}{c|}{\textbf{SSIM}} \\
 & RFJS & DPS1 & RFJS & DPS1 & RFJS & DPS1 \\
\midrule
1 & 20.30 & 19.98 & 0.363 & 0.353 & 0.549 & 0.541 \\
2 & 19.12 & 18.77 & 0.382 & 0.413 & 0.567 & 0.552 \\
3 & 20.95 & 20.98 & 0.254 & 0.273 & 0.609 & 0.596 \\
4 & 17.85 & 18.54 & 0.411 & 0.406 & 0.368 & 0.389 \\
\midrule
\textbf{Avg}
& \textbf{19.56} & \textbf{19.57} & \textbf{0.353} & \textbf{0.361} & \textbf{0.523}  & \textbf{0.520} \\
\bottomrule
\end{tabular}
\end{table}
}

}

\subsection{DPS}
\label{app:dps-more-examples}

\begin{figure}[H]
  \centering
  \makebox[\textwidth]{
  \scalebox{0.52}{
  \begin{tabular}{c c c c}
  \begin{tabular}{>{\centering\arraybackslash}m{0.5cm} *{6}{>{\centering\arraybackslash}m{2.2cm}}>{\centering\arraybackslash}m{0.5cm} *{6}{>{\centering\arraybackslash}m{2.2cm}}}
    & \small Side & \small Measurement & \small DPS & \small RGG & \small RFJS (Ours) & \small Ground Truth &
    & \small Side & \small Measurement & \small DPS & \small RGG & \small RFJS (Ours) & \small Ground Truth \\

    {\centering\rotatebox{90}{\footnotesize Box Inpainting }} &
    \includegraphics[width=\linewidth]{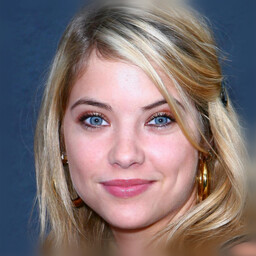} &
    \includegraphics[width=\linewidth]{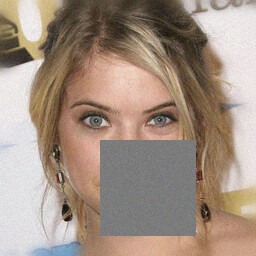} &
    \includegraphics[width=\linewidth]{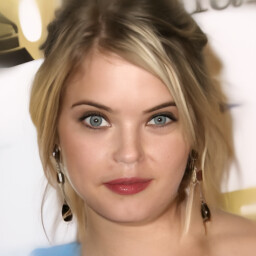} &
    \includegraphics[width=\linewidth]{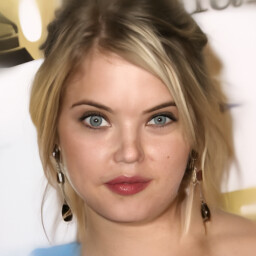} &
    \includegraphics[width=\linewidth]{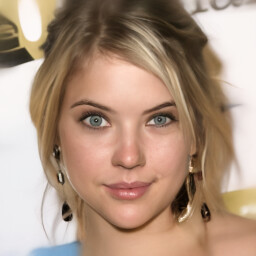} &
    \includegraphics[width=\linewidth]{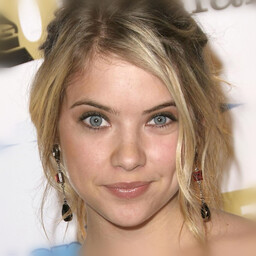} &

    {\centering\rotatebox{90}{\footnotesize Box Inpainting }} &
    \includegraphics[width=\linewidth]{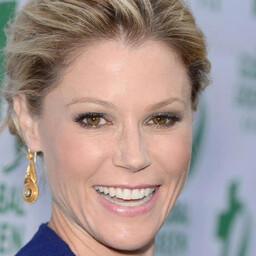} &
    \includegraphics[width=\linewidth]{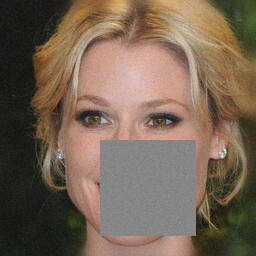} &
    \includegraphics[width=\linewidth]{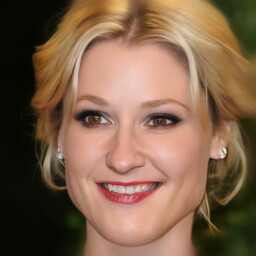} &
    \includegraphics[width=\linewidth]{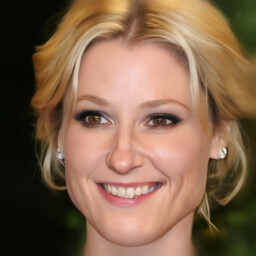} &
    \includegraphics[width=\linewidth]{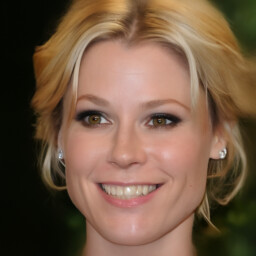} &
    \includegraphics[width=\linewidth]{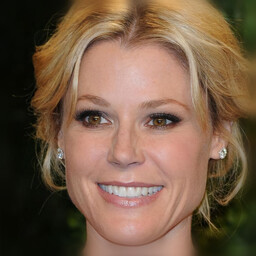} \\ [-2pt]

    {\centering\rotatebox{90}{\small Super Resolution}} &
    \includegraphics[width=\linewidth]{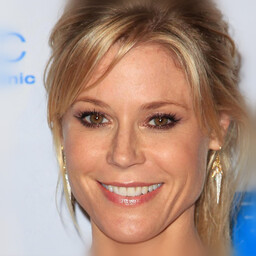} &
    \includegraphics[width=\linewidth]{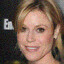} &
    \includegraphics[width=\linewidth]{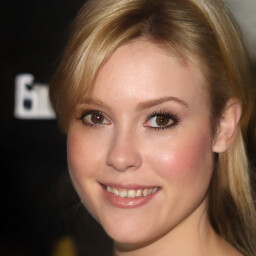} &
    \includegraphics[width=\linewidth]{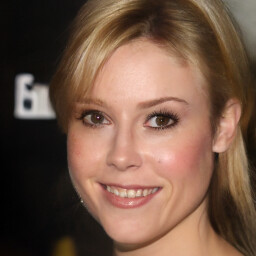} &
    \includegraphics[width=\linewidth]{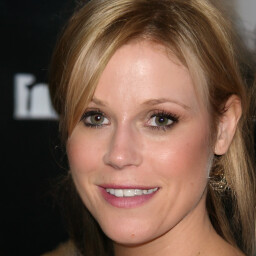} &
    \includegraphics[width=\linewidth]{pictures/DPS/label/img_00020.jpg} &

    {\centering\rotatebox{90}{\footnotesize Super Resolution }}
    & \includegraphics[width=\linewidth]{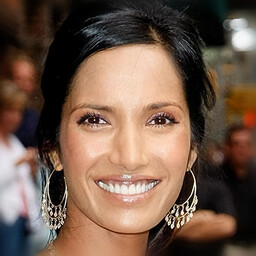} &
    \includegraphics[width=\linewidth]{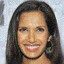} &
    \includegraphics[width=\linewidth]{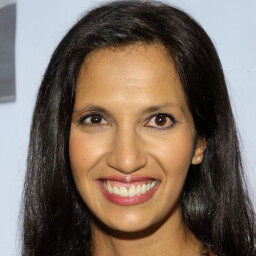} &
    \includegraphics[width=\linewidth]{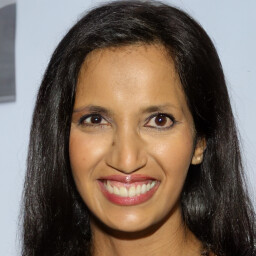} &
    \includegraphics[width=\linewidth]{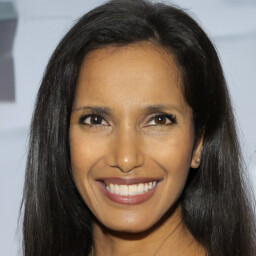} &
    \includegraphics[width=\linewidth]{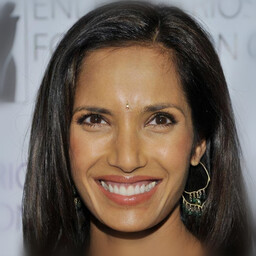}
    \\ [-2pt]

    {\centering\rotatebox{90}{\footnotesize Motion Deblur }} &
    \includegraphics[width=\linewidth]{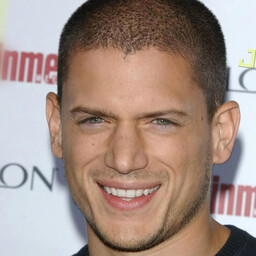} &
    \includegraphics[width=\linewidth]{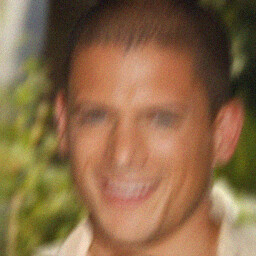} &
    \includegraphics[width=\linewidth]{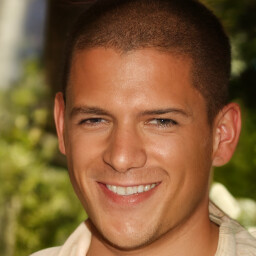} &
    \includegraphics[width=\linewidth]{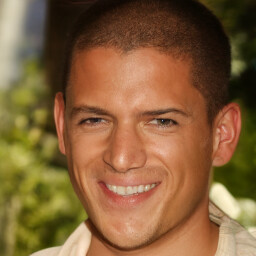} &
    \includegraphics[width=\linewidth]{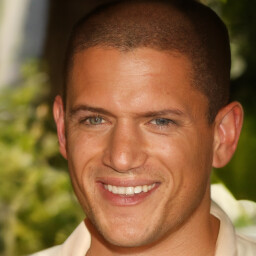} &
    \includegraphics[width=\linewidth]{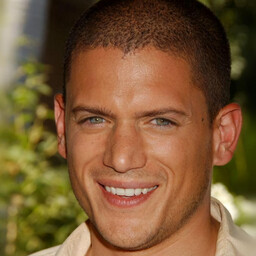} &

    {\centering\rotatebox{90}{\footnotesize Motion Deblur }} &
    \includegraphics[width=\linewidth]{pictures/DPS/label/img_00038.jpg} &
    \includegraphics[width=\linewidth]{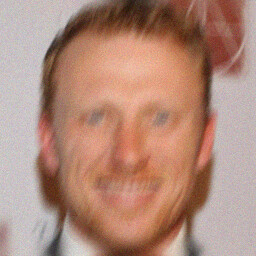} &
    \includegraphics[width=\linewidth]{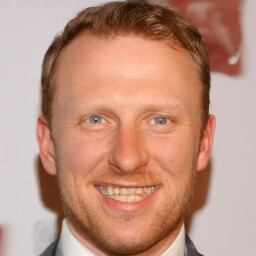} &
    \includegraphics[width=\linewidth]{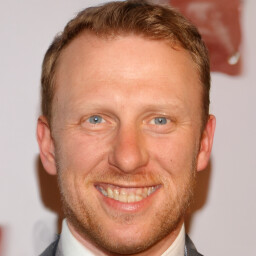} &
    \includegraphics[width=\linewidth]{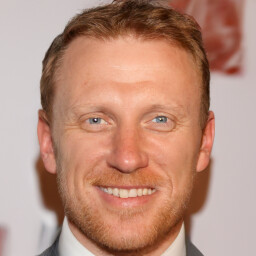} &
    \includegraphics[width=\linewidth]{pictures/DPS/label/img_00039.jpg} \\ [-2pt]

    {\centering\rotatebox{90}{\footnotesize Gaussian Deblur }} &
    \includegraphics[width=\linewidth]{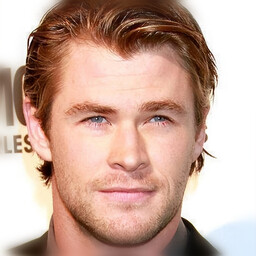} &
    \includegraphics[width=\linewidth]{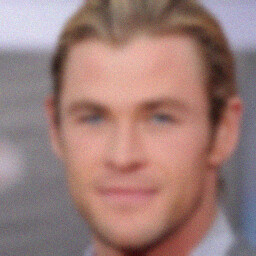} &
    \includegraphics[width=\linewidth]{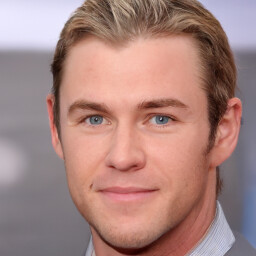} &
    \includegraphics[width=\linewidth]{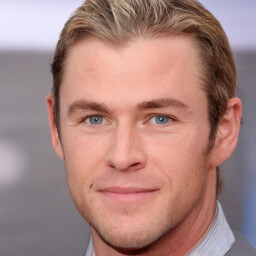} &
    \includegraphics[width=\linewidth]{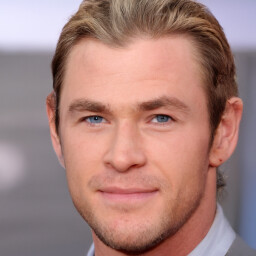} &
    \includegraphics[width=\linewidth]{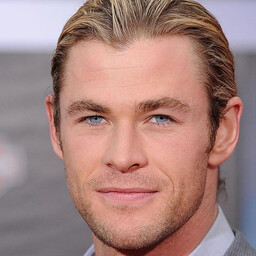} &

    {\centering\rotatebox{90}{\footnotesize Gaussian Deblur }} &
    \includegraphics[width=\linewidth]{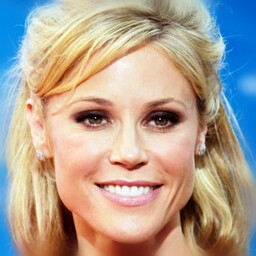} &
    \includegraphics[width=\linewidth]{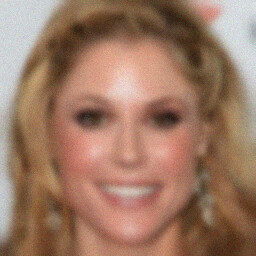} &
    \includegraphics[width=\linewidth]{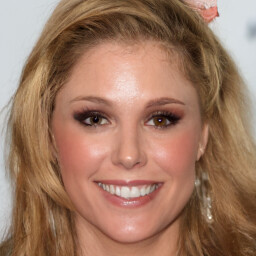} &
    \includegraphics[width=\linewidth]{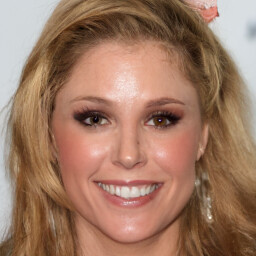} &
    \includegraphics[width=\linewidth]{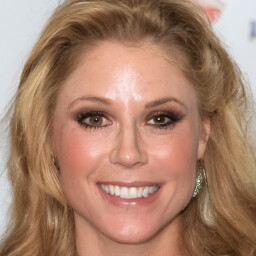} &
    \includegraphics[width=\linewidth]{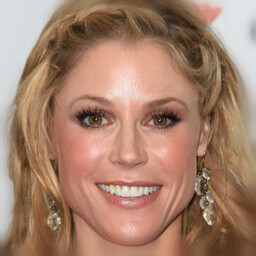} \\ [-2pt]

    {\centering\rotatebox{90}{\footnotesize Nonlinear Deblur }} &
    \includegraphics[width=\linewidth]{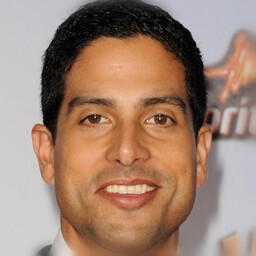} &
    \includegraphics[width=\linewidth]{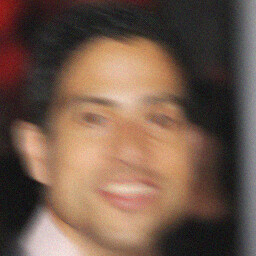} &
    \includegraphics[width=\linewidth]{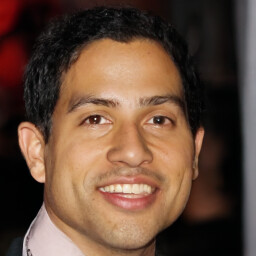} &
    \includegraphics[width=\linewidth]{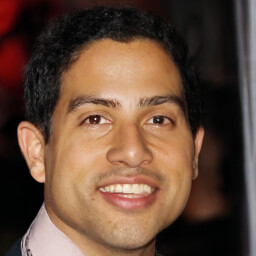} &
    \includegraphics[width=\linewidth]{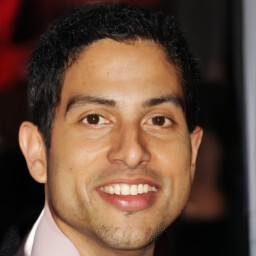} &
    \includegraphics[width=\linewidth]{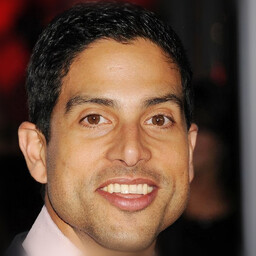} &

    {\centering\rotatebox{90}{\footnotesize Nonlinear Deblur }} &
    \includegraphics[width=\linewidth]{pictures/DPS/label/img_00048.jpg} &
    \includegraphics[width=\linewidth]{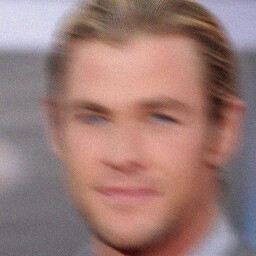} &
    \includegraphics[width=\linewidth]{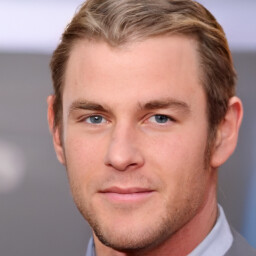} &
    \includegraphics[width=\linewidth]{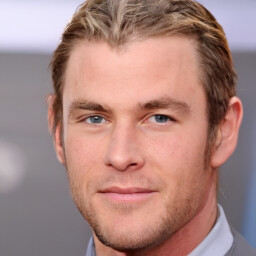} &
    \includegraphics[width=\linewidth]{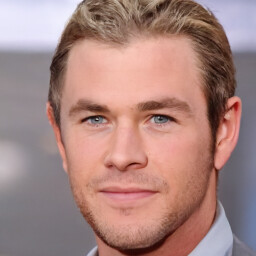} &
    \includegraphics[width=\linewidth]{pictures/DPS/label/img_00049.jpg} \\ [-2pt]

    {\centering\rotatebox{90}{\footnotesize Blind Deblur }} &
    \includegraphics[width=\linewidth]{pictures/DPS/label/img_00029.jpg} &
    \includegraphics[width=\linewidth]{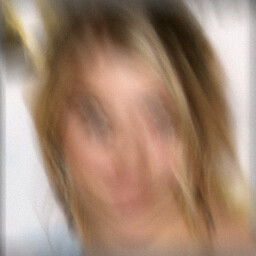} &
    \includegraphics[width=\linewidth]{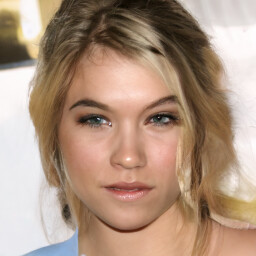} &
    \includegraphics[width=\linewidth]{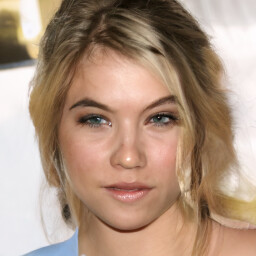} &
    \includegraphics[width=\linewidth]{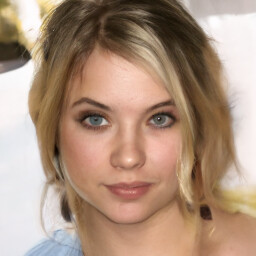} &
    \includegraphics[width=\linewidth]{pictures/DPS/label/img_00028.jpg} &

    {\centering\rotatebox{90}{\footnotesize Blind Deblur }} &
    \includegraphics[width=\linewidth]{pictures/DPS/label/img_00038.jpg} &
    \includegraphics[width=\linewidth]{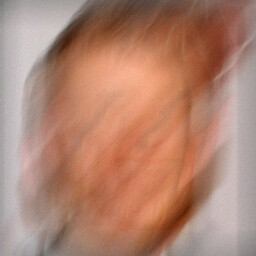} &
    \includegraphics[width=\linewidth]{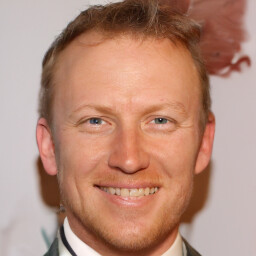} &
    \includegraphics[width=\linewidth]{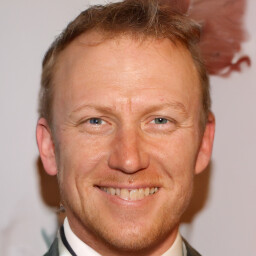} &
    \includegraphics[width=\linewidth]{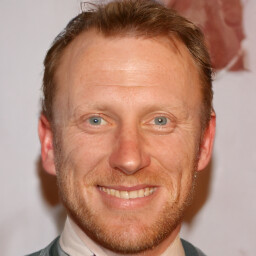} &
    \includegraphics[width=\linewidth]{pictures/DPS/label/img_00039.jpg} \\ [-2pt]

\end{tabular}

  \end{tabular}
  }} % END \makebox and \scalebox
  \caption{Additional samples using DPS as the base sampler.}
  \label{fig:qualitative_results_additional}
\end{figure}

We also present additional qualitative examples for our DPS experiments in Figures~\ref{fig:qualitative_results_additional} and~\ref{fig:qualitative_results_text_reward_additional}, corresponding to settings that use face and text as side information, respectively.

\newcommand{\tasklabel}[3][0pt]{%
  \multirow{2}{*}[#1]{%
    \parbox[c][#2][c]{0.5cm}{\centering\rotatebox{90}{\footnotesize #3}}%
  }%
}

{\setlength{\tabcolsep}{1pt}
\renewcommand{\arraystretch}{1.0}
\begin{figure}[H]
  \centering
  \makebox[\textwidth]{%
  \scalebox{0.75}{%
  \begin{tabular}{c c}
  \begin{tabular}{>{\centering\arraybackslash}m{0.5cm} *{4}{>{\centering\arraybackslash}m{2.2cm}} >{\centering\arraybackslash}m{0.5cm} *{4}{>{\centering\arraybackslash}m{2.2cm}} }
    & \small Measurement & \small DPS & \small RFJ Search & \small Ground Truth &
    & \small Measurement & \small DPS & \small RFJ Search & \small Ground Truth \\
    \tasklabel{0cm}{Box Inpainting (138)} &
    \includegraphics[width=\linewidth]{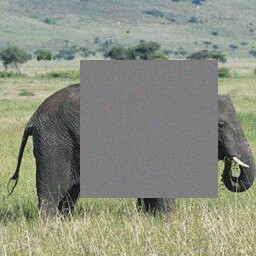} &
    \includegraphics[width=\linewidth]{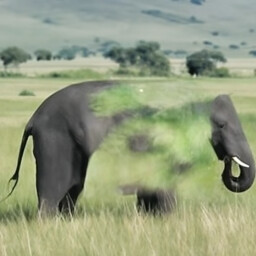} &
    \includegraphics[width=\linewidth]{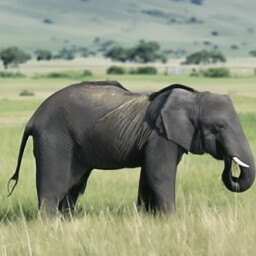} &
    \includegraphics[width=\linewidth]{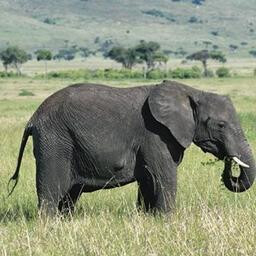}
    &
    \tasklabel{0cm}{Super-resolution (32)} &
    \includegraphics[width=\linewidth]{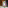} &
    \includegraphics[width=\linewidth]{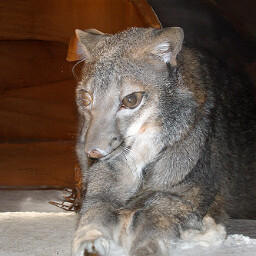} &
    \includegraphics[width=\linewidth]{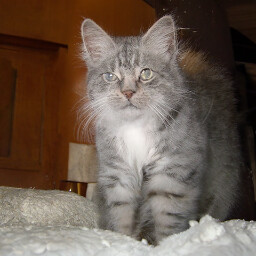} &
    \includegraphics[width=\linewidth]{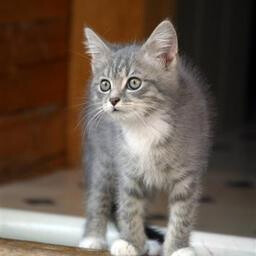}
     \\
    & \multicolumn{4}{l}{\parbox[c][1.5cm][c]{8.5cm}{Side: \textit{An elephant stands in tall grass on a savanna with distant trees.}}}
    & & \multicolumn{4}{l}{\parbox[c][1.5cm][c]{8.5cm}{Side: \textit{A fluffy gray kitten with white markings.}}} \\
    \tasklabel{0cm}{Motion Deblur} &
    \includegraphics[width=\linewidth]{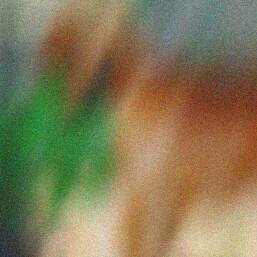} &
    \includegraphics[width=\linewidth]{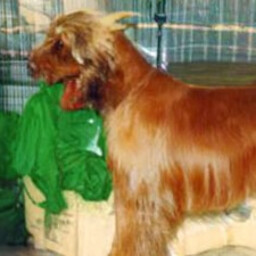} &
    \includegraphics[width=\linewidth]{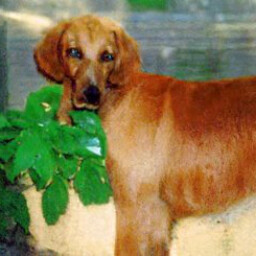} &
    \includegraphics[width=\linewidth]{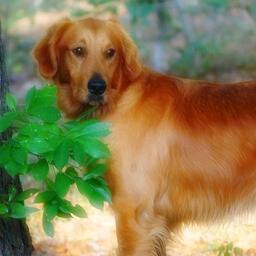}
    &
    \tasklabel{0cm}{Gaussian Deblur} &
    \includegraphics[width=\linewidth]{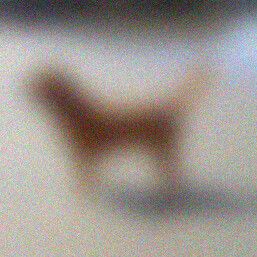} &
    \includegraphics[width=\linewidth]{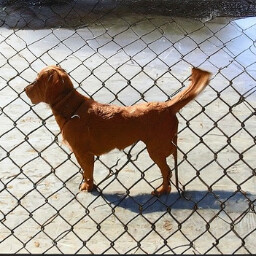} &
    \includegraphics[width=\linewidth]{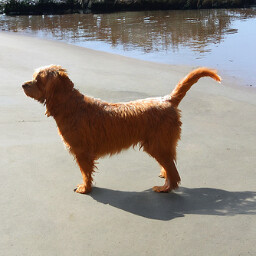} &
    \includegraphics[width=\linewidth]{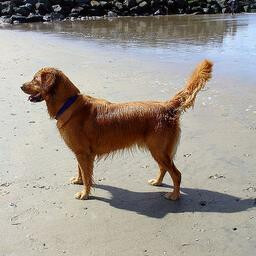}  \\
    & \multicolumn{4}{l}{\parbox[c][1.5cm][c]{8.5cm}{Side: \textit{A reddish-brown golden retriever stands upright in a forest, green leaves on the left.}}}
    & & \multicolumn{4}{l}{\parbox[c][1.5cm][c]{8.5cm}{Side: \textit{A wet golden retriever stands on a sandy beach, facing left.}}} \\

    \tasklabel{0cm}{Nonlinear Deblur} &
    \includegraphics[width=\linewidth]{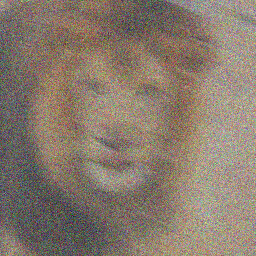} &
    \includegraphics[width=\linewidth]{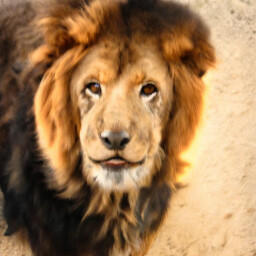} &
    \includegraphics[width=\linewidth]{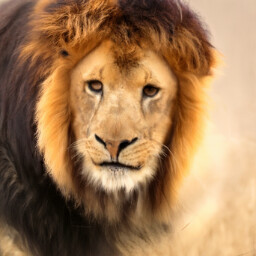} &
    \includegraphics[width=\linewidth]{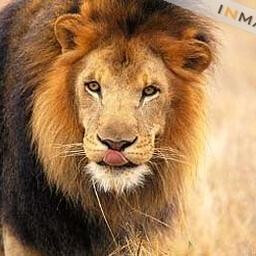}
    &
    \tasklabel{0cm}{Blind Deblur} &
    \includegraphics[width=\linewidth]{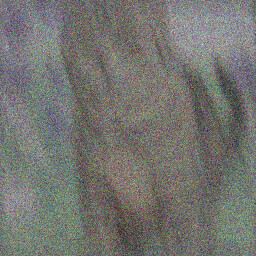} &
    \includegraphics[width=\linewidth]{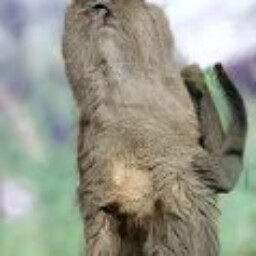} &
    \includegraphics[width=\linewidth]{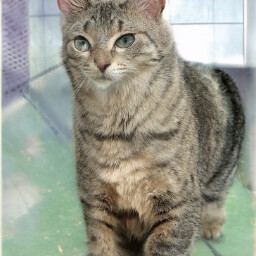} &
    \includegraphics[width=\linewidth]{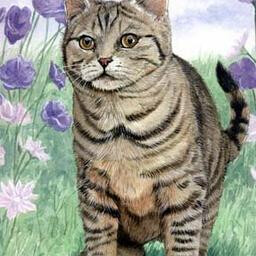}  \\
    & \multicolumn{4}{l}{\parbox[c][1.5cm][c]{8.5cm}{Side: \textit{A male lion with a black-to-rust mane, surrounded by soft dry grass.}}}
    & & \multicolumn{4}{l}{\parbox[c][1.5cm][c]{8.5cm}{Side: \textit{A tabby cat with dark stripes.}}} \\
  \end{tabular}
  \end{tabular}
  }}% END \scalebox
  \caption{Qualitative comparison on ImageNet with textual side information. For highly degraded measurements, DPS and BlindDPS often produce artifacts, whereas our method reduces these by better aligning reconstructions with the description.
}
  \label{fig:qualitative_results_text_reward_additional}
\end{figure}
}

\subsection{\rev{Expanded Text Evaluation on 75 Images}}
\label{app:text75}

\rev{The textual side-information results of \cref{tab:quantitative_results_text_reward} are
computed on the $25$ ImageNet images described in \cref{sec:experiment_setup}. To check that they
do not depend on that particular sample, we repeat the whole evaluation on an expanded set of $75$
images that contains the original ones as a subset, leaving every other setting unchanged
($N = 4$ particles, resampling base $B = 100$; see \cref{app: text_hyperparameters}). We also add
DreamSim \citep{fu2023dreamsim}, a perceptual distance fit to human similarity judgements, as a
fifth metric; like PSNR, SSIM and LPIPS it is computed against the ground truth and takes no part
in selection, which uses the side-information reward alone. \Cref{tab:text75} reports the
result.}

\begin{table}[ht!]
\centering
\footnotesize
\resizebox{\textwidth}{!}{%
\rev{%
\begin{tabular}{l|ccccc|ccccc}
\toprule
& \multicolumn{5}{c|}{Box Inpainting} & \multicolumn{5}{c}{Super Resolution ($\times$32)} \\
\midrule
Algorithm & CS ($\uparrow$) & PSNR ($\uparrow$) & SSIM ($\uparrow$) & LPIPS ($\downarrow$) & DS ($\downarrow$) & CS ($\uparrow$) & PSNR ($\uparrow$) & SSIM ($\uparrow$) & LPIPS ($\downarrow$) & DS ($\downarrow$) \\
\midrule
RFJS (ours) & $\mathbf{0.892}$ & $\mathbf{18.80}$ & $\mathbf{0.636}$ & $\mathbf{0.301}$ & $\mathbf{0.177}$ & $\mathbf{0.805}$ & $\underline{15.39}$ & $\mathbf{0.281}$ & $\mathbf{0.520}$ & $\mathbf{0.320}$ \\
GS (ours) & $\underline{0.890}$ & $18.59$ & $\underline{0.635}$ & $\underline{0.305}$ & $\underline{0.182}$ & $\underline{0.798}$ & $\mathbf{15.45}$ & $\underline{0.278}$ & $\underline{0.523}$ & $\underline{0.336}$ \\
BoN (w/ side) & $0.883$ & $\underline{18.70}$ & $0.634$ & $0.308$ & $0.191$ & $0.771$ & $15.35$ & $0.271$ & $0.526$ & $0.367$ \\
DPS & $0.861$ & $18.23$ & $0.629$ & $0.317$ & $0.233$ & $0.739$ & $15.38$ & $0.271$ & $0.535$ & $0.419$ \\
\midrule
& \multicolumn{5}{c|}{Gaussian Deblur} & \multicolumn{5}{c}{Motion Deblur} \\
\midrule
Algorithm & CS ($\uparrow$) & PSNR ($\uparrow$) & SSIM ($\uparrow$) & LPIPS ($\downarrow$) & DS ($\downarrow$) & CS ($\uparrow$) & PSNR ($\uparrow$) & SSIM ($\uparrow$) & LPIPS ($\downarrow$) & DS ($\downarrow$) \\
\midrule
RFJS (ours) & $\mathbf{0.843}$ & $16.17$ & $0.294$ & $\underline{0.468}$ & $\underline{0.249}$ & $\mathbf{0.865}$ & $\underline{17.13}$ & $\underline{0.344}$ & $\mathbf{0.441}$ & $\mathbf{0.202}$ \\
GS (ours) & $0.834$ & $\underline{16.34}$ & $\underline{0.297}$ & $\underline{0.468}$ & $0.250$ & $\underline{0.860}$ & $16.73$ & $0.327$ & $0.453$ & $\underline{0.212}$ \\
BoN (w/ side) & $\underline{0.841}$ & $\mathbf{16.82}$ & $\mathbf{0.319}$ & $\mathbf{0.454}$ & $\mathbf{0.243}$ & $0.848$ & $\mathbf{17.29}$ & $\mathbf{0.346}$ & $\underline{0.446}$ & $0.219$ \\
DPS & $0.788$ & $15.66$ & $0.285$ & $0.493$ & $0.337$ & $0.810$ & $16.59$ & $0.322$ & $0.472$ & $0.293$ \\
\midrule
& \multicolumn{5}{c|}{Non-linear Deblur} & \multicolumn{5}{c}{Blind Deblur} \\
\midrule
Algorithm & CS ($\uparrow$) & PSNR ($\uparrow$) & SSIM ($\uparrow$) & LPIPS ($\downarrow$) & DS ($\downarrow$) & CS ($\uparrow$) & PSNR ($\uparrow$) & SSIM ($\uparrow$) & LPIPS ($\downarrow$) & DS ($\downarrow$) \\
\midrule
RFJS (ours) & $\underline{0.874}$ & $18.52$ & $\underline{0.410}$ & $\underline{0.410}$ & $0.195$ & $\mathbf{0.894}$ & $19.94$ & $0.494$ & $\mathbf{0.354}$ & $\mathbf{0.138}$ \\
GS (ours) & $0.871$ & $\mathbf{18.64}$ & $0.409$ & $\mathbf{0.408}$ & $\underline{0.191}$ & $\underline{0.890}$ & $20.01$ & $0.497$ & $\mathbf{0.354}$ & $\underline{0.140}$ \\
BoN (w/ side) & $\mathbf{0.876}$ & $\underline{18.58}$ & $\mathbf{0.413}$ & $0.414$ & $\mathbf{0.190}$ & $0.884$ & $\mathbf{20.11}$ & $\underline{0.498}$ & $\underline{0.355}$ & $0.148$ \\
DPS / BlindDPS & $0.854$ & $18.55$ & $\mathbf{0.413}$ & $0.420$ & $0.222$ & $0.873$ & $\underline{20.07}$ & $\mathbf{0.502}$ & $0.367$ & $0.169$ \\
\bottomrule
\end{tabular}%
}}
\vspace{5pt}
\caption{\rev{\textbf{Text as side information, $75$ images.} The evaluation of
\cref{tab:quantitative_results_text_reward} repeated on a three-times-larger set, with DreamSim
(DS) added. Bold is best and underline second best within each task; values that tie at the
reported precision share the same emphasis. The baseline row is DPS for the first five tasks
and BlindDPS for blind deblur. RFJS and GS lead on CS and DreamSim on every task, while PSNR
and SSIM are led by Best-of-$N$ on several of them.}}
\label{tab:text75}
\end{table}

\rev{The conclusions of the main text carry over to the larger set. RFJS and GS lead the base
solver on the metrics that track agreement with the side information or with human perceptual
judgement -- CS, LPIPS and DreamSim -- on all six tasks, with DreamSim improving by between
$12\%$ and $31\%$ and CS by between $2\%$ and $9\%$. PSNR and SSIM stay in the same range as the
baseline, better on some tasks and level on the rest, and worse on none. This is the pattern
\cref{app:importance_of_target_metric} describes: search concentrates mass on reconstructions
that agree with $\vs$, which a pixel-wise average does not reward.}

\rev{Against Best-of-$N$, which draws the same $N$ samples and keeps the one the reward prefers
but never resamples during the trajectory, RFJS is ahead on CS and DreamSim on most of the tasks,
so the gain comes from resampling along the trajectory and not from the final reward-based pick
alone. Gaussian deblur is the exception: there Best-of-$N$ attains the better PSNR and SSIM while
the two are level on CS.}

\rev{RFJS and GS are close to each other throughout this set. We therefore read the expanded text
evaluation as evidence for reward-guided search as such, rather than for the recursive
fork--join schedule specifically over a greedy one. The two separate more clearly when the reward
is sharper: with face identity as side information, RFJS attains a lower face-similarity distance
than GS on every task in \cref{tab:quantitative_results_dps}.}

\subsection{DAPS}

\label{app:daps}

\textbf{Setup:} We employ the search and gradient modules to infuse side information using DAPS as the base sampler. We consider two challenging tasks, box inpainting with a box of size $96 \times 96$ and super-resolution with downsampling factor of $10$. For gradient guidance, we used a scale of $15$ with respect to the noise being added to $\hat{\vx}_{0\mid t}(\vx_t)$ after MCMC steps \citep{zhang2024daps}. DAPS uses fewer diffusion steps $(200)$ than DPS $(1000)$. Further, the algorithm is based on a graphical model that allows for more inherent exploration due to the decoupling between consecutive steps. Therefore, for search algorithms, relatively smaller bases are preferable, and hence $B=4$ is chosen.

\textbf{Results:} The qualitative results are given in Figure \ref{fig:qualitative_results_daps}, while the quantitative metrics are given in \ref{tab:quantitative_results_daps}. Observe that for the task of inpainting, our search algorithm shows a significant improvement over DAPS, particularly in the FS metric.

\begin{table}[ht!]
\caption{Comparison of metrics for various inverse problems using DAPS as the base sampler. For each metric, the best result is shown in \textbf{bold}, and the second best is \underline{underlined}. We observe that our RFJ Search-based algorithm has the best or the second-best performance in all the tasks.}
\label{tab:quantitative_results_daps}
\centering
\footnotesize
\resizebox{0.99\textwidth}{!}{%
\begin{tabular}{lcccc|cccc}
\toprule
& \multicolumn{4}{c|}{~Box Inpainting} & \multicolumn{4}{c}{~Super Resolution ($\times 10$)} \\
\midrule
\parbox[t]{1.5cm}{Method} & \parbox[t]{2.75cm}{\centering FS ($\downarrow$)} & \parbox[t]{1.75cm}{\centering PSNR ($\uparrow$)} & \parbox[t]{1.75cm}{\centering LPIPS ($\downarrow$)} & \parbox[t]{1.75cm}{\centering SSIM ($\uparrow$)}  & \parbox[t]{2.75cm}{\centering FS ($\downarrow$)} & \parbox[t]{1.75cm}{\centering PSNR ($\uparrow$)} & \parbox[t]{1.75cm}{\centering LPIPS ($\downarrow$)} & \parbox[t]{1.75cm}{\centering SSIM ($\uparrow$)}   \\
\midrule
RFJS (ours) & $\textbf{0.423}\scriptstyle \pm0.10$ & $\textbf{28.720}\scriptstyle \pm1.35$ & $\textbf{0.140}\scriptstyle \pm0.03$ & $\textbf{0.788}\scriptstyle \pm0.03$ & $\underline{0.654}\scriptstyle \pm0.11$ & $\underline{25.228}\scriptstyle \pm1.34$ & $\textbf{0.282}\scriptstyle \pm0.03$ & $\underline{0.661}\scriptstyle \pm0.04$ \\
GS (ours) & $0.511\scriptstyle \pm0.12$ & $28.640\scriptstyle \pm1.43$ & $\underline{0.140}\scriptstyle \pm0.03$ & $\underline{0.787}\scriptstyle \pm0.03$ & $0.760\scriptstyle \pm0.12$ & $\textbf{25.271}\scriptstyle \pm1.36$ & $0.285\scriptstyle \pm0.03$ & $\textbf{0.662}\scriptstyle \pm0.04$ \\
RGG & $\underline{0.436}\scriptstyle \pm0.12$ & $28.410\scriptstyle \pm1.39$ & $0.141\scriptstyle \pm0.03$ & $0.784\scriptstyle \pm0.03$ & $\textbf{0.579}\scriptstyle \pm0.13$ & $25.210\scriptstyle \pm1.34$ & $\underline{0.282}\scriptstyle \pm0.03$ & $0.659\scriptstyle \pm0.04$ \\
BON & $0.611\scriptstyle \pm0.14$ & $\underline{28.660}\scriptstyle \pm1.45$ & $0.141\scriptstyle \pm0.03$ & $0.787\scriptstyle \pm0.03$
& $0.909\scriptstyle \pm0.11$ & $25.220\scriptstyle \pm1.38$ & $0.285\scriptstyle \pm0.03$ & $0.660\scriptstyle \pm0.04$  \\
DAPS & $0.739\scriptstyle \pm0.18$ & $28.290\scriptstyle \pm1.53$ & $0.142\scriptstyle \pm0.03$ & $0.784\scriptstyle \pm 0.03$ & $1.020\scriptstyle \pm0.14$ & $25.170\scriptstyle \pm1.35$ & $0.285\scriptstyle \pm0.03$ & $0.659\scriptstyle \pm0.04$ \\
\bottomrule
\end{tabular}
}
\end{table}

{
\setlength{\tabcolsep}{1pt}
\renewcommand{\arraystretch}{1.0}
\begin{figure}[ht!]
  \centering
  \makebox[\textwidth]{
  \scalebox{0.52}{
  \begin{tabular}{c c}
  \begin{tabular}{>{\centering\arraybackslash}m{0.5cm} *{6}{>{\centering\arraybackslash}m{2.2cm}}>{\centering\arraybackslash}m{0.5cm} *{6}{>{\centering\arraybackslash}m{2.2cm}}}
    & \small Side & \small Measurement & \small DAPS & \small RGG & \small RFJS (Ours) & \small Ground Truth & & \small Side & \small Measurement & \small DAPS & \small Gradient & \small RFJS (Ours) & \small Ground Truth \\
    \multirow{3}{*}{\parbox[c][5cm][c]{0.5cm}{\centering\rotatebox{90}{\footnotesize Box Inpainting (96)}}}
    &
    \includegraphics[width=\linewidth]{pictures/DPS/label/img_00028.jpg} &
    \includegraphics[width=\linewidth]{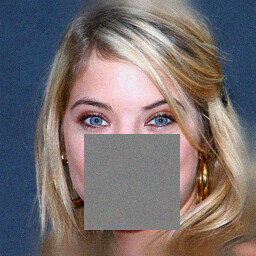} &
    \includegraphics[width=\linewidth]{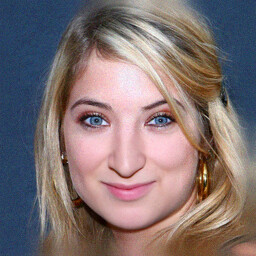} &
    \includegraphics[width=\linewidth]{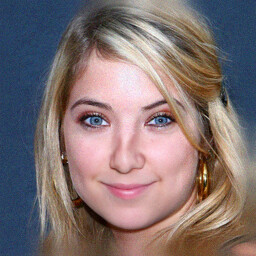} &
    \includegraphics[width=\linewidth]{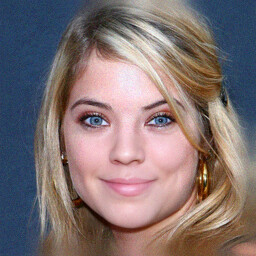} &
    \includegraphics[width=\linewidth]{pictures/DPS/label/img_00029.jpg}
    & \multirow{3}{*}{\parbox[c][5cm][c]{0.5cm}{\centering\rotatebox{90}{\footnotesize Super-resolution (10)}}}
     &
    \includegraphics[width=\linewidth]{pictures/DPS/label/img_00053.jpg} &
    \includegraphics[width=\linewidth]{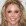} &
    \includegraphics[width=\linewidth]{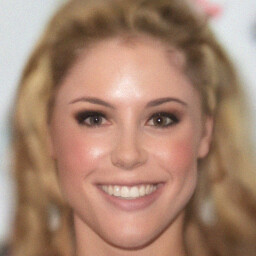} &
    \includegraphics[width=\linewidth]{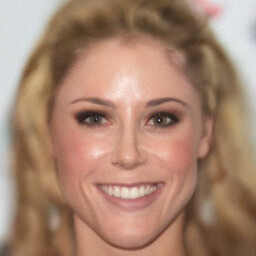} &
    \includegraphics[width=\linewidth]{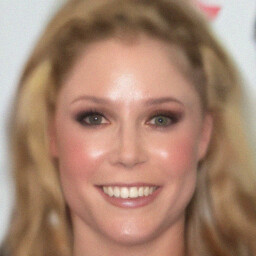} &
    \includegraphics[width=\linewidth]{pictures/DPS/label/img_00052.jpg} \\
     &
    \includegraphics[width=\linewidth]{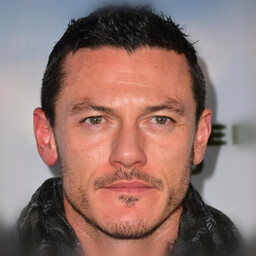} &
    \includegraphics[width=\linewidth]{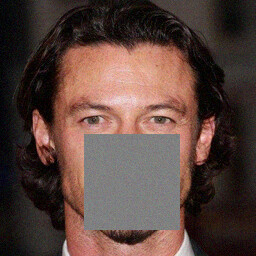} &
    \includegraphics[width=\linewidth]{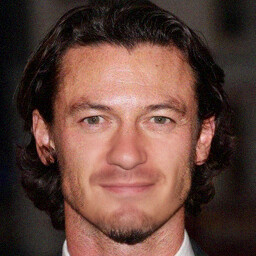} &
    \includegraphics[width=\linewidth]{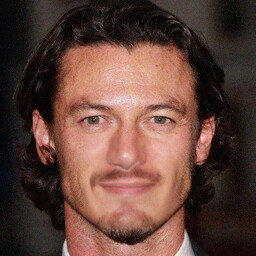} &
    \includegraphics[width=\linewidth]{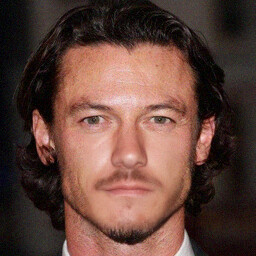} &
    \includegraphics[width=\linewidth]{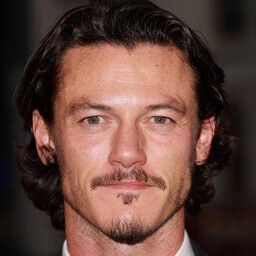} &
    &
    \includegraphics[width=\linewidth]{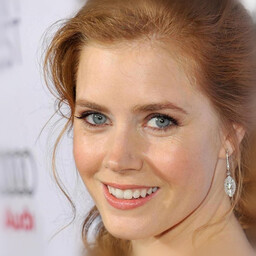} &
    \includegraphics[width=\linewidth]{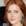} &
    \includegraphics[width=\linewidth]{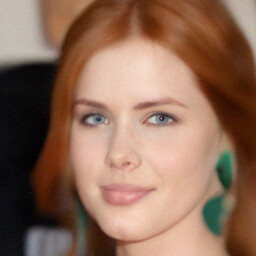} &
    \includegraphics[width=\linewidth]{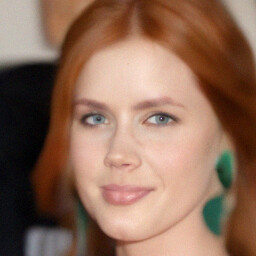} &
    \includegraphics[width=\linewidth]{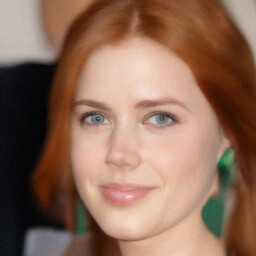} &
    \includegraphics[width=\linewidth]{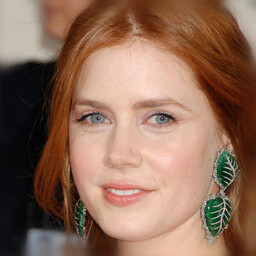} \\
     &
    \includegraphics[width=\linewidth]{pictures/DPS/label/img_00021.jpg} &
    \includegraphics[width=\linewidth]{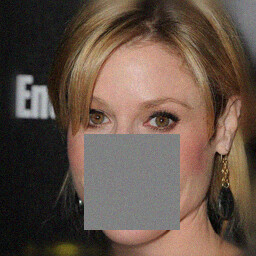} &
    \includegraphics[width=\linewidth]{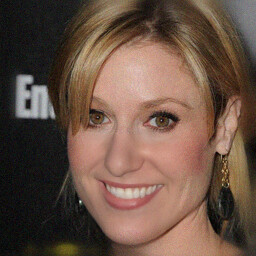} &
    \includegraphics[width=\linewidth]{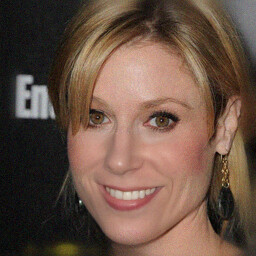} &
    \includegraphics[width=\linewidth]{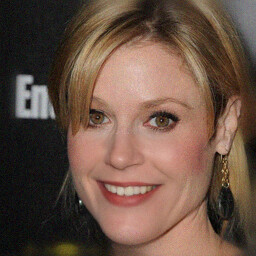} &
    \includegraphics[width=\linewidth]{pictures/DPS/label/img_00020.jpg} &
    &
    \includegraphics[width=\linewidth]{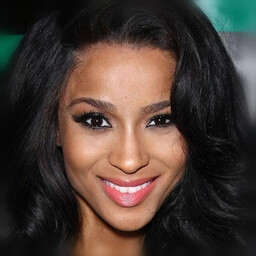} &
    \includegraphics[width=\linewidth]{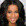} &
    \includegraphics[width=\linewidth]{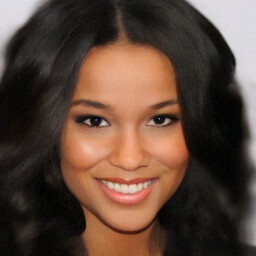} &
    \includegraphics[width=\linewidth]{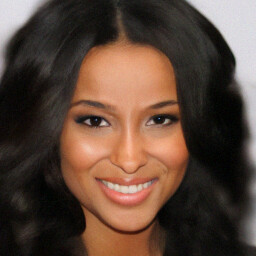} &
    \includegraphics[width=\linewidth]{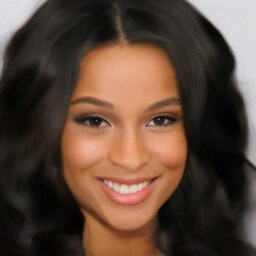} &
    \includegraphics[width=\linewidth]{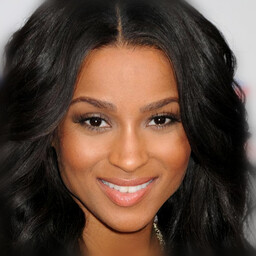} \\
\end{tabular}
  \end{tabular}
  }} % END \makebox and \scalebox
  \caption{Qualitative comparison of algorithms using DAPS as the base sampler. Our method offers better reconstructions aligned with the identity.}
  \label{fig:qualitative_results_daps}
\end{figure}
}

\subsection{MPGD}
\label{app:mpgd}

\textbf{Setup:} As in the paper \citet{he2024mpgd}, we choose super-resolution and Gaussian deblurring as the tasks, along with additional task of box inpainting. For box inpainting, we used a box of size $64\times 64$ at the center of the face. Further, we modified the down-sampling scale of super-resolution from $4$ to $6$, and the intensity of the kernel in Gaussian deblur from $3$ to $5$ to make the tasks more challenging. Even though MPGD uses $100$ DDIM steps in generation, its exploratory capabilities are similar to DPS. Therefore, we cannot use very large base $B$, whence, we choose $B=8$ for box inpainting, and super-resolution. For Gaussian deblur, we found that using $B=8$ becomes detrimental for other metrics, and so $B=16$ is used. For the gradient guidance, the scales of $0.5$ for box inpainting, and $0.25$ for super-resolution and Gaussian deblur are chosen carefully to avoid overfitting. This is the scale relative to the gradient with respect to the measurement. For more discussion and examples on the effect of gradient scale, see Appendix \ref{app:effect_of_grad_scale}.

\textbf{Results:} Experimental results with MPGD \citep{he2024mpgd} as baseline algorithm are given in Table \ref{tab:quantitative_results_mpgd}. We observe that using RFJ Search significantly enhances the FS metric, while improving the other metrics. The reconstructions that utilize the side information exhibit strong identity match with the ground truth, which is reflected in the FS metric.

\begin{table}[htbp]
\caption{Quantitative comparison of reconstruction metrics in case of inverse problems with MPGD as the base sampler. For each metric, the best result is shown in \textbf{bold}, and the second best is \underline{underlined}. Observe that our RFJ search algorithm has the best or the second-best performance in all the tasks. In Gaussian deblur, our search algorithm is only marginally worse than the best metrics attained. The value in the brackets indicates the resampling rate for search algorithms, and gradient scale for Gradient algorithm.}
\label{tab:quantitative_results_mpgd}
\centering
\footnotesize
\resizebox{0.99\textwidth}{!}{%
\begin{tabular}{l|llcccc}
\toprule
\parbox[t]{1.5cm}{Sampler} & \parbox[t]{2.5cm}{~Task} & \parbox[t]{1.5cm}{Method} & \parbox[t]{2.75cm}{\centering FS ($\downarrow$)} & \parbox[t]{1.75cm}{\centering PSNR ($\uparrow$)} & \parbox[t]{1.75cm}{\centering LPIPS ($\downarrow$)} & \parbox[t]{1.75cm}{\centering SSIM ($\uparrow$)} \\
\midrule
\multirow{20}{*}{MPGD} & \multirow{5}{*}{~Box Inpainting~
(64)} & RFJS (8) (ours) & $\mathbf{0.542} \scriptstyle \pm 0.08$ & $\mathbf{29.81} \scriptstyle \pm 1.44$ & $\mathbf{0.102} \scriptstyle \pm 0.02$ & $\mathbf{0.852} \scriptstyle \pm 0.02$ \\
 &  & GS (8) (ours) & $\underline{0.587} \scriptstyle \pm 0.10$ & $\underline{29.44} \scriptstyle \pm 1.75$ & $\underline{0.102} \scriptstyle \pm 0.02$ & $\underline{0.851}\scriptstyle \pm 0.02$ \\
 &  & RGG (0.5) & $0.609 \scriptstyle \pm 0.08$ & $29.24 \scriptstyle \pm 1.30$ & $0.103 \scriptstyle \pm 0.02$ & $0.850 \scriptstyle \pm 0.02$ \\
 &  & BON & $0.661 \scriptstyle \pm 0.08$ & $29.35 \scriptstyle \pm 1.82$ & $0.102 \scriptstyle \pm 0.02$ & $0.851 \scriptstyle \pm 0.02$ \\
 &  & MPGD & $0.766 \scriptstyle \pm 0.07$  & $29.09 \scriptstyle \pm 1.27$   & $0.103 \scriptstyle \pm 0.02$  & $0.848 \scriptstyle \pm 0.02$  \\
\cmidrule(lr){2-7}
 & \multirow{5}{*}{~Super Resolution~(6)} & RFJS (8) (ours) & $\mathbf{0.834} \scriptstyle \pm 0.09$ & $\mathbf{24.50} \scriptstyle \pm 1.48$ & $\mathbf{0.242} \scriptstyle \pm 0.04$ & $\mathbf{0.666} \scriptstyle \pm 0.06$ \\
 &  & GS (8) (ours) & $0.878 \scriptstyle \pm 0.08$ & $\underline{24.45} \scriptstyle \pm 1.47$ & $0.247 \scriptstyle \pm 0.03$ & $0.660 \scriptstyle \pm 0.06$ \\
 &  & RGG (0.25) & $\underline{0.854} \scriptstyle \pm 0.07$ & $24.39 \scriptstyle \pm 1.44$ & $0.246 \scriptstyle \pm 0.03$ & $0.656 \scriptstyle \pm 0.05$ \\
 &  & BON & $0.964 \scriptstyle \pm 0.09$ & $24.44 \scriptstyle \pm 1.58$ & $\underline{0.244} \scriptstyle \pm 0.04$ & $\underline{0.664} \scriptstyle \pm 0.06$ \\
 &  & MPGD & $1.037 \scriptstyle \pm 0.07$ & $24.39 \scriptstyle \pm 1.45$ & $0.249 \scriptstyle \pm 0.03$ & $0.657  \scriptstyle \pm 0.06$ \\
 \cmidrule(lr){2-7}
 & \multirow{5}{*}{~Gaussian Deblur~(5)} & RFJS (16) (ours) & $\underline{0.848}\scriptstyle \pm 0.07$ & $\underline{24.19} \scriptstyle \pm 1.40$ & $\mathbf{0.229} \scriptstyle \pm 0.03$ & $\underline{0.638} \scriptstyle \pm 0.06$  \\
 &  & GS (16) (ours) & $0.893 \scriptstyle \pm 0.07$ & $24.14 \scriptstyle \pm 1.39$ & $\underline{0.233} \scriptstyle \pm 0.03$ & $0.637 \scriptstyle \pm 0.06$ \\
 &  & RGG (0.25) & $\mathbf{0.846} \scriptstyle \pm 0.05$ & $24.11 \scriptstyle \pm 1.34$ & $0.235 \scriptstyle \pm 0.03$ & $0.634 \scriptstyle \pm 0.05$ \\
 &  & BON & $0.950 \scriptstyle \pm 0.07$ & $\mathbf{24.20} \scriptstyle \pm 1.38$ & $0.233 \scriptstyle \pm 0.03$ & $\mathbf{0.640} \scriptstyle \pm 0.06$ \\
 &  & MPGD & $1.026 \scriptstyle \pm 0.06$ & $24.09 \scriptstyle \pm 1.35 $ & $0.236 \scriptstyle \pm 0.03$ & $0.634 \scriptstyle \pm 0.06$ \\

 \bottomrule
\end{tabular}
}
\end{table}

{
\setlength{\tabcolsep}{1pt}
\renewcommand{\arraystretch}{1.0}
\begin{figure}[H]
  \centering
  \makebox[\textwidth]{
  \scalebox{0.51}{
  \begin{tabular}{c c}
  \begin{tabular}{>{\centering\arraybackslash}m{0.5cm} *{6}{>{\centering\arraybackslash}m{2.2cm}} >{\centering\arraybackslash}m{0.5cm} *{6}{>{\centering\arraybackslash}m{2.2cm}}}
    & \small Side & \small Measurement & \small MPGD & \small RGG & \small RFJS (Ours) & \small Ground Truth & & \small Side & \small Measurement & \small MPGD & \small RGG & \small RFJS (Ours) & \small Ground Truth \\
    {\centering\rotatebox{90}{\footnotesize Box Inpainting}}
    &
    \includegraphics[width=\linewidth]{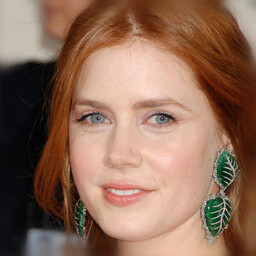} &
    \includegraphics[width=\linewidth]{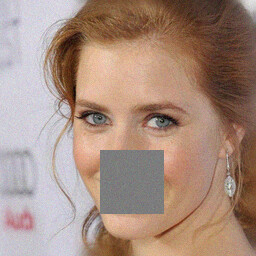} &
    \includegraphics[width=\linewidth]{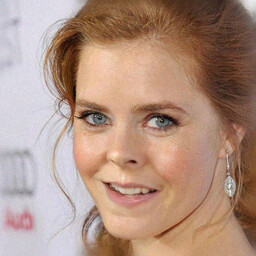} &
    \includegraphics[width=\linewidth]{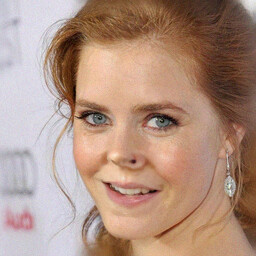} &
    \includegraphics[width=\linewidth]{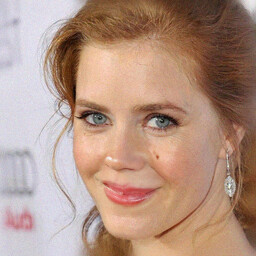} &
    \includegraphics[width=\linewidth]{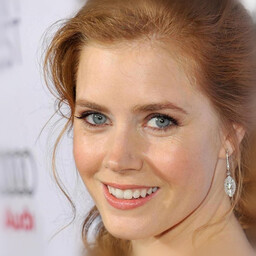}
     &
     {\centering\rotatebox{90}{\footnotesize Box Inpainting}}
    &
    \includegraphics[width=\linewidth]{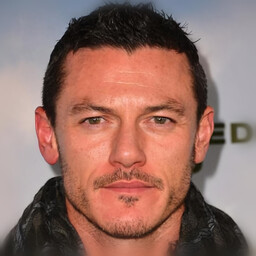} &
    \includegraphics[width=\linewidth]{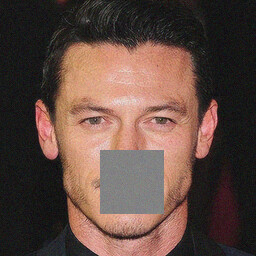} &
    \includegraphics[width=\linewidth]{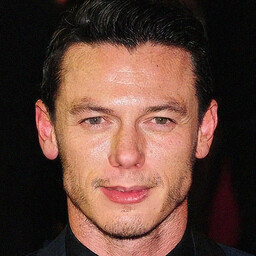} &
    \includegraphics[width=\linewidth]{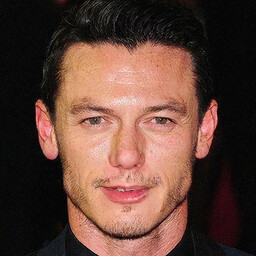} &
    \includegraphics[width=\linewidth]{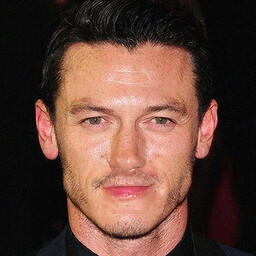} &
    \includegraphics[width=\linewidth]{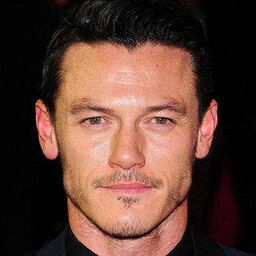} \\
    {\centering\rotatebox{90}{\footnotesize Super Resolution}}
     &
    \includegraphics[width=\linewidth]{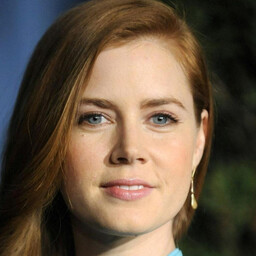} &
    \includegraphics[width=\linewidth]{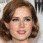} &
    \includegraphics[width=\linewidth]{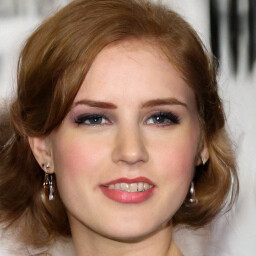} &
    \includegraphics[width=\linewidth]{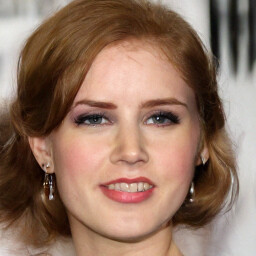} &
    \includegraphics[width=\linewidth]{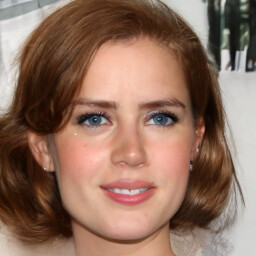} &
    \includegraphics[width=\linewidth]{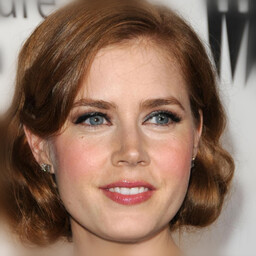}
     & {\centering\rotatebox{90}{\footnotesize Super Resolution}}
     &
    \includegraphics[width=\linewidth]{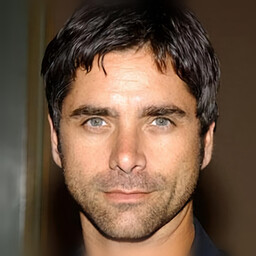} &
    \includegraphics[width=\linewidth]{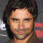} &
    \includegraphics[width=\linewidth]{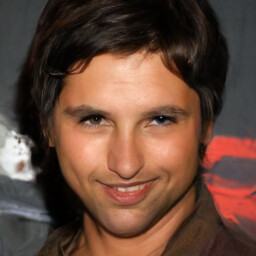} &
    \includegraphics[width=\linewidth]{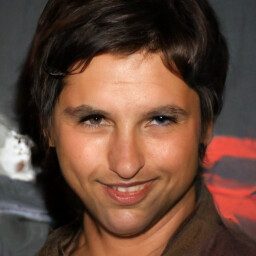} &
    \includegraphics[width=\linewidth]{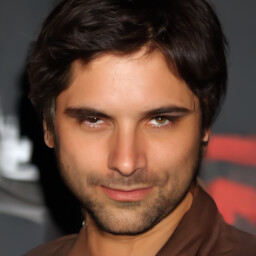} &
    \includegraphics[width=\linewidth]{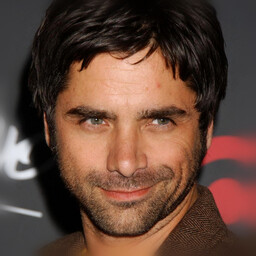} \\
    {\centering\rotatebox{90}{\footnotesize Gaussian Deblur}}
     &
    \includegraphics[width=\linewidth]{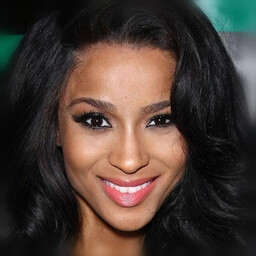} &
    \includegraphics[width=\linewidth]{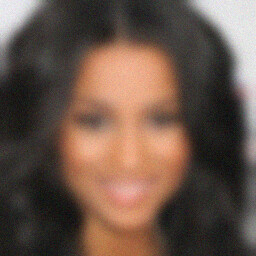} &
    \includegraphics[width=\linewidth]{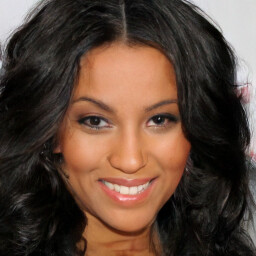} &
    \includegraphics[width=\linewidth]{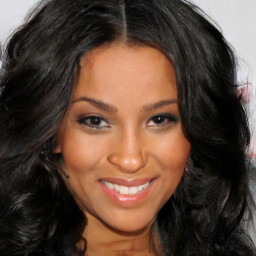} &
    \includegraphics[width=\linewidth]{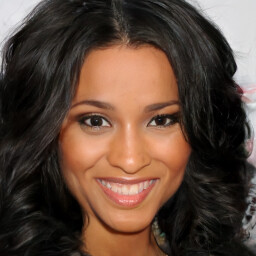} &
    \includegraphics[width=\linewidth]{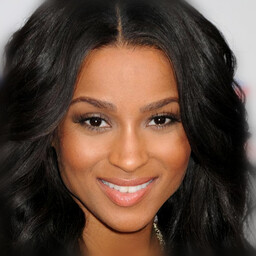}
     &
     {\centering\rotatebox{90}{\footnotesize Gaussian Deblur}}
     &
    \includegraphics[width=\linewidth]{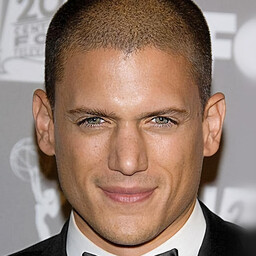} &
    \includegraphics[width=\linewidth]{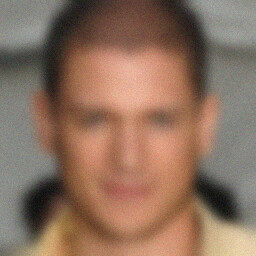} &
    \includegraphics[width=\linewidth]{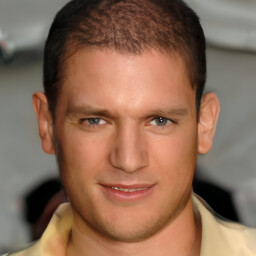} &
    \includegraphics[width=\linewidth]{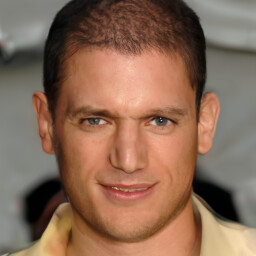} &
    \includegraphics[width=\linewidth]{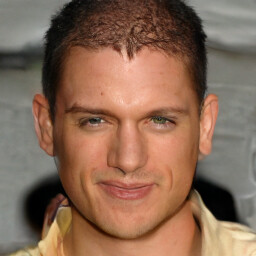} &
    \includegraphics[width=\linewidth]{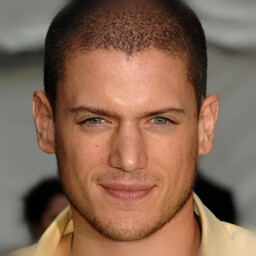} \\
\end{tabular}
  \end{tabular}
  }} % END \makebox and \scalebox
  \caption{Qualitative comparison of algorithms using MPGD as the base sampler. Our method offers better reconstructions aligned with the identity. Notice that while the identity is preserved, the exact reconstruction might not be possible as witnessed in, for example, Box inpainting. The ground truth has a smiling face whereas the reconstruction does not, although being the same person. Thus, PSNR improvements over the base sampler might be small, but FS improvements are significant.}
  \label{fig:qualitative_results_mpgd}
\end{figure}
}

\newpage

\subsection{Stable Diffusion}
\label{app:stable_diffusion}

To further demonstrate the generality of our framework, we evaluate it on Stable Diffusion, a large-scale latent diffusion model that natively conditions on text. A natural baseline in this setting is to simply pass the textual side information as the input prompt to Stable Diffusion and run the standard inverse problem solver (DAPS), which directly leverages the model's native text-conditioning mechanism without any modification. We compare this baseline against our inference-time search applied on top of the same text-conditioned solver, where particles are selected using a reward that scores consistency between reconstructions and the textual side information.

Even though Stable Diffusion already incorporates the text prompt through its native conditioning, we find that performing inference-time search over the resulting trajectories yields further improvements. This indicates that text-conditioning alone does not fully exploit the information contained in the side description: search allows the solver to explore multiple plausible reconstructions consistent with the prompt and select those that best align with the side information through the reward.

We additionally compare against a Best-of-$N$ variant that uses the same number of particles but selects the final reconstruction based on the measurement residual $\|\vy - \mA\vx_0^i\|$ rather than the side-information reward. This measurement-based selection strategy fails to improve, and in noisy or severely ill-posed regimes can actively degrade, the reconstruction quality, since it favors particles that explain the noise rather than the underlying signal. In contrast, our reward-based selection using side information consistently improves all metrics across box inpainting, super-resolution, motion deblur, and Gaussian deblur tasks.

\begin{figure}[H]
  \centering
  \makebox[\textwidth]{
  \scalebox{0.8}{
  \begin{tabular}{c c}
  \begin{tabular}{>{\centering\arraybackslash}m{0.5cm} *{4}{>{\centering\arraybackslash}m{2.2cm}}}
    & \small input & \small DAPS & \small Ours & \small label \\

    \multirow{3}{*}{\parbox[c][3cm][c]{0.5cm}{\centering\rotatebox{90}{\footnotesize Box Inpainting}}} &
    \includegraphics[width=\linewidth]{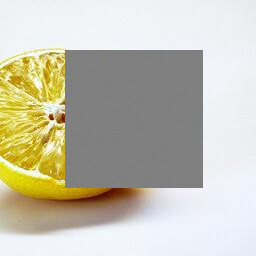} &
    \includegraphics[width=\linewidth]{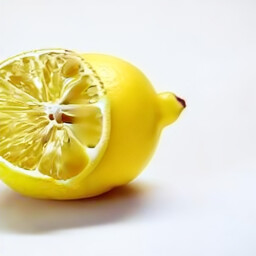} &
    \includegraphics[width=\linewidth]{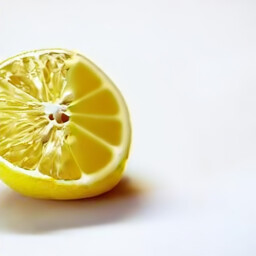} &
    \includegraphics[width=\linewidth]{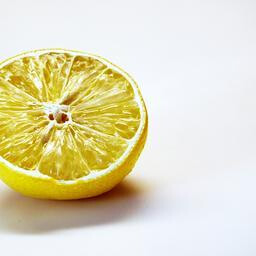}
    \\
    &
    \includegraphics[width=\linewidth]{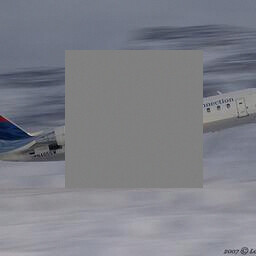} &
    \includegraphics[width=\linewidth]{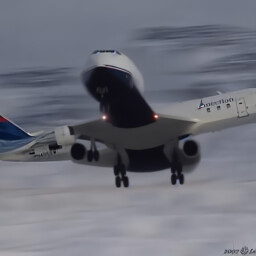} &
    \includegraphics[width=\linewidth]{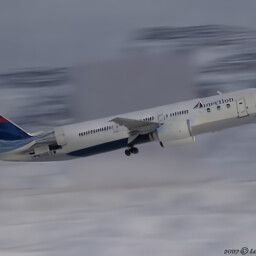} &
    \includegraphics[width=\linewidth]{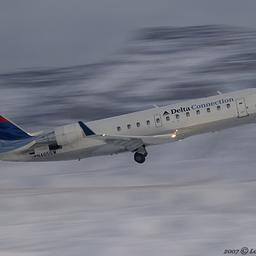}
    \\
    \multirow{4}{*}{\parbox[c][4cm][c]{0.5cm}{\centering\rotatebox{90}{\footnotesize Super Resolution}}} &
    \includegraphics[width=\linewidth]{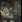} &
    \includegraphics[width=\linewidth]{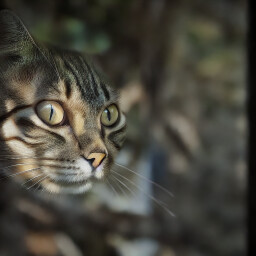} &
    \includegraphics[width=\linewidth]{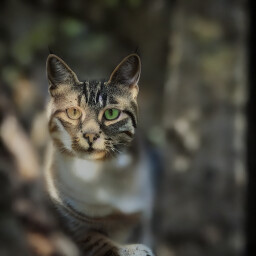} &
    \includegraphics[width=\linewidth]{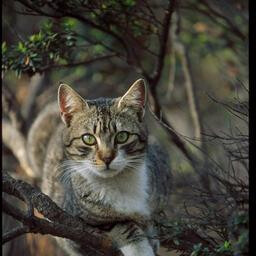}
    \\ &
    \includegraphics[width=\linewidth]{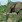} &
    \includegraphics[width=\linewidth]{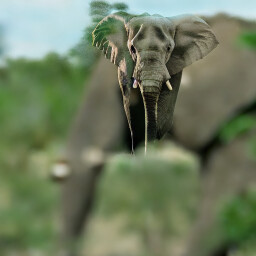} &
    \includegraphics[width=\linewidth]{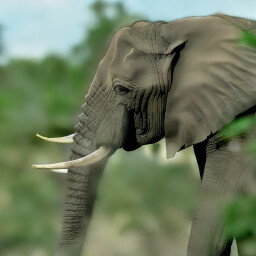} &
    \includegraphics[width=\linewidth]{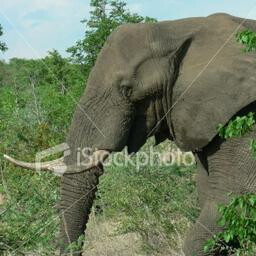}
    \\
    \multirow{2}{*}{\parbox[c][4cm][c]{0.5cm}{\centering\rotatebox{90}{\footnotesize Gaussian Deblur}}}
    &
    \includegraphics[width=\linewidth]{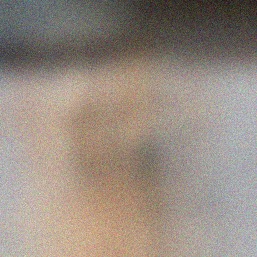} &
    \includegraphics[width=\linewidth]{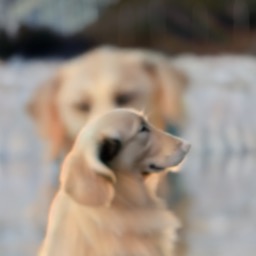} &
    \includegraphics[width=\linewidth]{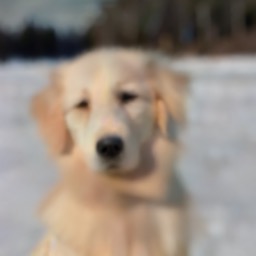} &
    \includegraphics[width=\linewidth]{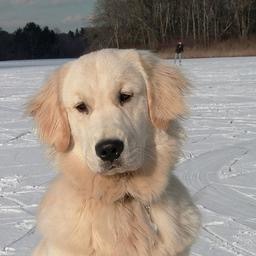}
    \\ &
    \includegraphics[width=\linewidth]{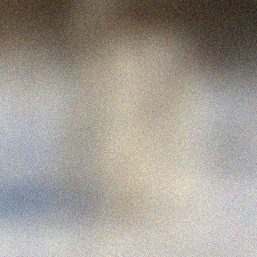} &
    \includegraphics[width=\linewidth]{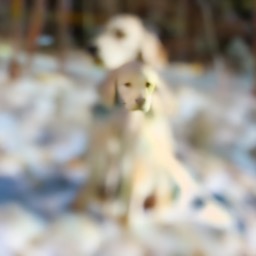} &
    \includegraphics[width=\linewidth]{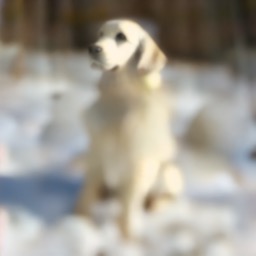} &
    \includegraphics[width=\linewidth]{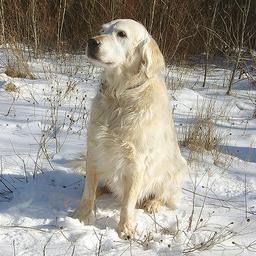}
    \\
    \multirow{3}{*}{\parbox[c][4cm][c]{0.5cm}{\centering\rotatebox{90}{\footnotesize Gaussian Deblur}}}
    &
    \includegraphics[width=\linewidth]{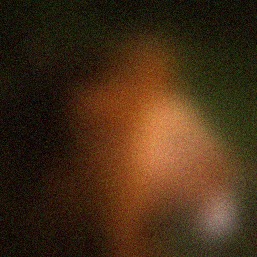} &
    \includegraphics[width=\linewidth]{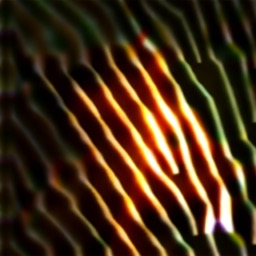} &
    \includegraphics[width=\linewidth]{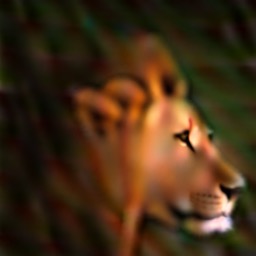} &
    \includegraphics[width=\linewidth]{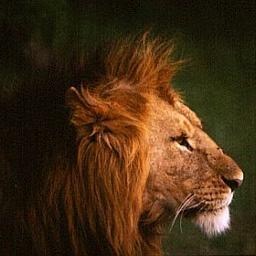}
    \\ &
    \includegraphics[width=\linewidth]{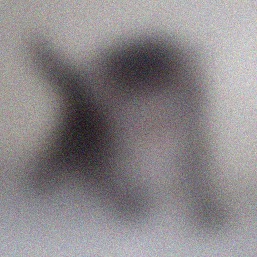} &
    \includegraphics[width=\linewidth]{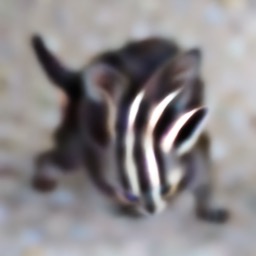} &
    \includegraphics[width=\linewidth]{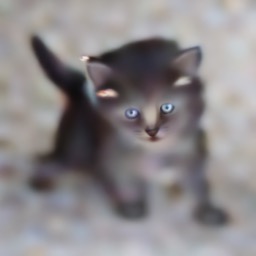} &
    \includegraphics[width=\linewidth]{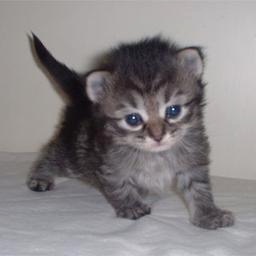}
    \\
    \end{tabular}

  \end{tabular}
  }} % END \makebox and \scalebox
  \caption{Stable Diffusion}
  \label{fig:qualitative_results_DAPS_StableDiffusion}
\end{figure}

Qualitative reconstructions are shown in Figure~\ref{fig:qualitative_results_DAPS_StableDiffusion}, where our method recovers perceptually faithful content that aligns with the textual description, while the DAPS baseline often produces artifacts or content inconsistent with the prompt. Quantitative results are reported in Table~\ref{tab:bon_si_ablation}, which confirms that selection by our reward $r(\vx_0; \vs)$ yields consistent gains over both the base sampler and Best-of-$N$ without side information. Together, these results show that our framework is complementary to native conditioning mechanisms and provides additional gains even when the base model is already trained to consume the same modality of side information.

\begin{table}[H]
\centering
\footnotesize
\def\tablescale{0.9}
\resizebox{\tablescale\linewidth}{!}{%
\begin{tabular}{l|cccc|cccc}
\toprule
& \multicolumn{4}{c|}{Box Inpainting} & \multicolumn{4}{c}{Gaussian Deblur} \\
\midrule
Algorithm & PSNR ($\uparrow$) & LPIPS ($\downarrow$) & SSIM ($\uparrow$) & CS ($\uparrow$) & PSNR ($\uparrow$) & LPIPS ($\downarrow$) & SSIM ($\uparrow$) & CS ($\uparrow$) \\
\midrule
BON (w/ SI, ours) & $\mathbf{19.51}$ & $\mathbf{0.262}$ & $\mathbf{0.750}$ & $\mathbf{0.893}$ & $\mathbf{20.30}$ & $\mathbf{0.527}$ & $\mathbf{0.494}$ & $\mathbf{0.789}$ \\
BON (w/o SI) & $18.02$ & $0.274$ & $0.738$ & $0.846$ & $\underline{20.05}$ & $\underline{0.540}$ & $\underline{0.487}$ & $\underline{0.748}$ \\
DAPS & $\underline{18.21}$ & $\underline{0.274}$ & $\underline{0.742}$ & $\underline{0.855}$ & $19.86$ & $0.540$ & $0.481$ & $0.735$ \\
\midrule
\midrule
& \multicolumn{4}{c|}{Super Resolution} & \multicolumn{4}{c}{Motion Deblur} \\
\midrule
Algorithm & PSNR ($\uparrow$) & LPIPS ($\downarrow$) & SSIM ($\uparrow$) & CS ($\uparrow$) & PSNR ($\uparrow$) & LPIPS ($\downarrow$) & SSIM ($\uparrow$) & CS ($\uparrow$) \\
\midrule
BON (w/ SI, ours) & $\mathbf{18.65}$ & $\mathbf{0.579}$ & $\mathbf{0.398}$ & $\mathbf{0.767}$ & $\mathbf{19.54}$ & $\mathbf{0.551}$ & $\mathbf{0.474}$ & $\mathbf{0.770}$ \\
BON (w/o SI) & $18.13$ & $0.590$ & $0.368$ & $\underline{0.729}$ & $18.18$ & $0.562$ & $0.462$ & $0.718$ \\
DAPS & $\underline{18.44}$ & $\underline{0.583}$ & $\underline{0.391}$ & $0.727$ & $\underline{19.49}$ & $\underline{0.556}$ & $\underline{0.466}$ & $\underline{0.722}$ \\
\bottomrule
\end{tabular}
}
\caption{Without side information, Best-of-$N$ selection by $\|\vy - \mA\vx_0^i\|$
fails to improve over the baseline and can degrade it. With side information,
selection by our reward $r(\vx_0;\vs)$ consistently improves all metrics (Box Inpainting and Super Resolution are reproduced from the table provided in the section Role of side information in the main paper).}
\label{tab:bon_si_ablation}
\end{table}

{% \color{blue}

\subsection{Ablation Studies}

In this section, we conduct ablation studies to study the performance and robustness of our algorithm by degrading: (i) Measurement quality and (ii) Side information quality.

\subsubsection{Measurement Quality: Hard Inverse Problems}

\label{app: measurement_quality}

In this section, we demonstrate the effectiveness of our method on hard inverse problems with severely degraded measurements. For DPS, we evaluate inpainting with a large mask that covers almost the entire face, super-resolution with aggressive downsampling ratios of 12× and 32×, and motion deblurring using a large 256-pixel kernel. For MPGD, we similarly consider challenging variants of inpainting, super-resolution, and Gaussian deblurring, with quantitative results reported in Tables~\ref{tab:dps_hard_tasks} and~\ref{tab:mpgd_hard_tasks}. Across all settings, our method consistently improves the target reward metric, which translates directly into stronger qualitative performance, and we also observe improvements in classical metrics on average.
The qualitative results are shown in Figures~\ref{fig:dps_hard_tasks_mb_256_sr_12x}, ~\ref{fig:dps_hard_tasks_box_ip}, ~\ref{fig:dps_hard_tasks_sr_32x}.
For textual side information, we already include challenging experimental settings in the main paper; see Figure~\ref{fig:qualitative_results_text_reward} and Table~\ref{tab:quantitative_results_text_reward}.

\begin{figure}[H]
  \centering
  \makebox[\textwidth]{
  \scalebox{0.75}{
  \begin{tabular}{c}
  \begin{tabular}{>{\centering\arraybackslash}m{0.5cm} *{6}{>{\centering\arraybackslash}m{2.2cm}}}
    & \small Measurement & \small Ground Truth & \small RFJS (Ours) & \small DPS(1) & \small DPS(2)  & \small DPS(3) \\

    {\centering\rotatebox{90}{\small Motion Deblur}} &
    \includegraphics[width=\linewidth]{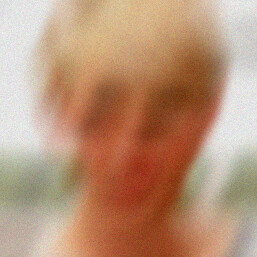} &
    \includegraphics[width=\linewidth]{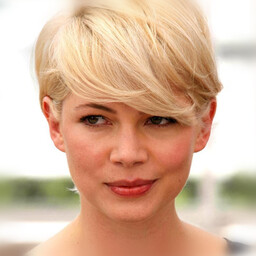} &
    \includegraphics[width=\linewidth]{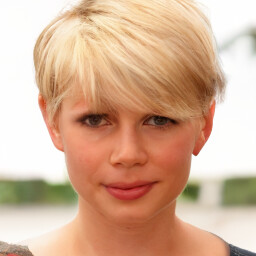} &
    \includegraphics[width=\linewidth]{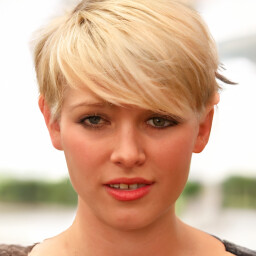} &
    \includegraphics[width=\linewidth]{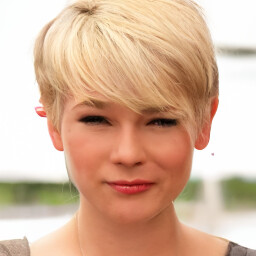} &
    \includegraphics[width=\linewidth]{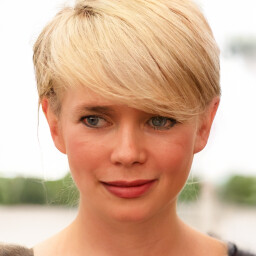}\\ [-2pt]

    {\centering\rotatebox{90}{\small Motion Deblur}} &
    \includegraphics[width=\linewidth]{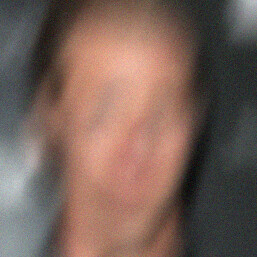} &
    \includegraphics[width=\linewidth]{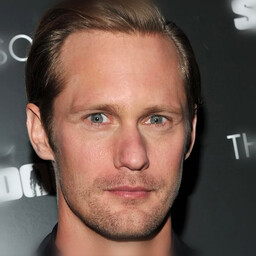} &
    \includegraphics[width=\linewidth]{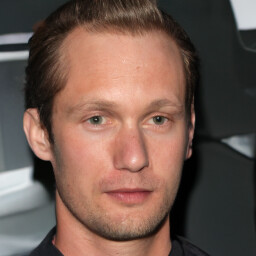} &
    \includegraphics[width=\linewidth]{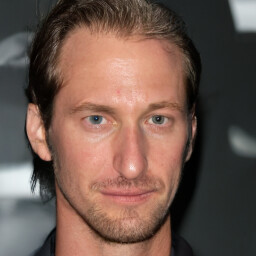} &
    \includegraphics[width=\linewidth]{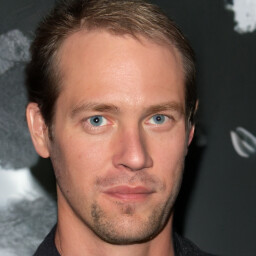} &
    \includegraphics[width=\linewidth]{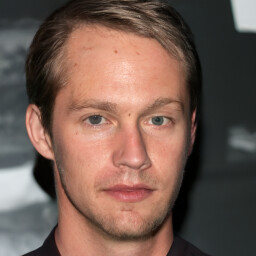}\\ [-2pt]

     {\centering\rotatebox{90}{\small Motion Deblur}} &
    \includegraphics[width=\linewidth]{pictures/hard_problems/DPS/motion/41inp.jpg} &
    \includegraphics[width=\linewidth]{pictures/hard_problems/DPS/motion/41gt.jpg} &
    \includegraphics[width=\linewidth]{pictures/hard_problems/DPS/motion/41s.jpg} &
    \includegraphics[width=\linewidth]{pictures/hard_problems/DPS/motion/41d3.jpg} &
    \includegraphics[width=\linewidth]{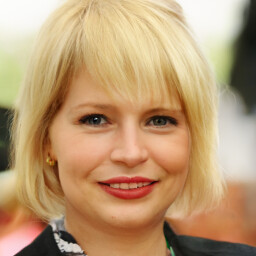} &
    \includegraphics[width=\linewidth]{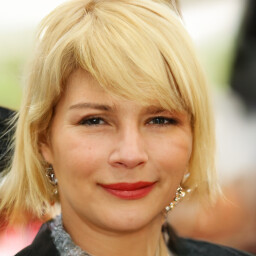}\\ [-2pt]

    {\centering\rotatebox{90}{\small Motion Deblur}} &
    \includegraphics[width=\linewidth]{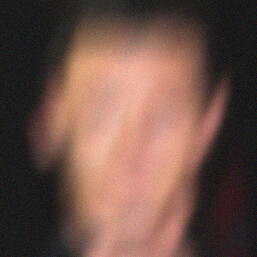} &
    \includegraphics[width=\linewidth]{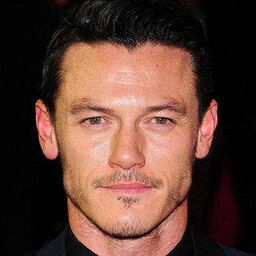} &
    \includegraphics[width=\linewidth]{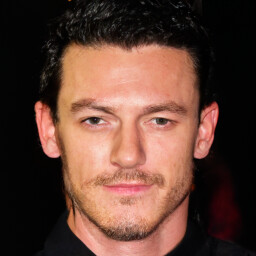} &
    \includegraphics[width=\linewidth]{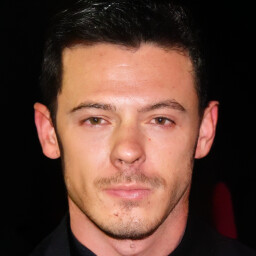} &
    \includegraphics[width=\linewidth]{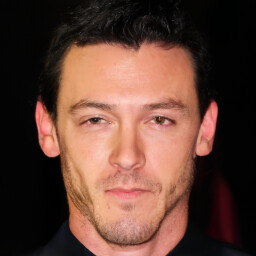} &
    \includegraphics[width=\linewidth]{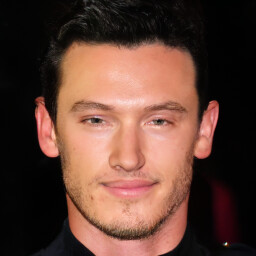}\\ [-2pt]

    {\centering\rotatebox{90}{\small Motion Deblur}} &
    \includegraphics[width=\linewidth]{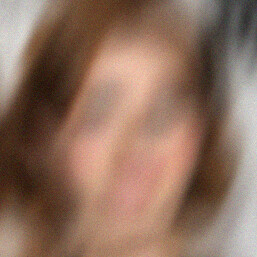} &
    \includegraphics[width=\linewidth]{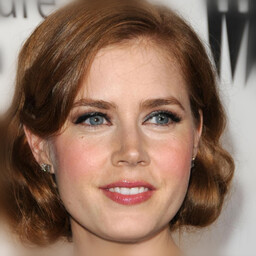} &
    \includegraphics[width=\linewidth]{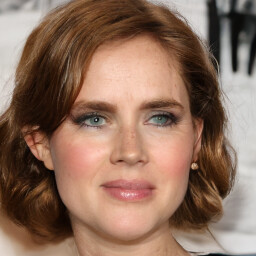} &
    \includegraphics[width=\linewidth]{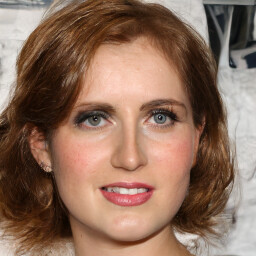} &
    \includegraphics[width=\linewidth]{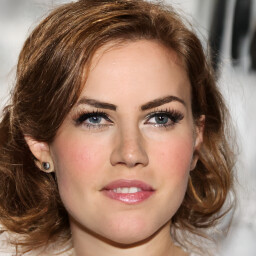} &
    \includegraphics[width=\linewidth]{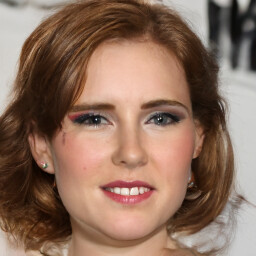}\\ [-2pt]

    {\centering\rotatebox{90}{\small Motion Deblur}} &
    \includegraphics[width=\linewidth]{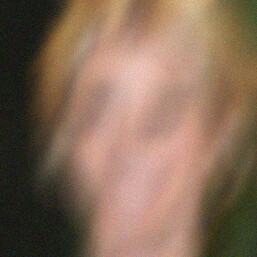} &
    \includegraphics[width=\linewidth]{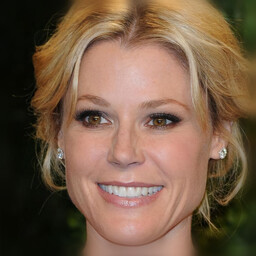} &
    \includegraphics[width=\linewidth]{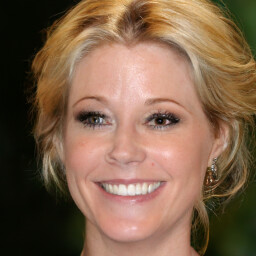} &
    \includegraphics[width=\linewidth]{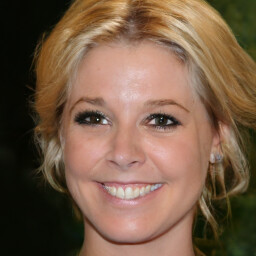} &
    \includegraphics[width=\linewidth]{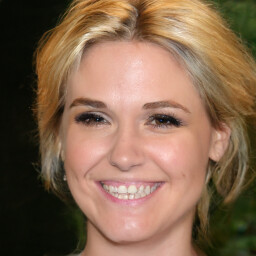} &
    \includegraphics[width=\linewidth]{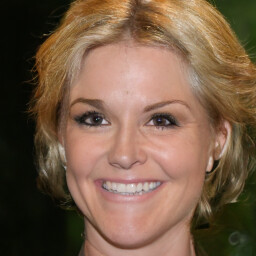}\\ [-2pt]

    {\centering\rotatebox{90}{\small SR(12)}} &
    \includegraphics[width=\linewidth]{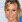} &
    \includegraphics[width=\linewidth]{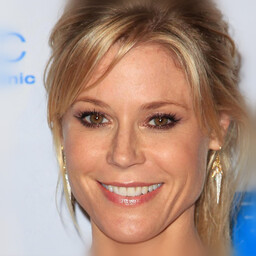} &
    \includegraphics[width=\linewidth]{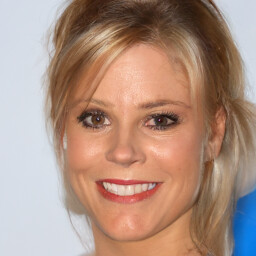} &
    \includegraphics[width=\linewidth]{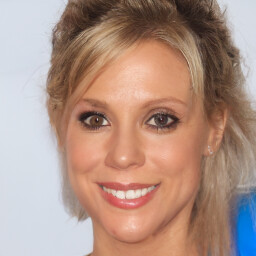} &
    \includegraphics[width=\linewidth]{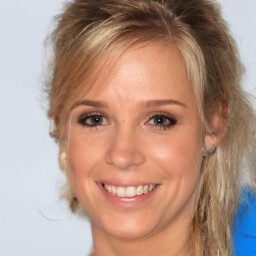} &
    \includegraphics[width=\linewidth]{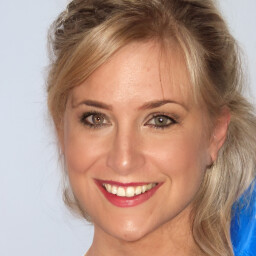}\\ [-2pt]

    {\centering\rotatebox{90}{\small SR(12)}} &
    \includegraphics[width=\linewidth]{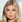} &
    \includegraphics[width=\linewidth]{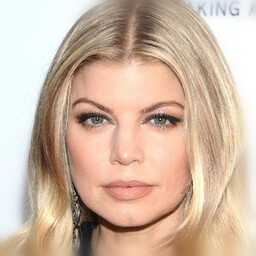} &
    \includegraphics[width=\linewidth]{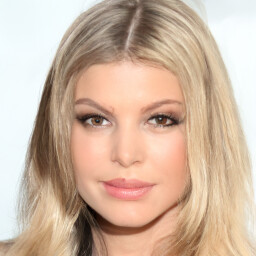} &
    \includegraphics[width=\linewidth]{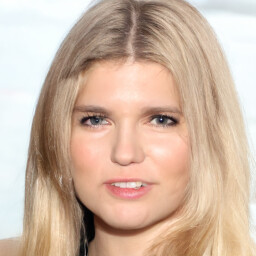} &
    \includegraphics[width=\linewidth]{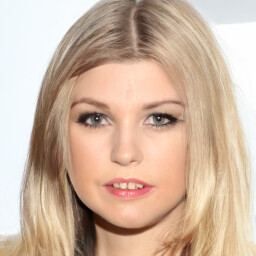} &
    \includegraphics[width=\linewidth]{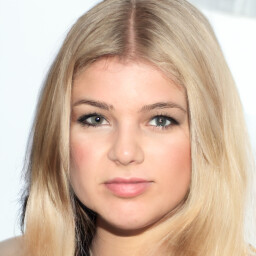}\\ [-2pt]

    {\centering\rotatebox{90}{\small SR(12)}} &
    \includegraphics[width=\linewidth]{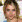} &
    \includegraphics[width=\linewidth]{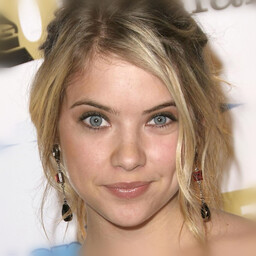} &
    \includegraphics[width=\linewidth]{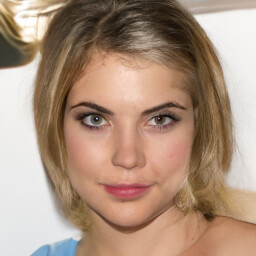} &
    \includegraphics[width=\linewidth]{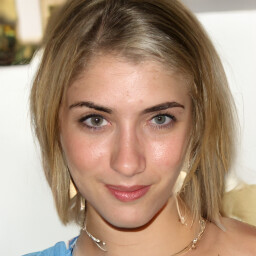} &
    \includegraphics[width=\linewidth]{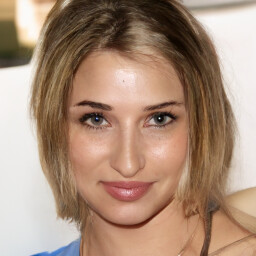} &
    \includegraphics[width=\linewidth]{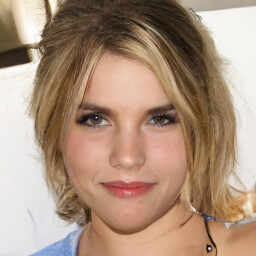}\\ [-2pt]

    {\centering\rotatebox{90}{\small SR(12)}} &
    \includegraphics[width=\linewidth]{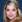} &
    \includegraphics[width=\linewidth]{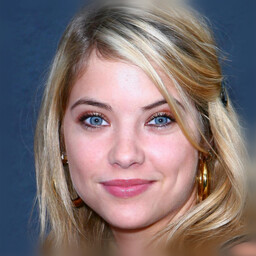} &
    \includegraphics[width=\linewidth]{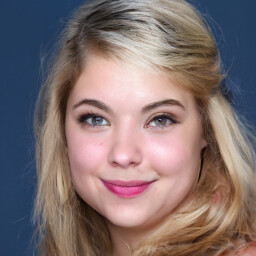} &
    \includegraphics[width=\linewidth]{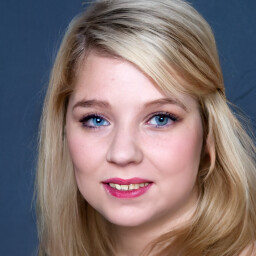} &
    \includegraphics[width=\linewidth]{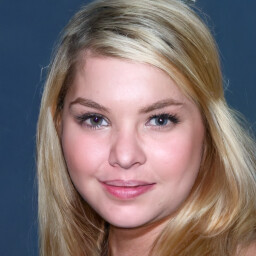} &
    \includegraphics[width=\linewidth]{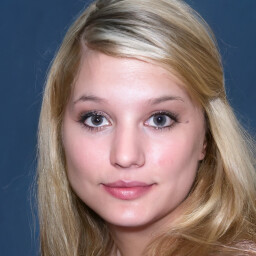}\\ [-2pt]

    {\centering\rotatebox{90}{\small SR(12)}} &
    \includegraphics[width=\linewidth]{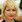} &
    \includegraphics[width=\linewidth]{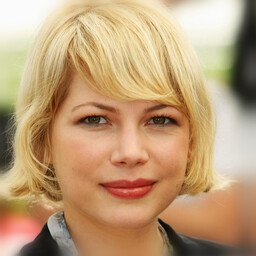} &
    \includegraphics[width=\linewidth]{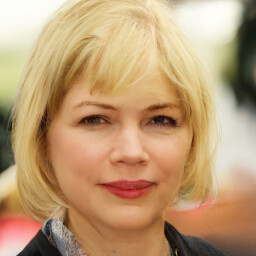} &
    \includegraphics[width=\linewidth]{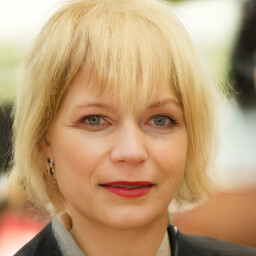} &
    \includegraphics[width=\linewidth]{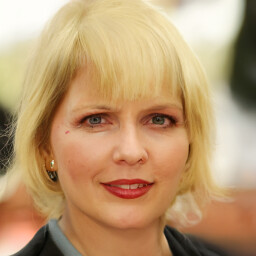} &
    \includegraphics[width=\linewidth]{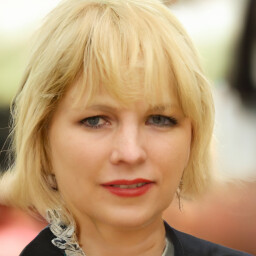}\\ [-2pt]

    {\centering\rotatebox{90}{\small SR(12)}} &
    \includegraphics[width=\linewidth]{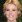} &
    \includegraphics[width=\linewidth]{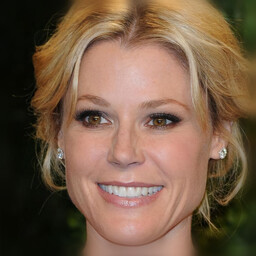} &
    \includegraphics[width=\linewidth]{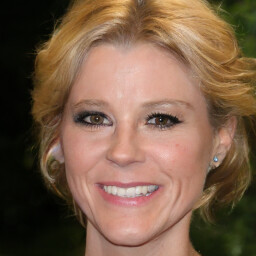} &
    \includegraphics[width=\linewidth]{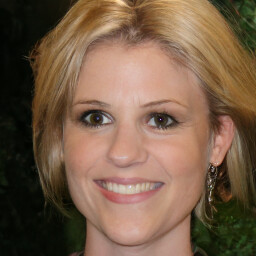} &
    \includegraphics[width=\linewidth]{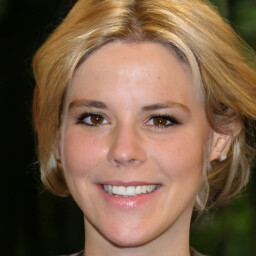} &
    \includegraphics[width=\linewidth]{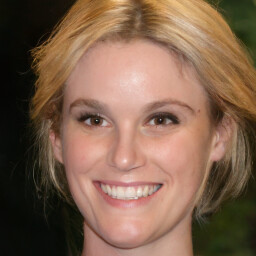}\\ [-2pt]

    {\centering\rotatebox{90}{\small SR(12)}} &
    \includegraphics[width=\linewidth]{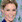} &
    \includegraphics[width=\linewidth]{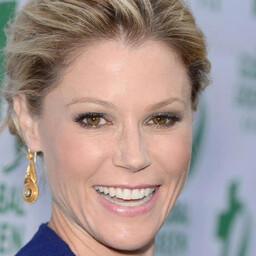} &
    \includegraphics[width=\linewidth]{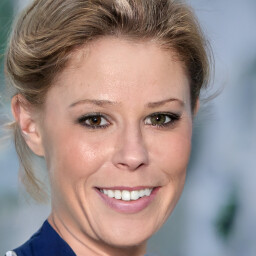} &
    \includegraphics[width=\linewidth]{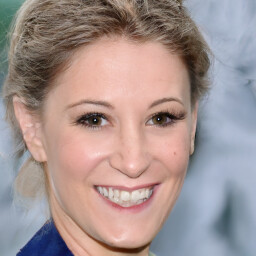} &
    \includegraphics[width=\linewidth]{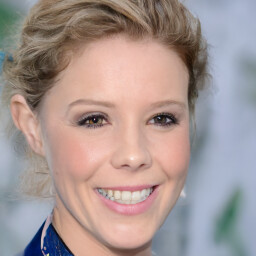} &
    \includegraphics[width=\linewidth]{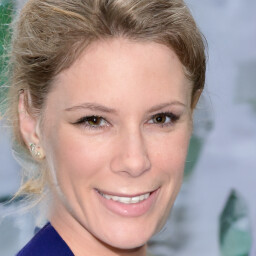}\\ [-2pt]

  \end{tabular}
  \end{tabular}
  }} % END \makebox and \scalebox
  \caption{Qualitative results on hard tasks using DPS. Our algorithms reconstructs faces with perceptual quality.}
  \label{fig:dps_hard_tasks_mb_256_sr_12x}
\end{figure}

\begin{figure}[H]
  \centering
  \makebox[\textwidth]{
  \scalebox{0.75}{
  \begin{tabular}{c}
  \begin{tabular}{>{\centering\arraybackslash}m{0.5cm} *{6}{>{\centering\arraybackslash}m{2.2cm}}}
    & \small Measurement & \small Ground Truth & \small RFJS(Ours) & \small DPS1 & \small DPS2  & \small DPS3 \\

    {\centering\rotatebox{90}{\small Box}} &
    \includegraphics[width=\linewidth]{pictures/hard_problems/DPS/box/47inp.jpg} &
    \includegraphics[width=\linewidth]{pictures/hard_problems/DPS/box/47gt.jpg} &
    \includegraphics[width=\linewidth]{pictures/hard_problems/DPS/box/47s.jpg} &
    \includegraphics[width=\linewidth]{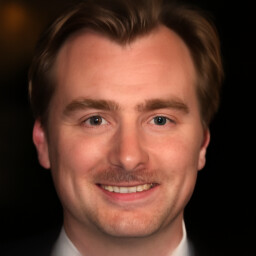} &
    \includegraphics[width=\linewidth]{pictures/hard_problems/DPS/box/47d2.jpg} &
    \includegraphics[width=\linewidth]{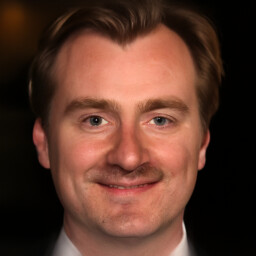}\\ [-2pt]

    {\centering\rotatebox{90}{\small Box}} &
    \includegraphics[width=\linewidth]{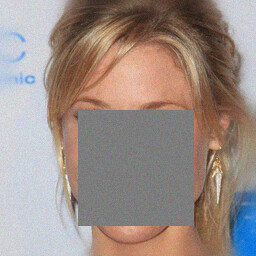} &
    \includegraphics[width=\linewidth]{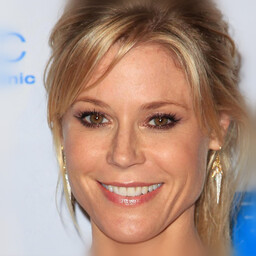} &
    \includegraphics[width=\linewidth]{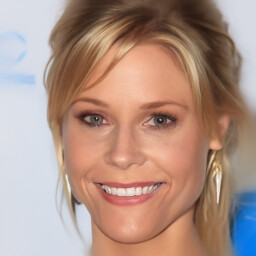} &
    \includegraphics[width=\linewidth]{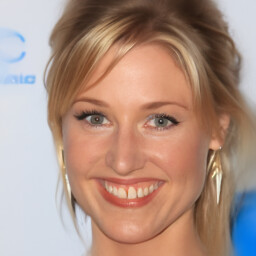} &
    \includegraphics[width=\linewidth]{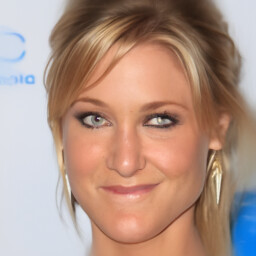} &
    \includegraphics[width=\linewidth]{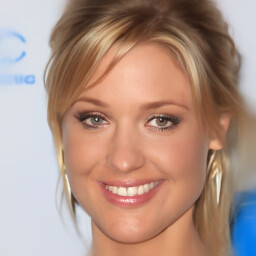}\\ [-2pt]

        {\centering\rotatebox{90}{\small Box}} &
    \includegraphics[width=\linewidth]{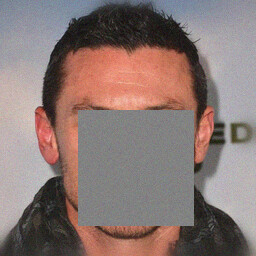} &
    \includegraphics[width=\linewidth]{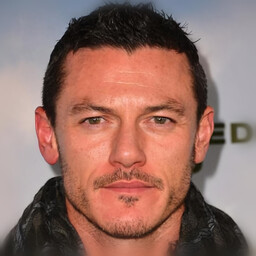} &
    \includegraphics[width=\linewidth]{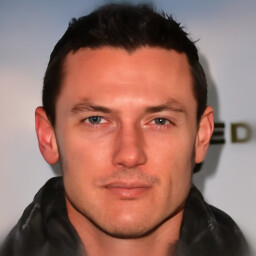} &
    \includegraphics[width=\linewidth]{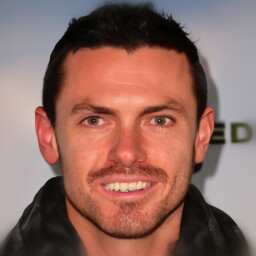} &
    \includegraphics[width=\linewidth]{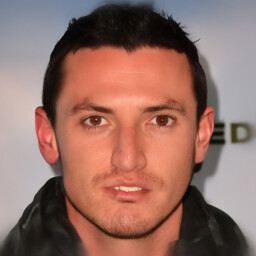} &
    \includegraphics[width=\linewidth]{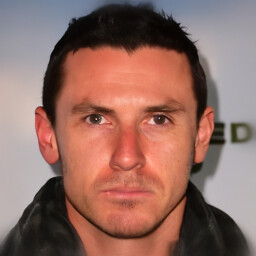}\\ [-2pt]

    {\centering\rotatebox{90}{\small Box}} &
    \includegraphics[width=\linewidth]{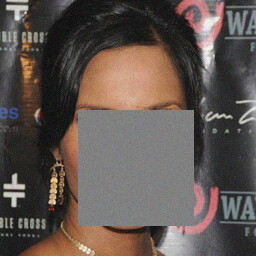} &
    \includegraphics[width=\linewidth]{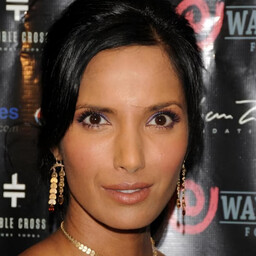} &
    \includegraphics[width=\linewidth]{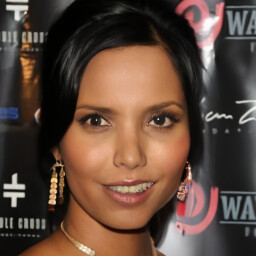} &
    \includegraphics[width=\linewidth]{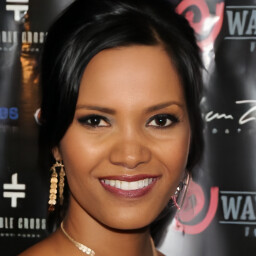} &
    \includegraphics[width=\linewidth]{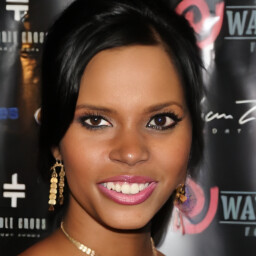} &
    \includegraphics[width=\linewidth]{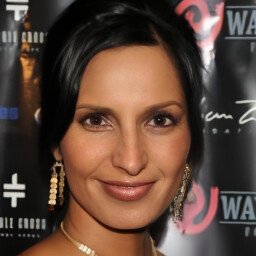}\\ [-2pt]

    {\centering\rotatebox{90}{\small Box}} &
    \includegraphics[width=\linewidth]{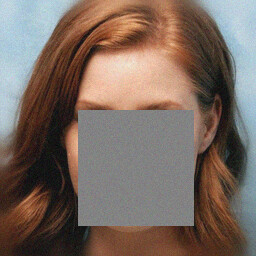} &
    \includegraphics[width=\linewidth]{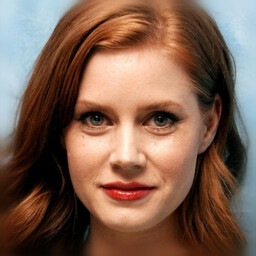} &
    \includegraphics[width=\linewidth]{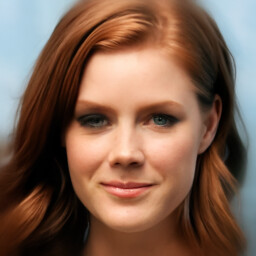} &
    \includegraphics[width=\linewidth]{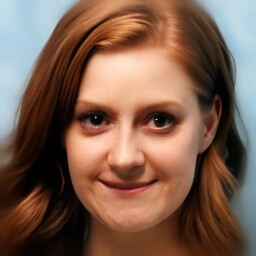} &
    \includegraphics[width=\linewidth]{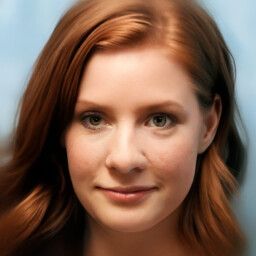} &
    \includegraphics[width=\linewidth]{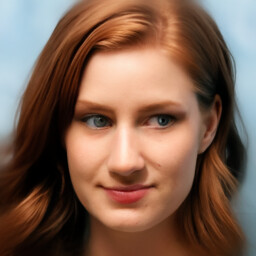}\\ [-2pt]

    {\centering\rotatebox{90}{\small Box}} &
    \includegraphics[width=\linewidth]{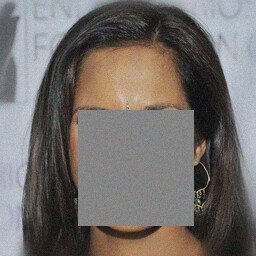} &
    \includegraphics[width=\linewidth]{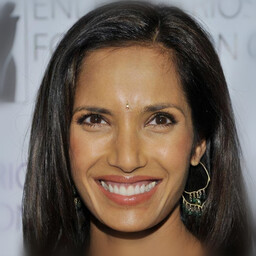} &
    \includegraphics[width=\linewidth]{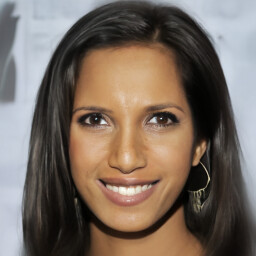} &
    \includegraphics[width=\linewidth]{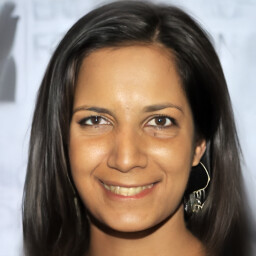} &
    \includegraphics[width=\linewidth]{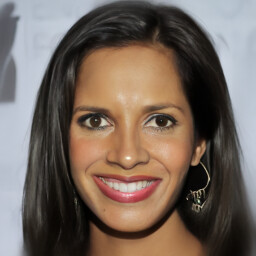} &
    \includegraphics[width=\linewidth]{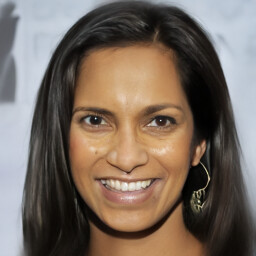}\\ [-2pt]

    {\centering\rotatebox{90}{\small Box}} &
    \includegraphics[width=\linewidth]{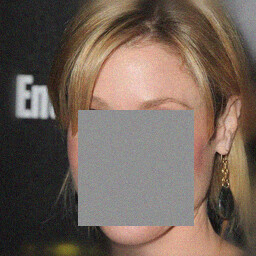} &
    \includegraphics[width=\linewidth]{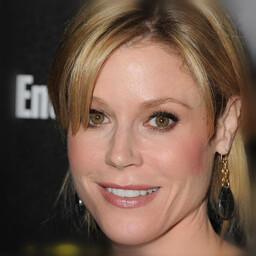} &
    \includegraphics[width=\linewidth]{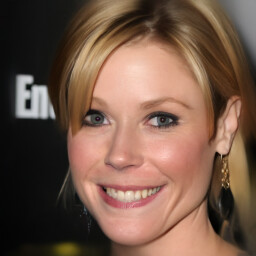} &
    \includegraphics[width=\linewidth]{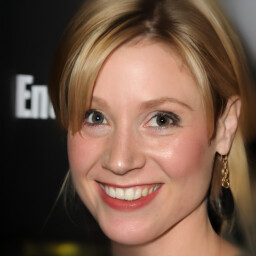} &
    \includegraphics[width=\linewidth]{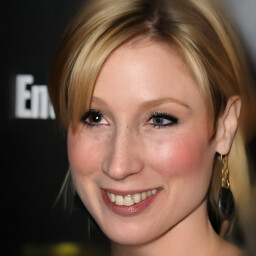} &
    \includegraphics[width=\linewidth]{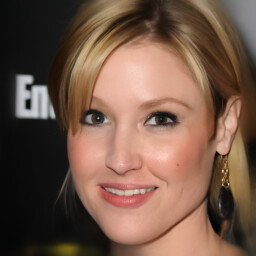}\\ [-2pt]

        {\centering\rotatebox{90}{\small Box}} &
    \includegraphics[width=\linewidth]{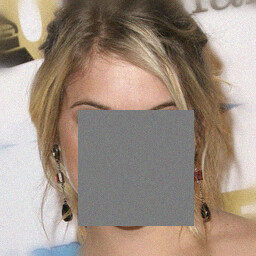} &
    \includegraphics[width=\linewidth]{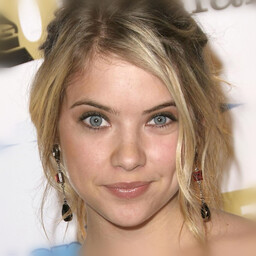} &
    \includegraphics[width=\linewidth]{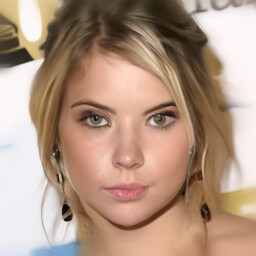} &
    \includegraphics[width=\linewidth]{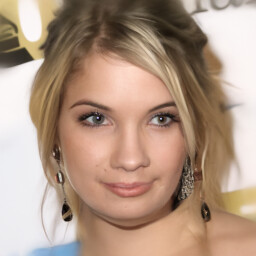} &
    \includegraphics[width=\linewidth]{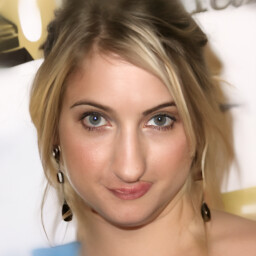} &
    \includegraphics[width=\linewidth]{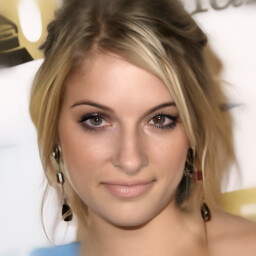}\\ [-2pt]

        {\centering\rotatebox{90}{\small Box}} &
    \includegraphics[width=\linewidth]{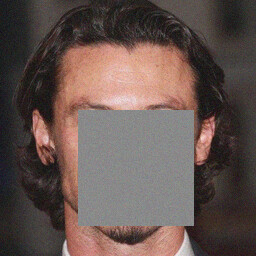} &
    \includegraphics[width=\linewidth]{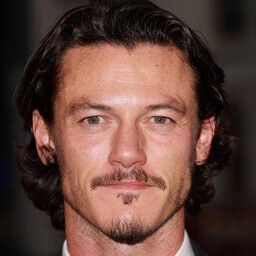} &
    \includegraphics[width=\linewidth]{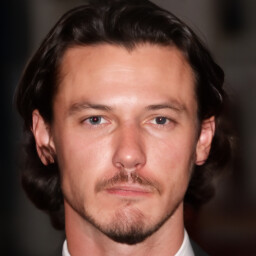} &
    \includegraphics[width=\linewidth]{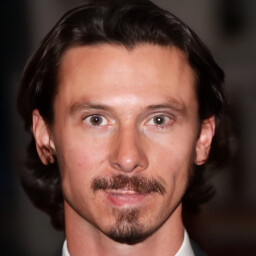} &
    \includegraphics[width=\linewidth]{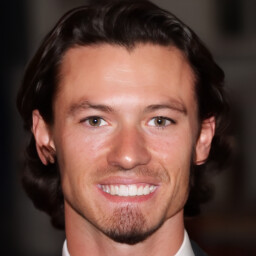} &
    \includegraphics[width=\linewidth]{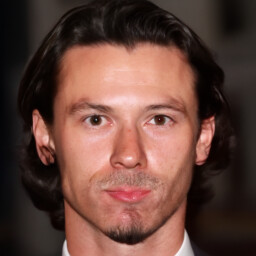}\\ [-2pt]

        {\centering\rotatebox{90}{\small Box}} &
    \includegraphics[width=\linewidth]{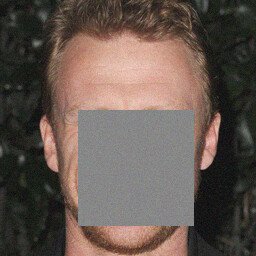} &
    \includegraphics[width=\linewidth]{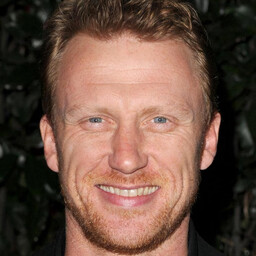} &
    \includegraphics[width=\linewidth]{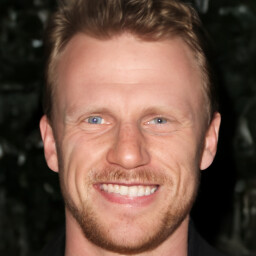} &
    \includegraphics[width=\linewidth]{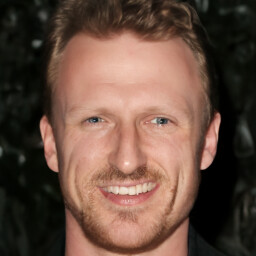} &
    \includegraphics[width=\linewidth]{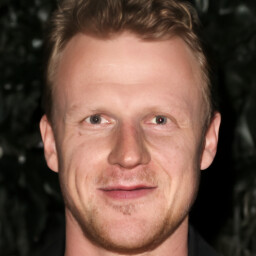} &
    \includegraphics[width=\linewidth]{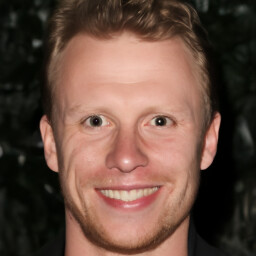}\\ [-2pt]

        {\centering\rotatebox{90}{\small Box}} &
    \includegraphics[width=\linewidth]{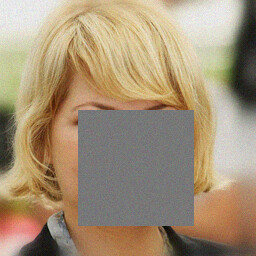} &
    \includegraphics[width=\linewidth]{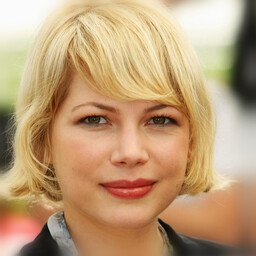} &
    \includegraphics[width=\linewidth]{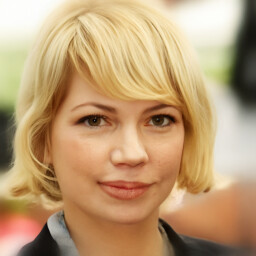} &
    \includegraphics[width=\linewidth]{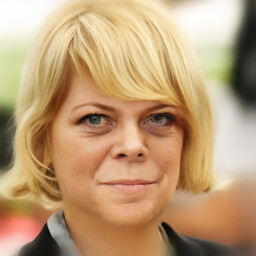} &
    \includegraphics[width=\linewidth]{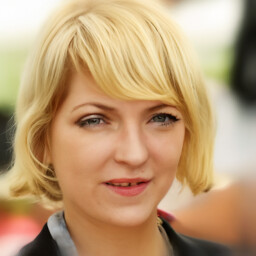} &
    \includegraphics[width=\linewidth]{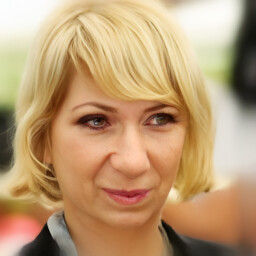}\\ [-2pt]

        {\centering\rotatebox{90}{\small Box}} &
    \includegraphics[width=\linewidth]{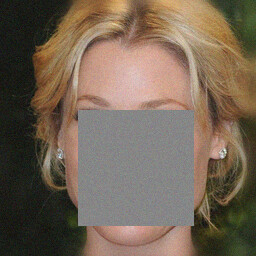} &
    \includegraphics[width=\linewidth]{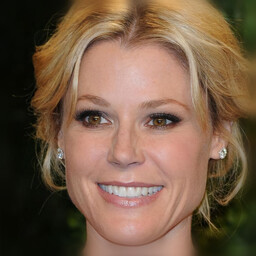} &
    \includegraphics[width=\linewidth]{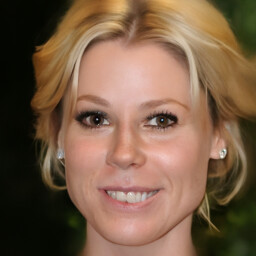} &
    \includegraphics[width=\linewidth]{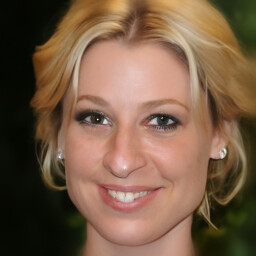} &
    \includegraphics[width=\linewidth]{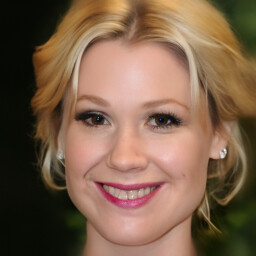} &
    \includegraphics[width=\linewidth]{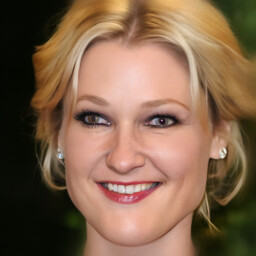}\\ [-2pt]

  \end{tabular}
  \end{tabular}
  }} % END \makebox and \scalebox
  \caption{Qualitative results on box inpainting using DPS sampler. Since the face is completely masked in the measurement, side information provides useful clues so that the reconstruction preserves identity.}
  \label{fig:dps_hard_tasks_box_ip}
\end{figure}

{\setlength{\tabcolsep}{6pt}  % default is 6pt
\begin{table}[ht]
\caption{Quantitative results using DPS for hard tasks with ${\rm dps~scale} = 0.8$ (guidance strength from the measurement $y$)}
    \label{tab:dps_hard_tasks}
    \centering
    \begin{tabular}{|c|c|c|c|c|c|c|}
    \toprule
       Task Name & Task Parameters & Algo & FS ($\downarrow$) & PSNR  ($\uparrow$) & LPIPS  ($\downarrow$) & SSIM ($\uparrow$) \\
       \midrule
       \multirow{2}{*}{Box Inpainting}
        & \multirow{2}{*}{$M=116$}
        & RFJS & \textbf{0.47} & \textbf{26.62} & \textbf{0.140} & \textbf{0.839} \\
        & & DPS & 1.07 & 26.21 & 0.151 & 0.826 \\
    \midrule
        \multirow{4}{*}{Super Resolution} &\multirow{2}{*}{$S=12$} & RFJS & \textbf{0.603} & \textbf{22.92} & \textbf{0.262} & \textbf{0.619} \\
       & & DPS & 1.23 & 22.78 & 0.265 & 0.613 \\
       &\multirow{2}{*}{$S=32$} & RFJS & \textbf{0.748} & \textbf{18.66} & \textbf{0.354} & \textbf{0.496} \\
       & & DPS & 1.38 & 18.43 & 0.363 & 0.489 \\
    \midrule
       \multirow{2}{*}{Motion Deblur}
        & \multirow{2}{*}{$K=256$}
        & RFJS & \textbf{0.545} & 22.67 & \textbf{0.257} & 0.614 \\
        & & DPS & 1.21 & \textbf{22.69} & 0.260 & \textbf{0.615} \\
    \bottomrule
    \end{tabular}
\end{table}
}

{\setlength{\tabcolsep}{6pt}  % default is 6pt
\begin{table}[ht]
\caption{Quantitative results using MPGD for hard tasks with ${\rm scale} = 6.0$ (guidance strength from the measurement $y$)}
    \label{tab:mpgd_hard_tasks}
    \centering
    \begin{tabular}{|c|c|c|c|c|c|c|}
    \toprule
       Task Name & Task Parameters & Algo & FS ($\downarrow$) & PSNR  ($\uparrow$) & LPIPS  ($\downarrow$) & SSIM ($\uparrow$) \\
    \midrule
       \multirow{2}{*}{Inpainting}
        & \multirow{2}{*}{$M=96$}
        & RFJS & \textbf{0.594} & \textbf{24.83} & \textbf{0.171} & \textbf{0.749} \\
        & & MPGD & 0.831 & 24.79 & 0.172 & 0.747 \\
    \midrule
        \multirow{4}{*}{Super Resolution} &\multirow{2}{*}{$S=12$} & RFJS & \textbf{1.032} & \textbf{21.40} & \textbf{0.312} & \textbf{0.555} \\
       & & MPGD & 1.282 & 21.32 & 0.318 & 0.553 \\
       &\multirow{2}{*}{$S=16$} & RFJS & \textbf{1.142} & \textbf{19.03} & 0.383 & \textbf{0.486} \\
       & & MPGD & 1.356 & 19.01 & \textbf{0.381} & 0.486 \\
       \midrule
        \multirow{2}{*}{Gaussian Deblur}
        & \multirow{2}{*}{$K=81,I=5.0$}
        & RFJS & \textbf{0.820} & \textbf{24.49} & \textbf{0.238} & \textbf{0.645} \\
        & & MPGD & 0.993 & 24.39 & 0.241 & 0.641 \\
    \bottomrule
    \end{tabular}
\end{table}}

\subsubsection{Side Information Quality}
\label{app: side_info_quality}

In this subsection, we evaluate how the quality of side information affects the performance of our method. For the face experiments, we study a $4\times$ super-resolution task using DPS as the base sampler. We degrade the side information by applying a Gaussian blur with a $31\times 31$ kernel and varying blur intensities. Figure \ref{fig:face_sim_vs_degradation} reports the resulting FS scores. As expected, performance consistently improves as the side information becomes more reliable. The case with blur intensity $0.0$ corresponds to the unaltered side information and matches the results reported in the main paper. Representative qualitative examples are provided in Figure \ref{fig:qualitative_results_face_side_info_degradation}. Notably, even with heavily degraded side information (intensity 5.0), our method (FS: 0.53) outperforms the baseline (FS: 1.042), and the performance steadily improves as the fidelity of the side information increases.

\begin{figure}[H]
    \centering
    \includegraphics[width=0.5\linewidth]{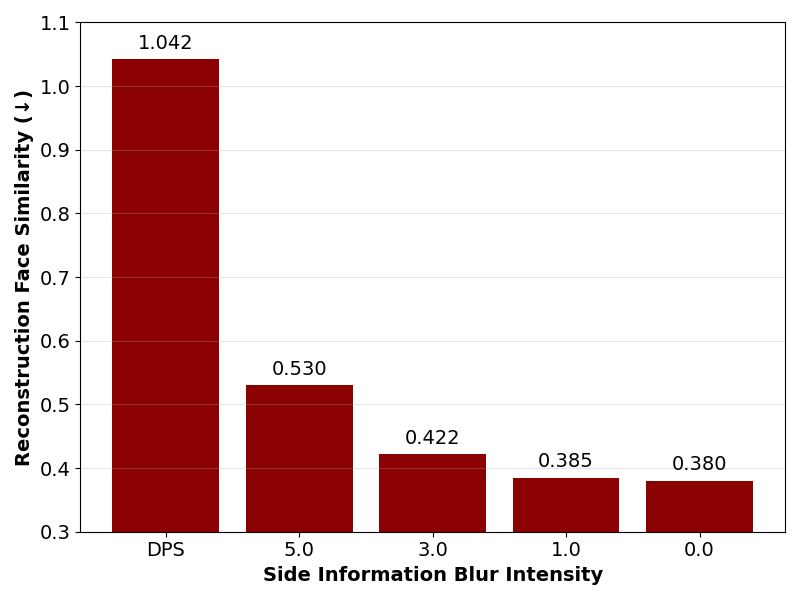}
    \caption{Performance of RFJS algorithm with side information blurred with Gaussian kernel in the face data experiments}
    \label{fig:face_sim_vs_degradation}
\end{figure}

\begin{figure}[H]
  \centering
  \makebox[\textwidth]{
  \scalebox{1}{
  \begin{tabular}{c}
   \begin{tabular}{>{\centering\arraybackslash}m{0.5cm} *{4}{>{\centering\arraybackslash}m{2.2cm}}}
    & \small GT/Input
    & \small Blur 1.0 & \small Blur 3.0 & \small Blur 5.0 \\

    {\centering\rotatebox{90}{\small GT/RFJS}} & \includegraphics[width=\linewidth]{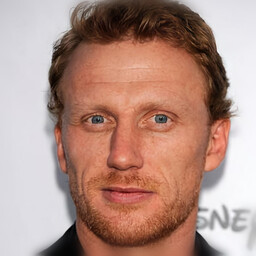} & \includegraphics[width=\linewidth]{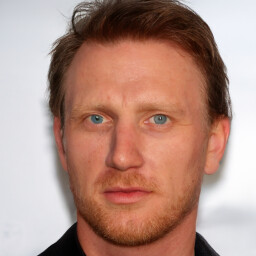} &
    \includegraphics[width=\linewidth]{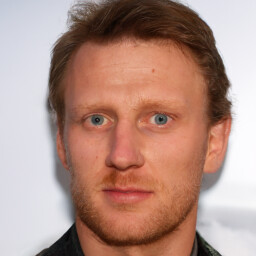} &
    \includegraphics[width=\linewidth]{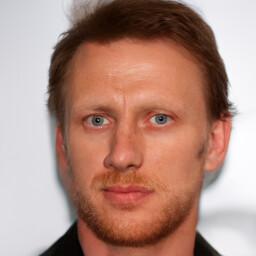}
    \\

    {\centering\rotatebox{90}{\small Input/Side Info}} & \includegraphics[width=\linewidth]{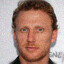} & \includegraphics[width=\linewidth]{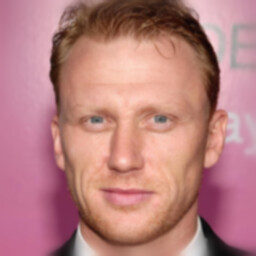} &
    \includegraphics[width=\linewidth]{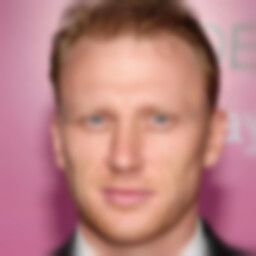} &
    \includegraphics[width=\linewidth]{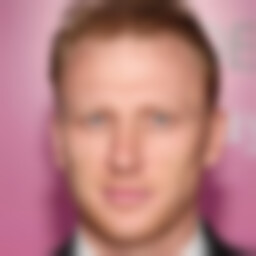}
    \\

    {\centering\rotatebox{90}{\small GT/RFJS}} & \includegraphics[width=\linewidth]{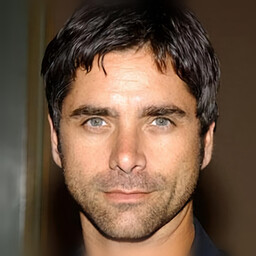} & \includegraphics[width=\linewidth]{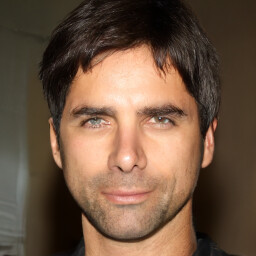} &
    \includegraphics[width=\linewidth]{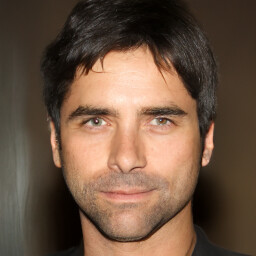} &
    \includegraphics[width=\linewidth]{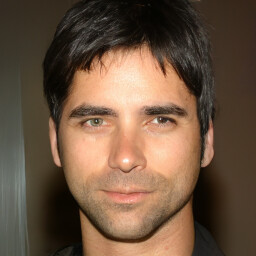}
    \\

    {\centering\rotatebox{90}{\small Input/Side Info}} & \includegraphics[width=\linewidth]{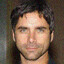} & \includegraphics[width=\linewidth]{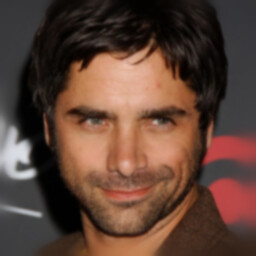} &
    \includegraphics[width=\linewidth]{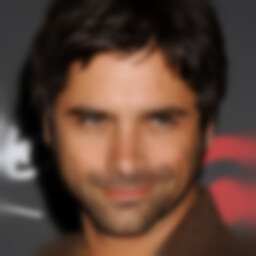} &
    \includegraphics[width=\linewidth]{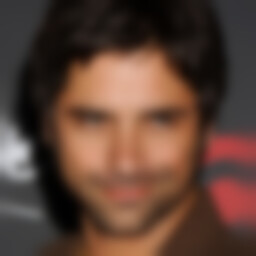}
    \\

    {\centering\rotatebox{90}{\small GT/RFJS}} & \includegraphics[width=\linewidth]{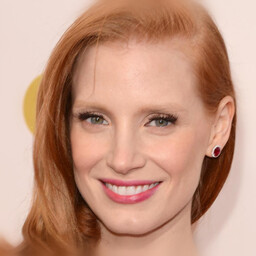} & \includegraphics[width=\linewidth]{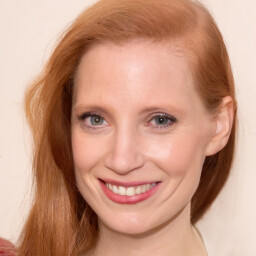} &
    \includegraphics[width=\linewidth]{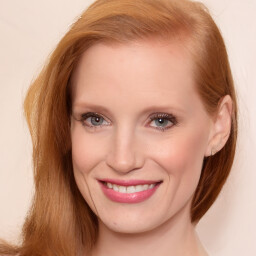} &
    \includegraphics[width=\linewidth]{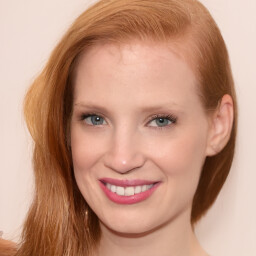}
    \\

    {\centering\rotatebox{90}{\small Input/Side Info}} & \includegraphics[width=\linewidth]{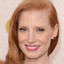} & \includegraphics[width=\linewidth]{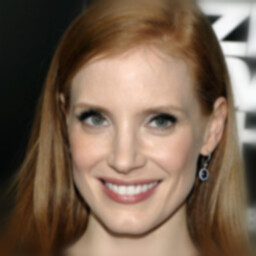} &
    \includegraphics[width=\linewidth]{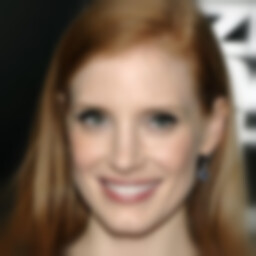} &
    \includegraphics[width=\linewidth]{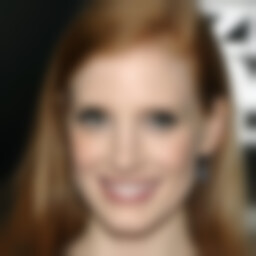}
    \\

    {\centering\rotatebox{90}{\small GT/RFJS}} & \includegraphics[width=\linewidth]{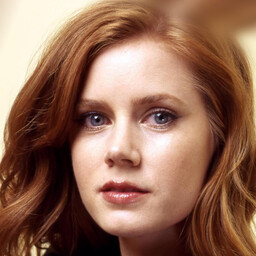} & \includegraphics[width=\linewidth]{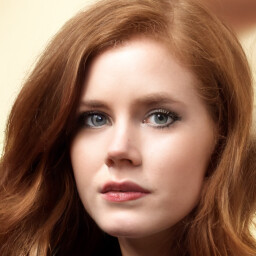} &
    \includegraphics[width=\linewidth]{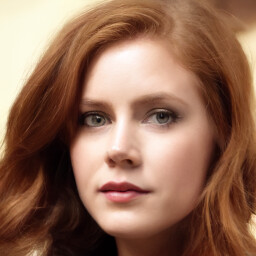} &
    \includegraphics[width=\linewidth]{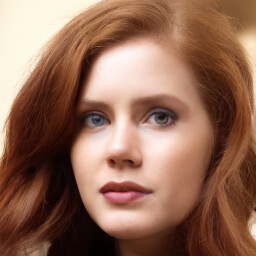}
    \\

    {\centering\rotatebox{90}{\small Input/Side Info}} & \includegraphics[width=\linewidth]{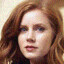} & \includegraphics[width=\linewidth]{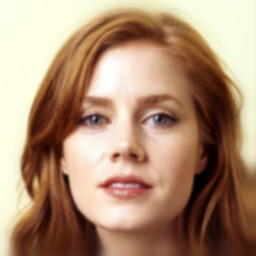} &
    \includegraphics[width=\linewidth]{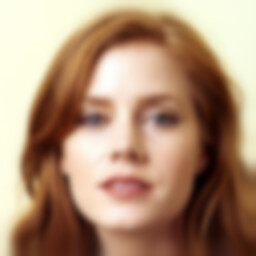} &
    \includegraphics[width=\linewidth]{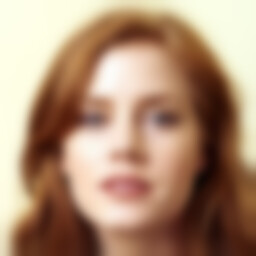}
    \\

  \end{tabular}
  \end{tabular}
  }} % END \makebox and \scalebox
  \caption{Qualitative comparison of generations with side information blurred at different levels. Some sharp features that preserve identity are lost when the side information is severely blurred. In the last three columns, first rows indicate reconstruction and second rows indicate the blurred side information that is used to guide the generation process.}
  \label{fig:qualitative_results_face_side_info_degradation}
\end{figure}

For the case of textual side information, Figure~\ref{fig:qualitative_results_text_side_info_degradation} shows that our method remains consistent under small variations in the prompt. In this example, the key missing information in the measurement is the type of animal or object present in the scene. Once the side information specifies that there is a golden retriever, the reconstruction improves significantly. Further modifying the prompt does not substantially affect the output. If the side information is "white dog", then our algorithm reconstructs a dog which is not necessarily a golden retriever (Figure~\ref{fig:qualitative_results_text_side_info_degradation}, last column).

\begin{figure}[t]
  \centering
  \makebox[\textwidth]{
  \scalebox{1}{
  \begin{tabular}{c}
  \begin{tabular}{>{\centering\arraybackslash}m{0.5cm} *{5}{>{\centering\arraybackslash}m{2.2cm}}}
    & \small Measurement & \small Ground Truth
    & \small Golden Retriever sitting on frozen lake looking forward & \small Golden Retriever & \small White Dog \\

    {\centering\rotatebox{90}{\small 1st Run}} &
    \includegraphics[width=\linewidth]{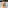} &
    \includegraphics[width=\linewidth]{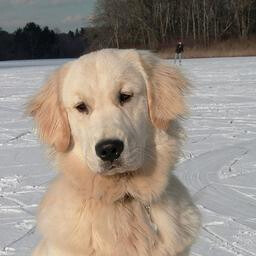} &
    \includegraphics[width=\linewidth]{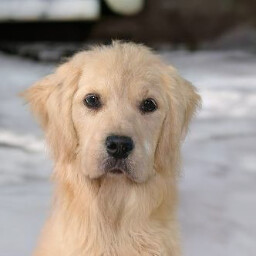} &
    \includegraphics[width=\linewidth]{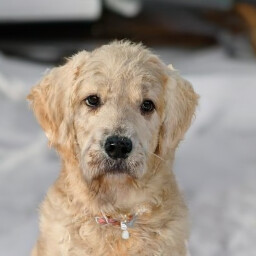} &
    \includegraphics[width=\linewidth]{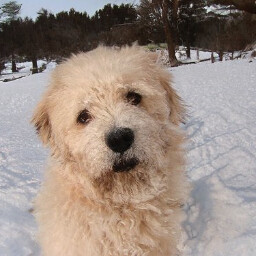} \\ [-2pt]

    {\centering\rotatebox{90}{\small 2nd Run}} &
    \includegraphics[width=\linewidth]{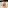} &
    \includegraphics[width=\linewidth]{pictures/change_side_info/gt.jpg} &
    \includegraphics[width=\linewidth]{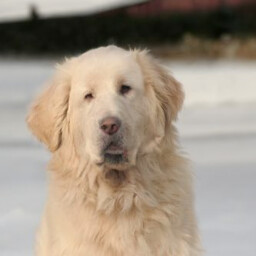} &
    \includegraphics[width=\linewidth]{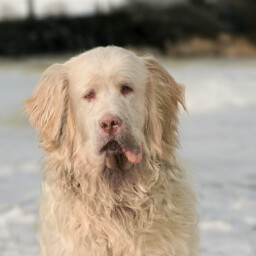} &
    \includegraphics[width=\linewidth]{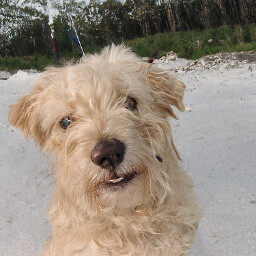} \\ [-2pt]

    {\centering\rotatebox{90}{\small 3rd Run}} &
    \includegraphics[width=\linewidth]{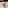} &
    \includegraphics[width=\linewidth]{pictures/change_side_info/gt.jpg} &
    \includegraphics[width=\linewidth]{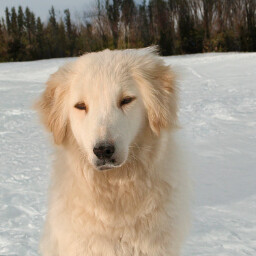} &
    \includegraphics[width=\linewidth]{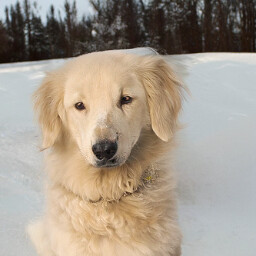} &
    \includegraphics[width=\linewidth]{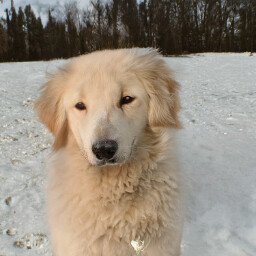} \\ [-2pt]

  \end{tabular}
  \end{tabular}
  }} % END \makebox and \scalebox
  \caption{Qualitative comparison of generations with different textual descriptions as side information. Due to randomness and robustness of the algorithm due to relative ordering, the quality assessment is difficult.}
  \label{fig:qualitative_results_text_side_info_degradation}
\end{figure}

}

\paragraph{\rev{Semantically incorrect side information.}}\rev{The degradation study above keeps the side information correct and reduces its fidelity. A different and more adversarial question is what happens when the side information is simply wrong, that is, when it carries confident information about the wrong target. We test this on $4\times$ super-resolution by replacing the reference image of each target with an image of a \emph{different} identity, leaving everything else unchanged. \Cref{tab:wrong_side_info} reports the result. Identity fidelity degrades sharply, which is the expected and correct behaviour, but the distortion metrics move very little. The reason is structural: the search selects among candidates produced by the measurement-conditioned transition kernel of the base solver, and never overrides $\vy$. An unreliable reward can therefore cause the algorithm to select the wrong identity among reconstructions that all remain plausible under the measurement, but it does not let the reconstruction drift away from the measurement altogether. In a severely ill-posed problem, where several semantically distinct reconstructions explain the same $\vy$, this is a genuine failure mode, and we discuss it further in \cref{app:limitations}.}

\begin{table}[H]
\centering
\scriptsize
\setlength{\tabcolsep}{6pt}
\rev{%
\begin{tabular}{lcccc}
\toprule
\textbf{Side information} & \textbf{FS}$\downarrow$ & \textbf{PSNR}$\uparrow$ & \textbf{LPIPS}$\downarrow$ & \textbf{SSIM}$\uparrow$ \\
\midrule
Correct reference            & \textbf{0.53} & \textbf{27.5} & \textbf{0.175} & \textbf{0.773} \\
Reference of a different identity & 0.91 & 27.1 & 0.183 & 0.759 \\
\bottomrule
\end{tabular}}
\caption{\rev{Effect of semantically incorrect side information on $4\times$ super-resolution with DPS as the base solver. Replacing the reference with an image of a different identity costs most of the identity gain, while the distortion metrics are almost unchanged, because the search only chooses among measurement-consistent candidates. The run uses $N=8$ particles, $B=16$ and a DPS step size of $0.8$, and is evaluated on $8$ images; the absolute values are therefore not comparable with the $4\times$ super-resolution row of \cref{tab:quantitative_results_dps}, and only the comparison between the two rows here is meaningful.}}
\label{tab:wrong_side_info}
\end{table}

\paragraph{\rev{Multiple reference images.}}\rev{\Cref{tab:multi_reference} compares the two aggregations of \cref{ssec:reward-modeling} for $K=2$, on box inpainting with a $116 \times 116$ mask, DPS as the base solver, $N=8$ particles and $B=16$. The gain over a single reference is consistent but modest: the mean improves three of the four metrics and the maximum improves all four. That ordering is what one expects when the two references differ in how closely they match the pose and lighting of the target, since rewarding alignment to the closest reference then ignores the less informative one.}

\begin{table}[H]
\centering
\scriptsize
\setlength{\tabcolsep}{6pt}
\rev{%
\begin{tabular}{lcccc}
\toprule
\textbf{Reference set} & \textbf{PSNR}$\uparrow$ & \textbf{LPIPS}$\downarrow$ & \textbf{SSIM}$\uparrow$ & \textbf{FS}$\downarrow$ \\
\midrule
$K=1$                                  & 26.11 & 0.1389 & 0.8451 & 0.2750 \\
$K=2$, mean reward                     & 26.42 & 0.1404 & 0.8491 & 0.2696 \\
$K=2$, max reward (nearest reference)  & \textbf{26.58} & \textbf{0.1337} & \textbf{0.8498} & \textbf{0.2613} \\
\bottomrule
\end{tabular}}
\caption{\rev{Aggregating the rewards of several reference images, for box inpainting with a $116 \times 116$ mask, DPS as the base solver, $N=8$ particles and $B=16$, evaluated on $8$ images.}}
\label{tab:multi_reference}
\end{table}

\subsection{Experiments on Effect of Gradient Scale}
\label{app:effect_of_grad_scale}

\begin{figure}[H]
    \centering
    \scalebox{0.5}{ % Change this value to make the figure smaller (<1) or larger (>1)
        \includegraphics[width=\linewidth]{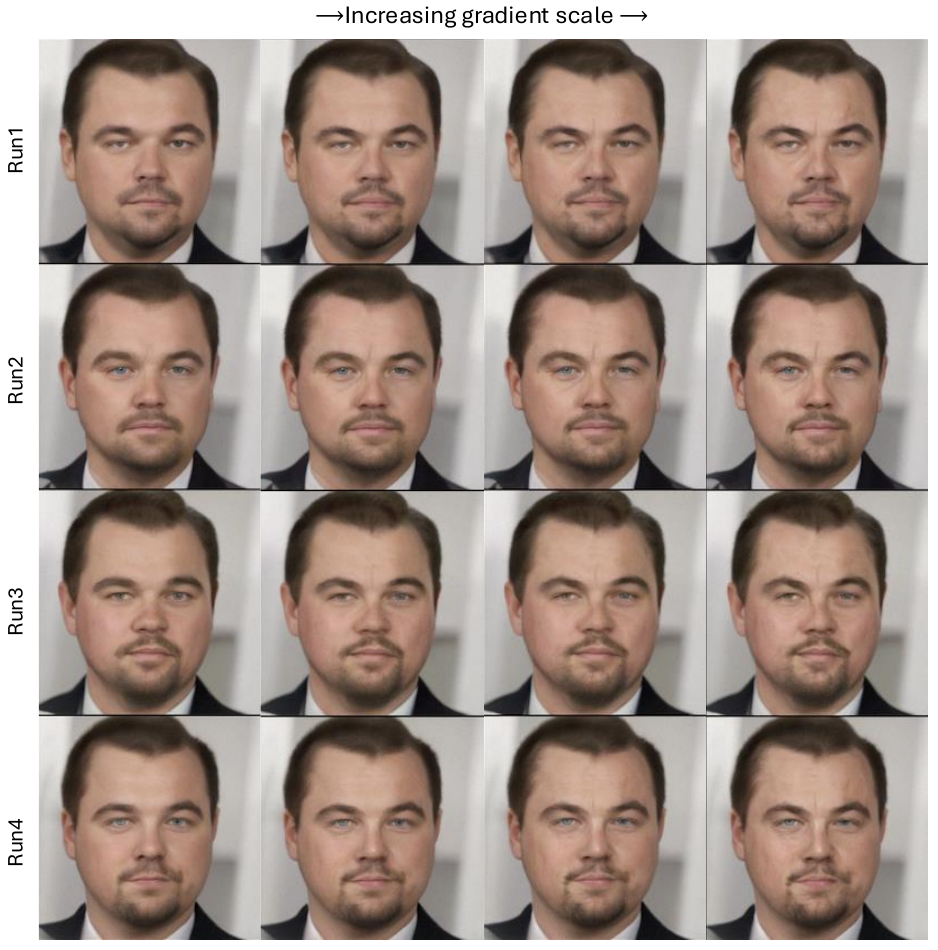}
    }
    \caption{
    \textbf{Effect of reward-gradient guidance in diffusion-based inverse problems.}
    We show 4 runs with different random seeds (rows), and for each seed we vary the gradient scale across 4 settings (columns).
    Within each row, the noise realization is identical and only the gradient scale changes; within each column, the gradient scale is fixed while the random seed varies.
    The ground truth and degraded input are the same for all reconstructions.
    This arrangement reveals two key observations:
    (1) The reward gradient influences fine details, such as wrinkles and facial lines without altering the overall facial structure; the structure is primarily determined by the initial noise realization.
    (2) Different seeds reconstruct different face structures, highlighting the multi-modal nature of the problem.
    This demonstrates why using multiple particles and performing search across them is beneficial: it enables exploration of structurally different hypotheses while the reward gradient refines locally.
    }
    \label{fig:grad_effect}
\end{figure}

\textbf{Gradient Guidance Limitations.} While guiding the reverse diffusion process with reward gradients can help generate images with higher reward scores, this approach has several limitations. First, as shown in Figures  \ref{fig:grad_effect}, \ref{fig:DPSGradEffect}, gradient-based guidance primarily adds fine details, such as wrinkles or texture, to the reconstruction, but it cannot significantly alter the global structure of the face. To isolate the effect of the gradient, we used fixed noise realizations for both the gradient-based and baseline methods. The results show that changes are mostly confined to local details, implying that if the sampling trajectory is poor, gradient guidance alone cannot compensate. This highlights the need for search-based methods that can explore a wider range of trajectories during inference.

Second, this method is sensitive to the choice of gradient scale. In the visual examples, we used a relatively large scale of 1.6 to make the gradient’s effect more visible; however, such high scales often degrade other metrics like PSNR and SSIM and introduce artifacts. Empirically, we found that a scale around $0.5$ yields the best balance when the base sampler is DPS or BlindDPS, consistently improving the FS metric \rev{while leaving the other evaluation metrics in the same range} and avoiding artifacts (see Figure \ref{fig:DPSGradEffect}). Moreover, the sensitivity to gradient scale increases when the number of reverse diffusion steps is small. For instance, in DAPS (Figure~\ref{fig:grad_effect}) and MPGD, where the number of steps is limited to 200 and 100 respectively, larger scales quickly lead to visible artifacts, as demonstrated in Figures \ref{fig:qualitative_results_mpgd_bi_64}, and \ref{fig:qualitative_results_mpgd_sr_6x}.

It is well established in deep learning research that deep (convolutional) neural networks are vulnerable to gradient-based adversarial attacks \citep{goodfellow2015explainingadvattacks}. Consequently, using such networks to provide reward-based guidance inherits these vulnerabilities. However, when combined with diffusion samplers, this susceptibility is partially alleviated, as the diffusion process can help steer trajectories away from adversarially induced local minima. This mitigating effect is particularly evident when using a large number of sampling steps (e.g., 1000 steps in DPS). In contrast, samplers with fewer steps (e.g., 100 steps in MPGD) exhibit increased sensitivity to the gradient scale, as illustrated in Figures \ref{fig:qualitative_results_mpgd_bi_64} and \ref{fig:qualitative_results_mpgd_sr_6x}.

{
\setlength{\tabcolsep}{1pt}
\renewcommand{\arraystretch}{1.0}
\begin{figure}[H]
  \centering
  \makebox[\textwidth]{
  \scalebox{0.66}{
  \begin{tabular}{c c}
  \begin{tabular}{>{\centering\arraybackslash}m{0.5cm} *{6}{>{\centering\arraybackslash}m{2.2cm}}}
    & \small DPS & \small Gradient (1.6) & \small Ground Truth \\

    \multirow{4}{*}{\parbox[c][8cm][c]{0.5cm}{\centering\rotatebox{90}{\footnotesize Super Resolution (4)}}} &
    \includegraphics[width=\linewidth]{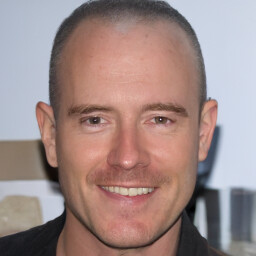} &
    \includegraphics[width=\linewidth]{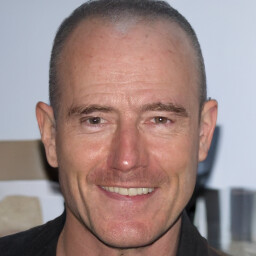} &
    \includegraphics[width=\linewidth]{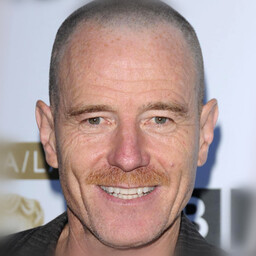}  \\[-2pt]

    &
    \includegraphics[width=\linewidth]{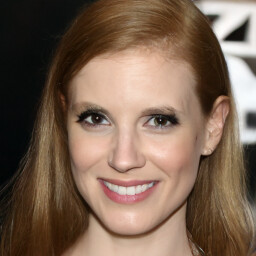} &
    \includegraphics[width=\linewidth]{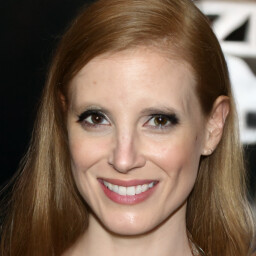} &
    \includegraphics[width=\linewidth]{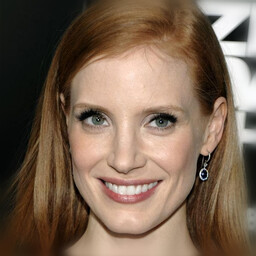}  \\[-2pt]

     &
    \includegraphics[width=\linewidth]{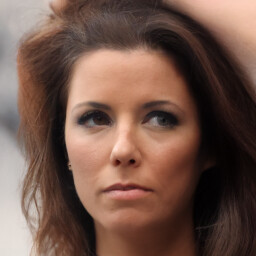} &
    \includegraphics[width=\linewidth]{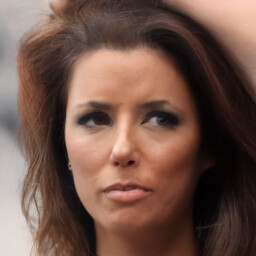} &
    \includegraphics[width=\linewidth]{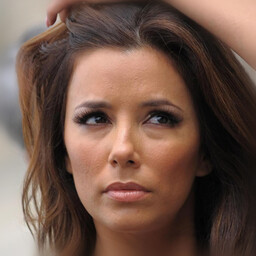}  \\[-2pt]

    &
    \includegraphics[width=\linewidth]{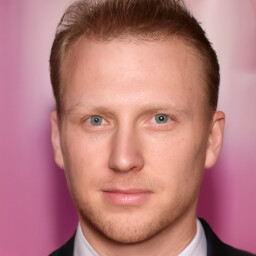} &
    \includegraphics[width=\linewidth]{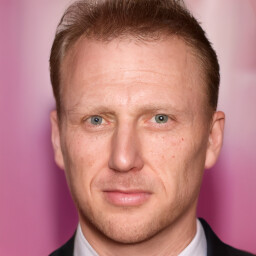} &
    \includegraphics[width=\linewidth]{pictures/DPS/label/img_00056.jpg}  \\[-2pt]

     &
    \includegraphics[width=\linewidth]{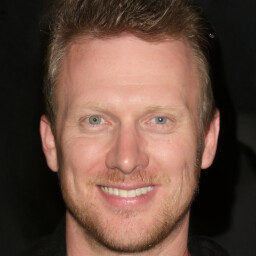} &
    \includegraphics[width=\linewidth]{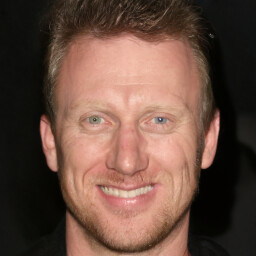} &
    \includegraphics[width=\linewidth]{pictures/DPS/label/img_00038.jpg}  \\
\end{tabular}

  \end{tabular}
  }} % END \makebox and \scalebox
  \caption{Qualitative comparison of the effect of gradient scale on reconstruction paths.}
  \label{fig:DPSGradEffect}
\end{figure}
}

{
\setlength{\tabcolsep}{1pt}
\renewcommand{\arraystretch}{1.0}

\begin{figure}[H]
  \centering
  \makebox[\textwidth]{
  \scalebox{0.6}{
  \begin{tabular}{c c}
  \begin{tabular}{>{\centering\arraybackslash}m{0.5cm} *{12}{>{\centering\arraybackslash}m{2.2cm}}}
    & \small MPGD & \small Gradient (0.5) & \small Gradient (1.0) & \small Gradient (1.5) & \small Ground Truth  & \small MPGD & \small Gradient (0.5) & \small Gradient (1.0) & \small Gradient (1.5) & \small Ground Truth \\

    \multirow{2}{*}{\parbox[c][2.5cm][c]{0.5cm}{\centering\rotatebox{90}{\footnotesize Box Inpainting (64)}}} &
    \includegraphics[width=\linewidth]{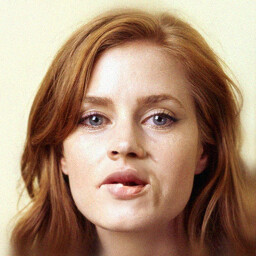} &
    \includegraphics[width=\linewidth]{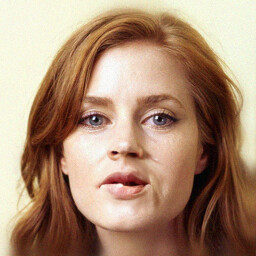} &
    \includegraphics[width=\linewidth]{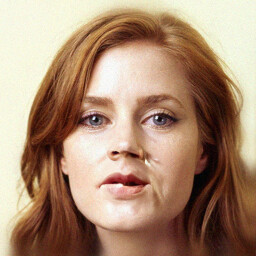} &
    \includegraphics[width=\linewidth]{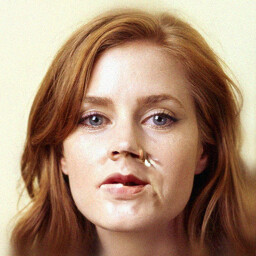} &
    \includegraphics[width=\linewidth]{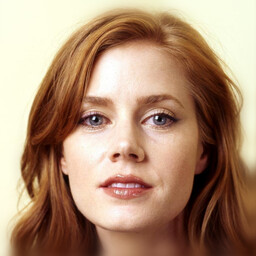}
    &
    \includegraphics[width=\linewidth]{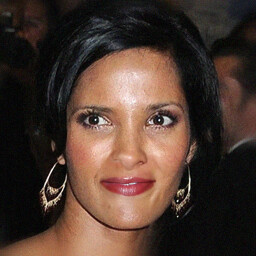} &
    \includegraphics[width=\linewidth]{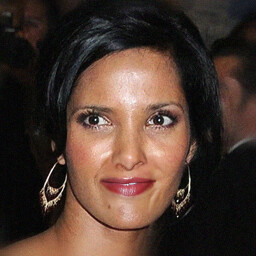} &
    \includegraphics[width=\linewidth]{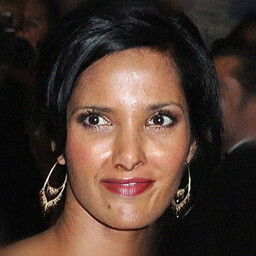} &
    \includegraphics[width=\linewidth]{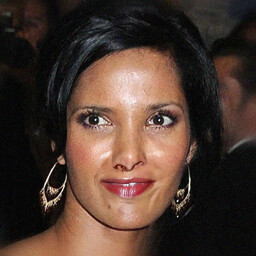} &
    \includegraphics[width=\linewidth]{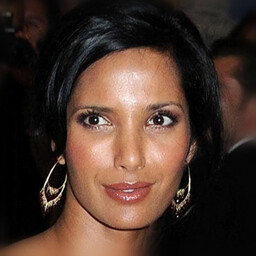} \\

     &
    \includegraphics[width=\linewidth]{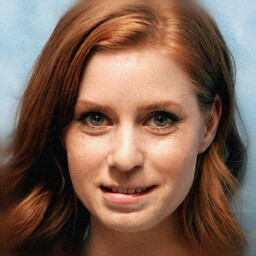} &
    \includegraphics[width=\linewidth]{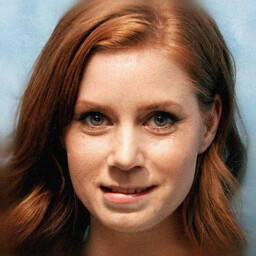} &
    \includegraphics[width=\linewidth]{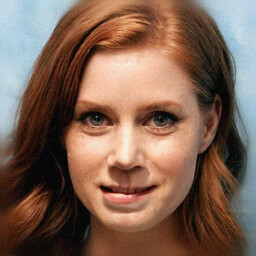} &
    \includegraphics[width=\linewidth]{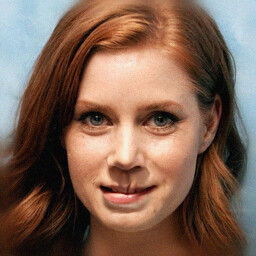} &
    \includegraphics[width=\linewidth]{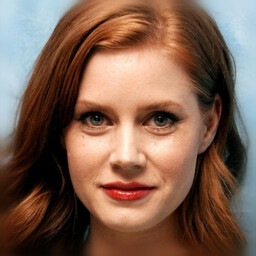}
     &
    \includegraphics[width=\linewidth]{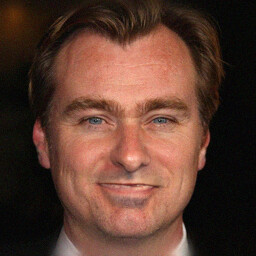} &
    \includegraphics[width=\linewidth]{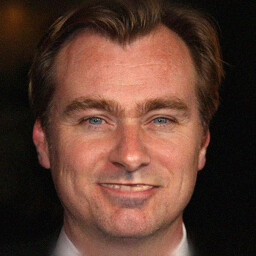} &
    \includegraphics[width=\linewidth]{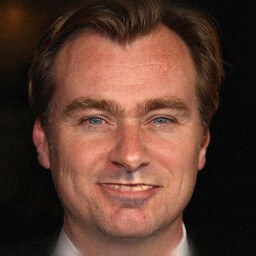} &
    \includegraphics[width=\linewidth]{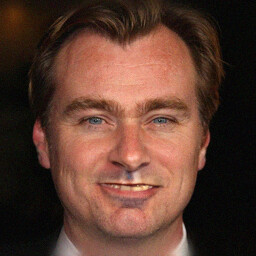} &
    \includegraphics[width=\linewidth]{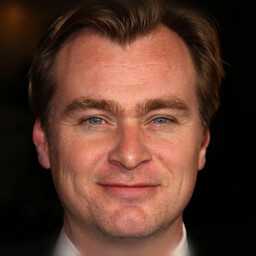} \\
\end{tabular}

  \end{tabular}
  }} % END \makebox and \scalebox
  \caption{Qualitative comparison of the effect of gradient scale on reconstruction paths. Notice that while the base reconstruction is reasonable, adding the gradient can degrade it if the scale is very large. The final scale used in the experiments is $0.5$.}
  \label{fig:qualitative_results_mpgd_bi_64}
\end{figure}
}

{
\setlength{\tabcolsep}{1pt}
\renewcommand{\arraystretch}{1.0}

\begin{figure}[H]
  \centering
  \makebox[\textwidth]{
  \scalebox{0.6}{
  \begin{tabular}{c c}
  \begin{tabular}{>{\centering\arraybackslash}m{0.5cm} *{12}{>{\centering\arraybackslash}m{2.2cm}}}
    & \small MPGD & \small Gradient (0.25) & \small Gradient (0.5) & \small Gradient (1.0) & \small Ground Truth & \small MPGD & \small Gradient (0.25) & \small Gradient (0.5) & \small Gradient (1.0) & \small Ground Truth \\

    \multirow{2}{*}{\parbox[c][3cm][c]{0.5cm}{\centering\rotatebox{90}{\footnotesize Super Resolution (6)}}} &
    \includegraphics[width=\linewidth]{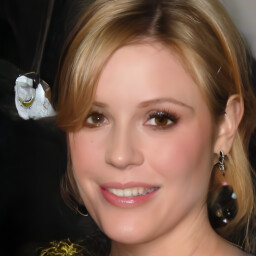} &
    \includegraphics[width=\linewidth]{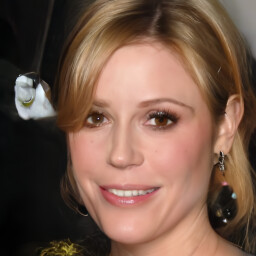} &
    \includegraphics[width=\linewidth]{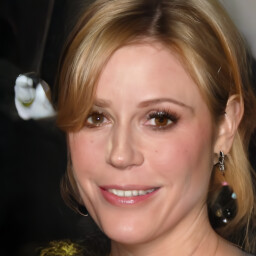} &
    \includegraphics[width=\linewidth]{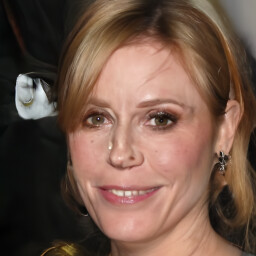} &
    \includegraphics[width=\linewidth]{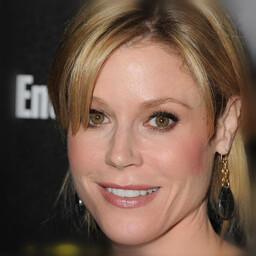}
    &
    \includegraphics[width=\linewidth]{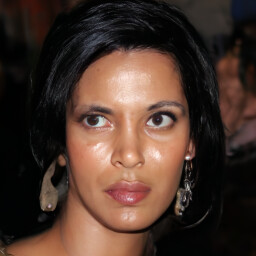} &
    \includegraphics[width=\linewidth]{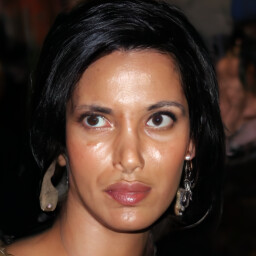} &
    \includegraphics[width=\linewidth]{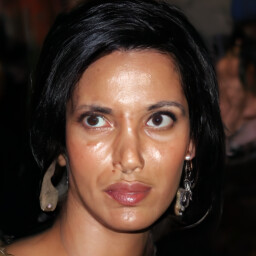} &
    \includegraphics[width=\linewidth]{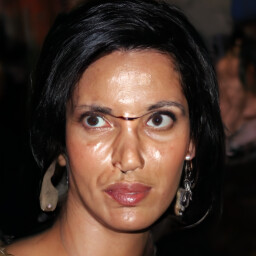} &
    \includegraphics[width=\linewidth]{pictures/mpgd/label/005_label.jpg} \\
    &
    \includegraphics[width=\linewidth]{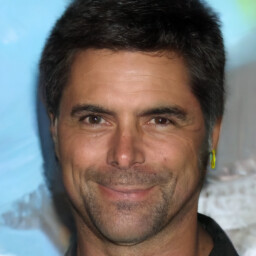} &
    \includegraphics[width=\linewidth]{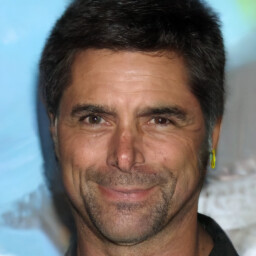} &
    \includegraphics[width=\linewidth]{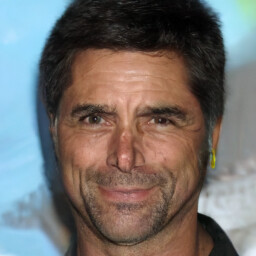} &
    \includegraphics[width=\linewidth]{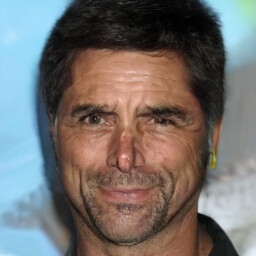} &
    \includegraphics[width=\linewidth]{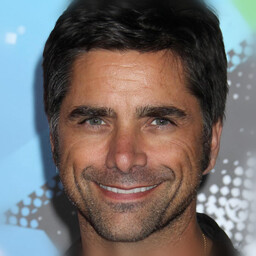}
    &
    \includegraphics[width=\linewidth]{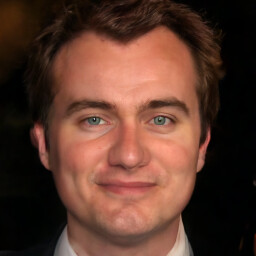} &
    \includegraphics[width=\linewidth]{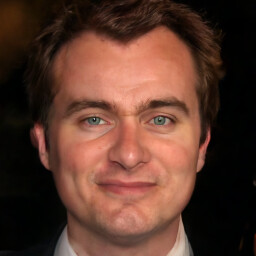} &
    \includegraphics[width=\linewidth]{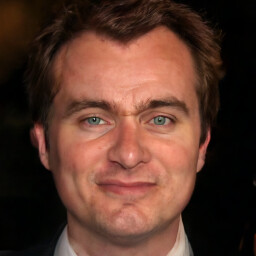} &
    \includegraphics[width=\linewidth]{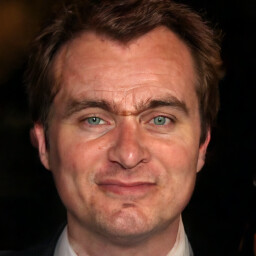} &
    \includegraphics[width=\linewidth]{pictures/mpgd/label/047_label.jpg} \\

\end{tabular}

  \end{tabular}
  }} % END \makebox and \scalebox
  \caption{Qualitative comparison of the effect of gradient scale on reconstruction paths for super resolution task. Notice that while the base reconstruction is reasonable, adding the gradient can degrade it if the scale is very large. The final scale used in the experiments is $0.25$.}
  \label{fig:qualitative_results_mpgd_sr_6x}
\end{figure}
}

\subsection{Effect of Number of Particles}
\label{app:num_particles_effect}

\begin{figure}[H]
    \centering
    \includegraphics[width=0.5\linewidth]{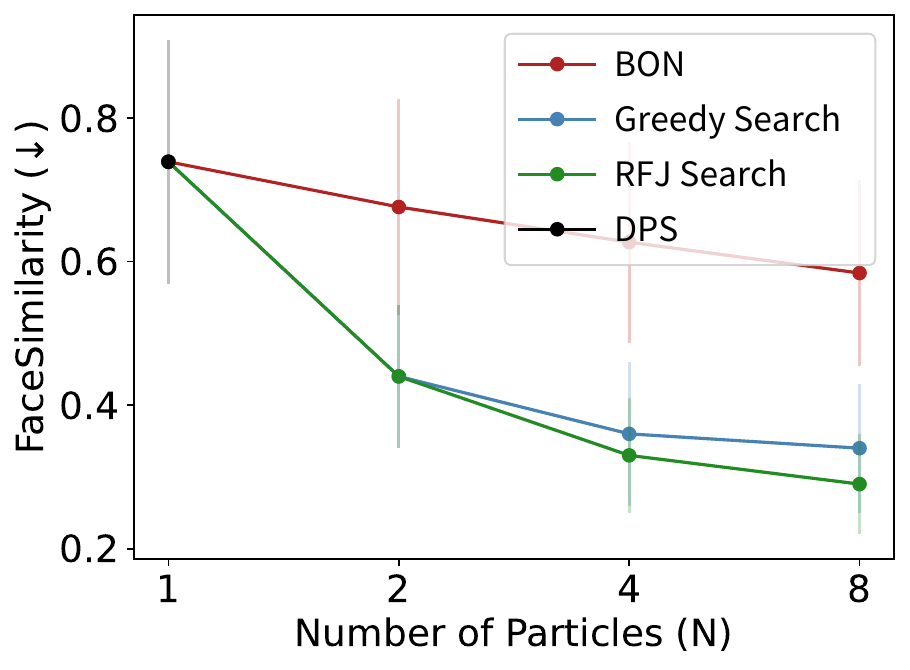}
    \caption{Scaling of search algorithms with respect to the number of particles.}
    \label{fig:facereward_vs_n}
\end{figure}

\begin{figure}[H]
    \centering
    \begin{subfigure}[b]{0.24\textwidth}
        \centering
        \includegraphics[width=\textwidth]{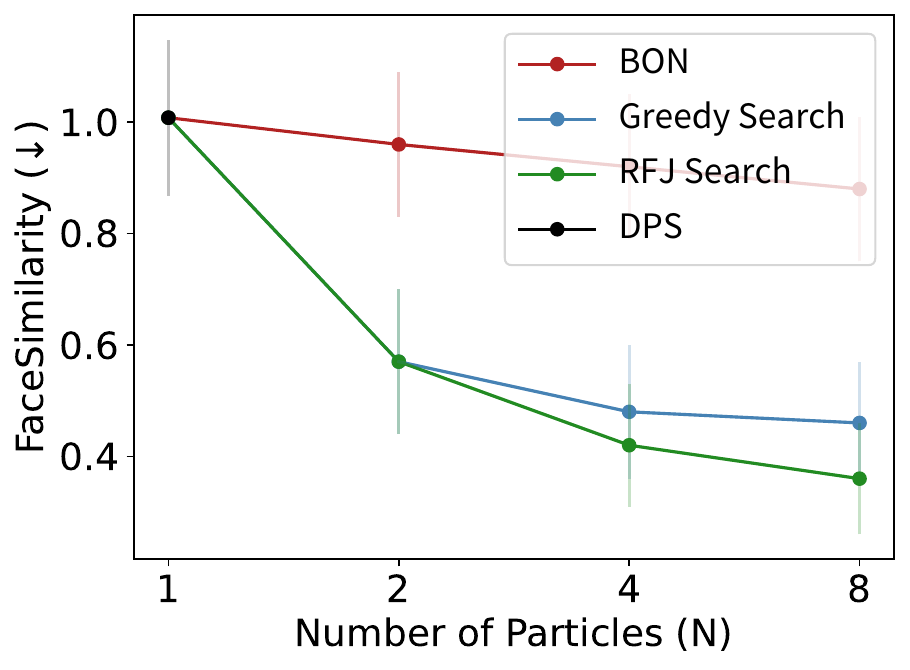}
        \label{fig:DPS-Nonlinear}
        \caption{Nonlinear Deblur}
    \end{subfigure}
     \hfill % spacing between subfigures
    \begin{subfigure}[b]{0.24\textwidth}
        \centering
        \includegraphics[width=\textwidth]{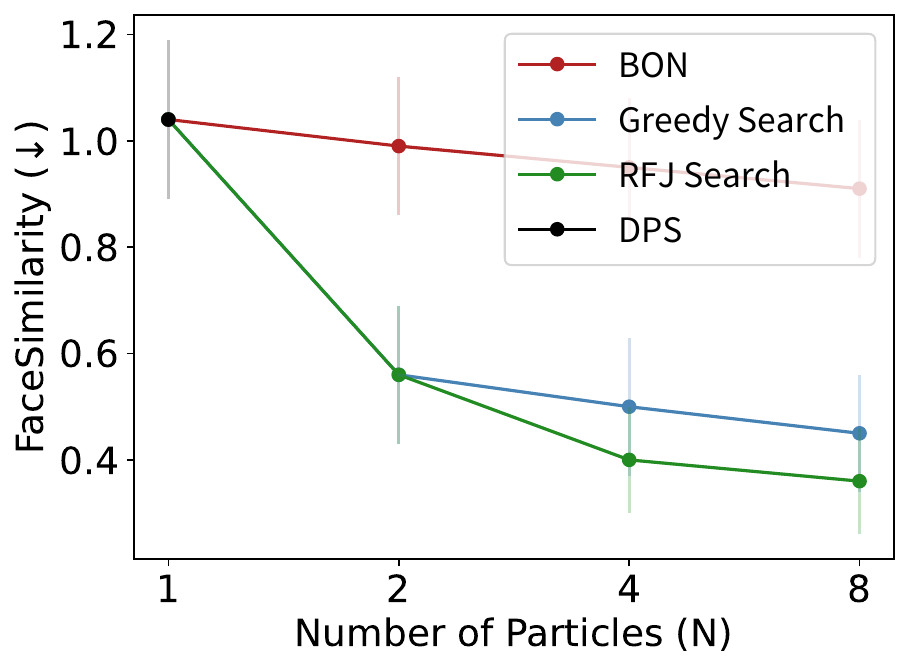}
        \label{fig:DPS-Super}
        \caption{Super Resolution (4)}
    \end{subfigure}
    \hfill
    \begin{subfigure}[b]{0.24\textwidth}
        \centering
        \includegraphics[width=\textwidth]{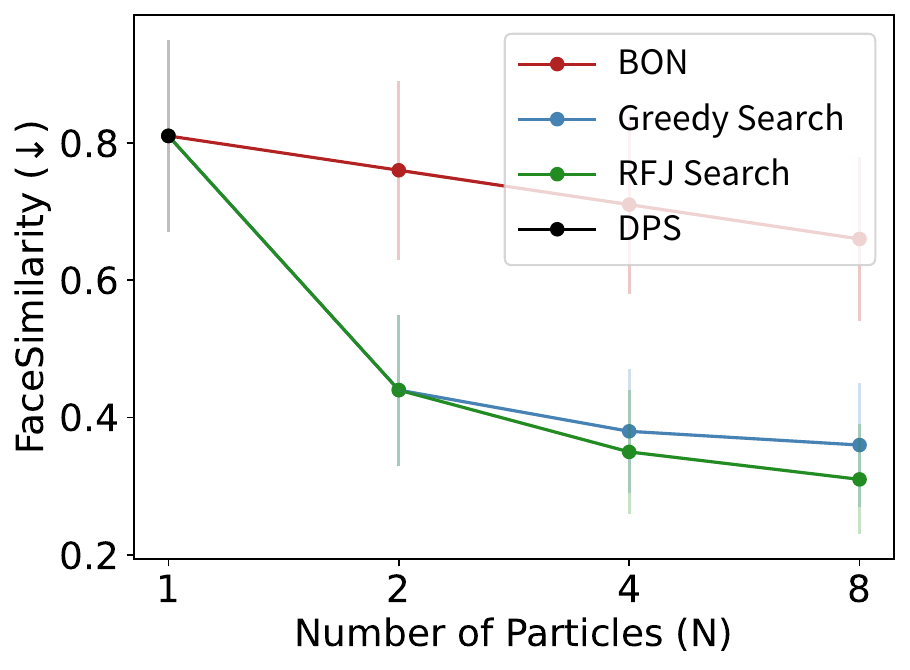}
        \label{fig:DPS-Gaussian}
        \caption{Gaussian Deblur}
    \end{subfigure}
    \hfill
    \begin{subfigure}[b]{0.24\textwidth}
        \centering
        \includegraphics[width=\textwidth]{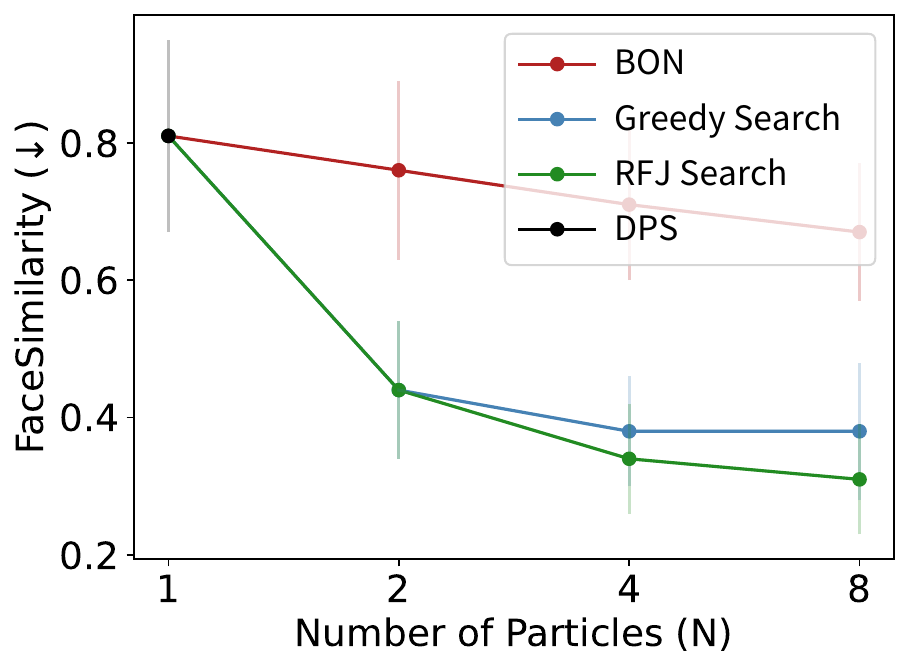}
        \label{fig:DPS-Motion}
        \caption{Motion Deblur}
    \end{subfigure}
    \caption{Effect of the number of particles $N$ on the FS metric in DPS. As $N$ increases, the performance improves.}
    \label{fig:face_sim_vs_n_dps}
\end{figure}

\begin{figure}
    \centering
    \begin{subfigure}[b]{0.31\textwidth}
        \centering
        \includegraphics[width=\textwidth]{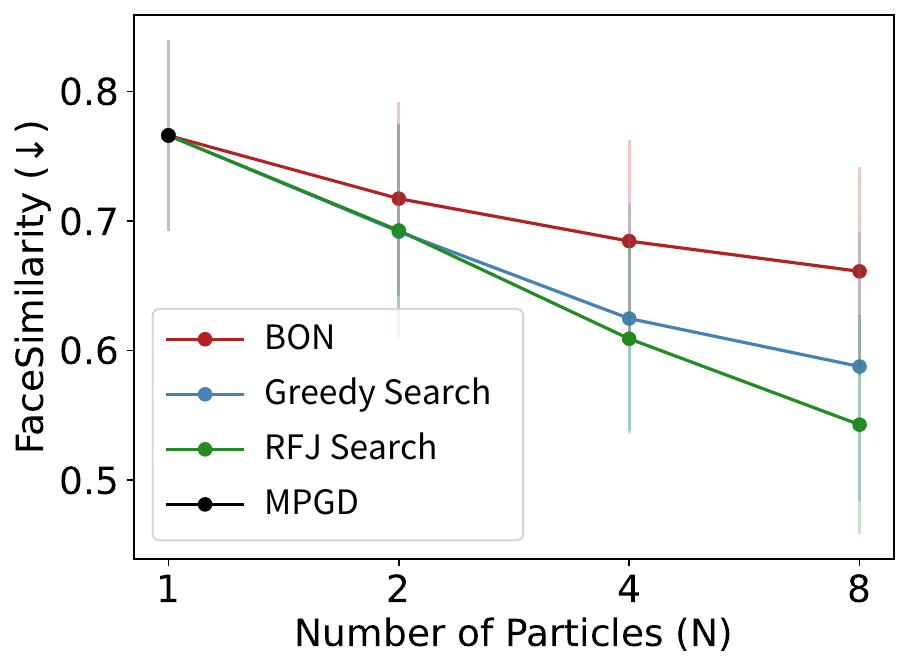}
        \label{fig:box_inpainting_64_face_sim_vs_n}
        \caption{Box Inpainting (Box: 64)}
    \end{subfigure}
    \hfill  % spacing between subfigures
    \begin{subfigure}[b]{0.31\textwidth}
        \centering
        \includegraphics[width=\textwidth]{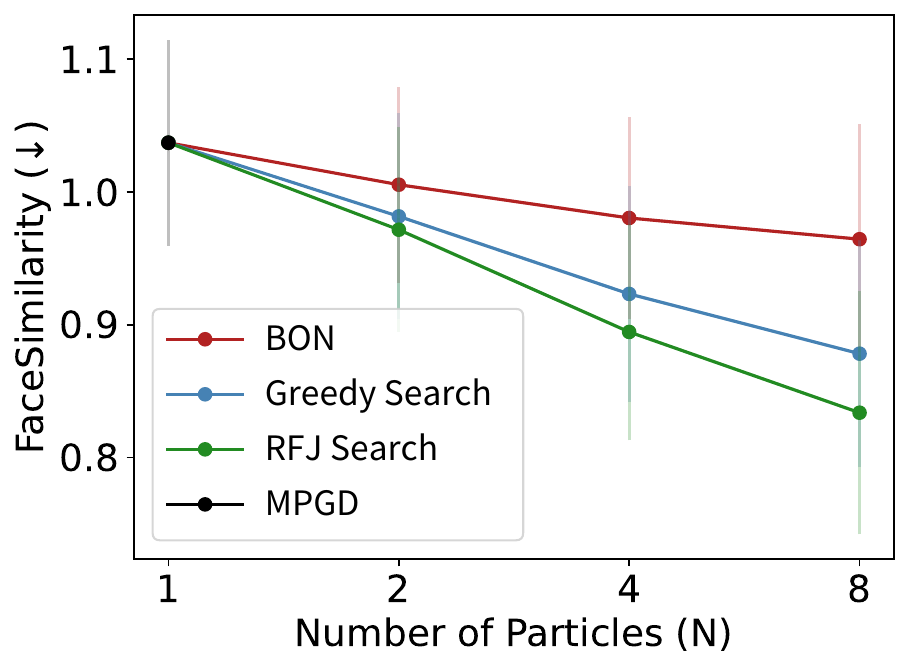}
        \label{fig:super_resolution_6x_face_sim_vs_n}
        \caption{Super-resolution (Scale: 6)}
    \end{subfigure}
     \hfill % spacing between subfigures
    \begin{subfigure}[b]{0.31\textwidth}
        \centering
        \includegraphics[width=\textwidth]{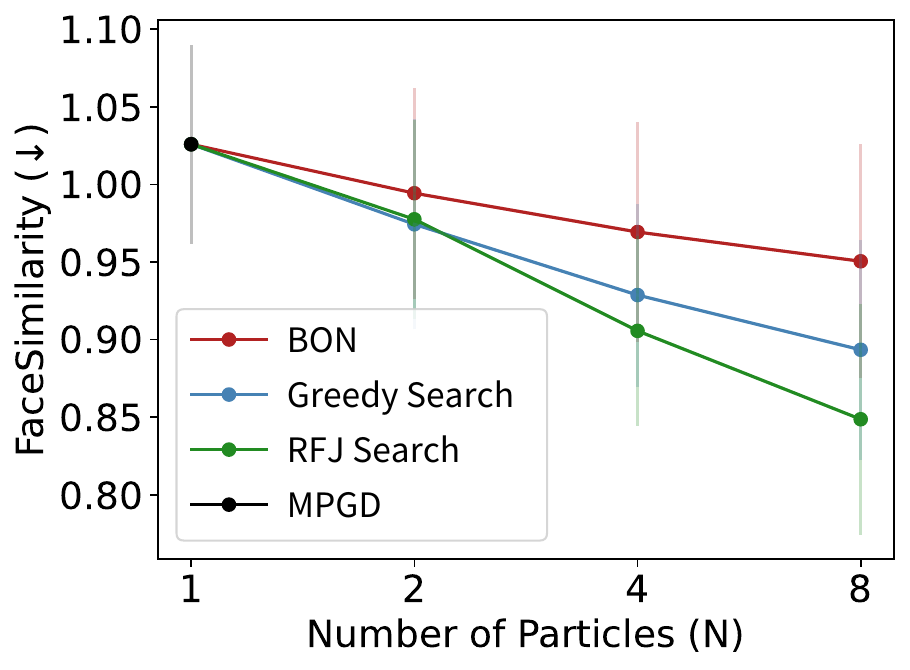}
        \label{fig:gaussian_blur_5_face_sim_vs_n}
        \caption{Gaussian Blur (Intensity: 5)}
    \end{subfigure}
    \caption{Effect of number of particles $N$ on the FS metric. RFJ Search algorithm offers the best scaling performance, followed by Greedy Search algorithm. Finally, BestOfN performance improves, but only marginally. $64$ indicates the size of the box for inpainting, $6$ indicates the down-sampling factor in super-resolution, and $5$ is the intensity of the Gaussian kernel in Gaussian deblur.}
    \label{fig:face_sim_vs_n}
\end{figure}

{% \color{blue}
In this section, we study how performance scales with the number of particles. Figure \ref{fig:facereward_vs_n} presents the results for the box inpainting task using RFJS with DPS as the baseline sampler, while Figure \ref{fig:face_sim_vs_n_dps} reports the corresponding results for the remaining tasks. The evaluations using MPGD as the baseline are shown in Figure \ref{fig:face_sim_vs_n}. Across all tasks and for both baseline samplers, we observe a consistent improvement in performance as the number of particles $N$ increases, aligning with the expected behavior of particle-based search methods. Furthermore, our results indicate that RFJS scales more efficiently with increasing $N$ than both Greedy Search and Best-of-N.
}

\subsection{Runtimes}

\label{app:runtimes}

Table~\ref{tab:time_particles_daps_mpgd} reports wall-clock runtimes (in seconds) for our search algorithms compared to Best-of-N (BON) and Greedy Search across different baselines (DPS, DAPS, and MPGD) with varying numbers of particles. When the number of particles is set to $1$, the runtime corresponds to the baseline method without search. As the number of particles increases, amount of computation scales linearly with $N$, but thanks to parallelization, the wall-clock overhead remains moderate: with $N=8$, runtimes are only $4-5\times$ those of the baseline. We also note that our RFJ Search method take slightly more time than BON and Greedy Search, but consistently achieve better reconstruction quality, highlighting the practical efficiency of our approach.

{% \color{blue}
Finally, resampling of the particles requires computing the rewards, which involves call to the reward evaluation model. Therefore, decreasing the base $B$ (more frequent resampling) increases the wall-clock time. Denoting $c_d, c_r$ the computation cost per function call to diffusion model and reward model respectively, a rough estimate of the time taken to run the algorithm is given by:
\begin{align*}
    N(O( T c_d) + O((T/B) c_r)) \le N(O( T c_d) + O(T c_r)),
\end{align*}
where $T$ denotes the number of diffusion steps and $N$ the number of particles. However, it is important to note that the wall-clock time is dominated by the call to the diffusion model, which is a much larger network than the reward model, i.e, $c_d \gg c_r$ in practice, and so the reward evaluation does not add a significant overhead.
}

\textbf{The compute--quality tradeoff is a feature, not a limitation.}
Our framework provides a controllable knob: increasing $N$ yields
better reconstructions, while $N=1$ recovers the base sampler exactly
at no additional cost. This is precisely the inference-time scaling
behavior that has driven recent progress in LLMs~\citep{snell2025scaling,
setlur2025rewarding}, and is a desirable property rather than a
drawback. We also note that any additional runtime cost is
\emph{inherent to inference-time search as a paradigm}, it is not
specific to our method. BON, Greedy Search, and any other
particle-based approach share the same linear scaling with $N$.
What distinguishes our framework is that the added compute is spent
meaningfully: as shown in Fig.~1 (right), scaling $N$ without side
information yields negligible gains, whereas our reward-guided search
produces consistent and significant improvements with every additional
particle. \textbf{The runtime cost is therefore justified by gains
that no amount of unguided compute scaling can achieve.}

{\setlength{\tabcolsep}{2 pt}
\begin{table}[H]
\centering
\caption{Runtime (seconds) vs number of particles for BON, Greedy Search, and RFJ Search methods on DPS, DAPS and MPGD $(B=8)$. The baseline algorithm corresponds to $N=1$.}
\label{tab:time_particles_daps_mpgd}
\resizebox{\textwidth}{!}{  % scales to fit full line width
\begin{tabular}{@{}c@{}}
\begin{subtable}[t]{0.5\textwidth}
\centering
\begin{tabular}{c|c|c|c}
\toprule
\textbf{Particles} & \textbf{BON} & \textbf{Greedy Search} & \textbf{RFJ Search} \\
\midrule
1      & 55  & -     & -     \\ %& 55"  & -     & -     \\
2        & 65 & 75 & 75 \\ %& 1:05" & 1:15" & 1:15" \\
4        & 102 & 118 & 131 \\ %& 1:42" & 1:58" & 2:11" \\
8        & 180 & 195 & 241 \\ %& 3:00" & 3:15" & 4:01" \\
\bottomrule
\end{tabular}
\caption{Task: Box inpainting. Baseline: DPS}
\label{tab:dps_box_time}
\end{subtable}
\hfill
\begin{subtable}[t]{0.45\textwidth}
\centering
\begin{tabular}{c|c|c|c}
\toprule
\textbf{Particles} & \textbf{BON} & \textbf{Greedy Search} & \textbf{RFJS} \\
\midrule
1      & 61  &   -  &  -  \\
2        & 72 & 91 & 91 \\
4        & 125 & 141 & 157\\
8        &  229 & 245 & 290 \\
\bottomrule
\end{tabular}
\caption{Task: Box inpainting. Baseline: DAPS}
\label{tab:daps_box_time}
\end{subtable}
\hfill
\begin{subtable}[t]{0.58\textwidth}
\centering
\begin{tabular}{c|c|c|c}
\toprule
\textbf{Particles} & \textbf{BON} & \textbf{Greedy Search} & \textbf{RFJS} \\
\midrule
1  & 3 & - & - \\ %& 1:16" & -     &   -    \\
2     & 4 & 5  & 5  \\ %& 1:20" & 2:03" & 2:03" \\
4     & 5 & 8 & 9 \\ %& 2:19" & 2:22" & 2:42" \\
8     & 8 & 12 & 23 \\ %& 3:50" & 4:02" & 4:50" \\
\bottomrule
\end{tabular}
\caption{Task: Box inpainting. Baseline:  MPGD}
\label{tab:mpgd_time}
\end{subtable}
\end{tabular}
}
\end{table}
}

\subsection{Ablation on Reward Function}
\label{app:reward_ablation}

\paragraph{Choice of reward function.}
A natural question is how to choose the reward function $r(x_0; s)$ and how sensitive the framework is to this choice. As discussed in Section~4.1, the reward function need only satisfy a simple and intuitive condition: it should assign high values to pairs $(x_0, s)$ that are consistent, and low values to inconsistent pairs. Any pretrained model that captures this notion of pairwise consistency is a valid candidate. In practice, such models are widely available for common side-information modalities, face recognition networks for image side information, and text-image alignment models for textual side information, and require no additional training.

To identify the best candidate for each modality, we conducted a brief literature review of pretrained models known to capture the relevant notion of consistency in prior recognition or generative modeling settings. For face side information, we evaluated AdaFace~\cite{kim2022adaface} and FaceNet, both trained to assign higher similarity to image pairs from the same identity. For textual side information, we evaluated ImageReward~\cite{xu2023imagereward} and CLIP score~\cite{radford2021clip}, where the former is trained to reflect human preference and text-image alignment, and the latter uses a contrastive objective for matched image-text pairs.

We note that during diffusion sampling, the reward is evaluated on intermediate denoised estimates $\hat{x}_{0|t}(x_t)$ (the Tweedie estimate) rather than directly on the noisy state $x_t$, since $\hat{x}_{0|t}$ lies closer to the image manifold and therefore falls more reliably within the distribution on which these networks were trained. Robustness to such intermediate inputs was an additional practical consideration in our selection.

Crucially, \emph{all reward candidates we tested improved over the baseline}, confirming that the framework is not sensitive to the specific choice of reward, as long as it satisfies the consistency condition above. We ultimately selected AdaFace and ImageReward as they yielded the strongest empirical performance. The full comparison is reported in Tables~\ref{tab:reward_ablation_face} and~\ref{tab:reward_ablation_text}.

\paragraph{Clarification on optimization target vs.\ evaluation metric.}
We wish to explicitly address a potential source of confusion. Our method optimizes the reward $r(\hat{x}_0; s)$, which measures consistency between the \emph{reconstructed image} $\hat{x}_0$ and the \emph{side information} $s$. However, the FS metric reported in our tables measures the distance between the \emph{reconstructed image} $\hat{x}_0$ and the \emph{ground truth} $x_0^*$.

These are \textbf{not the same quantity}. The ground truth $x_0^*$ is never observed at inference time and plays no role in the reward computation or the search procedure. The side information $s$ is a \emph{different} image of the same identity, taken under different conditions (pose, lighting, etc.), and is not equal to $x_0^*$. So the improvements we get are not a consequence of optimizing the same quantity that is being measured, and it is not reward hacking.

Below, we provide an ablation on choice of reward functions for face as side information and text as side information.

\begin{table}[H]
\centering
\scriptsize
\setlength{\tabcolsep}{3.5pt}
\begin{tabular}{llccccc}
\toprule
\textbf{Task} & \textbf{Param} & \textbf{Algo} & \textbf{FS}$\downarrow$ & \textbf{PSNR}$\uparrow$ & \textbf{LPIPS}$\downarrow$ & \textbf{SSIM}$\uparrow$ \\
\midrule
\multirow{3}{*}{Box Inpainting} & \multirow{3}{*}{$M{=}96$}
& DPS     & 0.739 & \underline{27.93} & \underline{0.139} & 0.852 \\
& & FaceNet & \underline{0.534} & 26.55 & 0.144 & \underline{0.852} \\
& & AdaFace & \textbf{0.308} & \textbf{28.29} & \textbf{0.136} & \textbf{0.855} \\
\midrule
\multirow{3}{*}{Super Resolution} & \multirow{3}{*}{$S{=}12$}
& DPS     & 1.23 & \underline{22.78} & 0.265 & \underline{0.613} \\
& & FaceNet & \underline{0.962} & 22.69 & \underline{0.265} & 0.608 \\
& & AdaFace & \textbf{0.603} & \textbf{22.92} & \textbf{0.262} & \textbf{0.619} \\
\midrule
\multirow{3}{*}{Motion Deblur} & \multirow{3}{*}{$K{=}256$}
& DPS     & 1.21 & \textbf{22.69} & 0.260 & \textbf{0.615} \\
& & FaceNet & \underline{0.945} & 21.88 & 0.277 & 0.587 \\
& & AdaFace & \textbf{0.545} & \underline{22.67} & \textbf{0.257} & \underline{0.614} \\
\bottomrule
\end{tabular}
\caption{Comparison of RFJS with different reward functions for image side information.
FS is evaluated against the \emph{ground truth} (not the side information used during inference).
Both FaceNet and AdaFace improve over the DPS baseline, confirming that any reward satisfying the consistency condition is beneficial; AdaFace achieves the best overall performance and is used in the main paper.}
\label{tab:reward_ablation_face}
\end{table}

\begin{table}[H]
\centering
\scriptsize
\setlength{\tabcolsep}{3.5pt}
\begin{tabular}{llccccc}
\toprule
\textbf{Task} & \textbf{Param} & \textbf{Algo} & \textbf{CS}$\uparrow$ & \textbf{PSNR}$\uparrow$ & \textbf{LPIPS}$\downarrow$ & \textbf{SSIM}$\uparrow$ \\
\midrule
\multirow{3}{*}{Box Inpainting} & \multirow{3}{*}{$M{=}138$}
& DPS         & 0.866 & 19.87 & 0.314 & 0.667 \\
& & CLIP        & \underline{0.875} & \underline{19.93} & \underline{0.305} & \underline{0.670} \\
& & ImageReward & \textbf{0.895} & \textbf{20.75} & \textbf{0.290} & \textbf{0.681} \\
\midrule
\multirow{3}{*}{Super Resolution} & \multirow{3}{*}{$S{=}32$}
& DPS         & 0.737 & \underline{16.83} & 0.521 & \underline{0.326} \\
& & CLIP        & \underline{0.784} & 16.37 & \underline{0.511} & 0.311 \\
& & ImageReward & \textbf{0.814} & \textbf{17.10} & \textbf{0.493} & \textbf{0.343} \\
\midrule
\multirow{3}{*}{Motion Deblur} & \multirow{3}{*}{$K{=}256$}
& DPS         & 0.783 & \underline{17.70} & 0.468 & \underline{0.364} \\
& & CLIP        & \underline{0.836} & 17.60 & \underline{0.465} & 0.353 \\
& & ImageReward & \textbf{0.866} & \textbf{18.27} & \textbf{0.439} & \textbf{0.367} \\
\bottomrule
\end{tabular}
\caption{Comparison of RFJS with different reward functions for textual side information.
CLIPScore is evaluated against the \emph{ground truth} image (not the text prompt used during inference).
Both CLIP-based and ImageReward-based rewards improve over the DPS baseline; ImageReward achieves the best performance and is used in the main paper.}
\label{tab:reward_ablation_text}
\end{table}

\subsection{Hyperparameter Sensitivity}
\label{app:hyperparameter}

Our framework introduces very few hyperparameters. The temperature $\tau$ controls the
softness of particle resampling, but as discussed in Section~4.2, RFJS naturally balances
exploration and exploitation through its hierarchical grouping structure, because of this, in practice, we set $\tau$ to a sufficiently small
value so that resampling becomes deterministic (always retaining the highest-reward particle
within each group), eliminating $\tau$ as a tuning parameter entirely.

The primary remaining hyperparameter is the resampling base $B$. We conduct an ablation
study over $B \in \{4, 8, 16, 32\}$ on the box inpainting task using DPS as the base sampler,
with $N=8$ particles. The results are reported in Table~\ref{tab:B_ablation}. Across all
values of $B$, RFJS substantially improves the perceptual FS metric over
the DPS baseline (from $0.739$ to a range of $0.296$--$0.434$), while keeping classical
metrics (PSNR, LPIPS, SSIM) in a consistently better range. Crucially, the variation across
choices of $B$ is small, confirming that the method is not sensitive to this hyperparameter.

In practice, we select $B$ once per base sampler, reflecting properties of the sampler
such as its number of diffusion steps and degree of inherent exploration, and apply the
same value uniformly across all tasks using that sampler. No per-task tuning of $B$ is performed.

\begin{table}[H]
\centering
\scriptsize
\setlength{\tabcolsep}{4pt}
\begin{tabular}{lcccc}
\toprule
\textbf{Method} & \textbf{PSNR}$\uparrow$ & \textbf{LPIPS}$\downarrow$ & \textbf{SSIM}$\uparrow$ & \textbf{FS}$\downarrow$ \\
\midrule
DPS (baseline)   & 27.929 & 0.1391 & 0.8519 & 0.739 \\
\midrule
RFJS ($B{=}4$)  & \underline{28.360} & 0.1365 & \textbf{0.8562} & \textbf{0.296} \\
RFJS ($B{=}8$)  & 28.186 & \textbf{0.1354} & 0.8551 & \underline{0.330} \\
RFJS ($B{=}16$) & \textbf{28.385} & \underline{0.1363} & \textbf{0.8562} & 0.372 \\
RFJS ($B{=}32$) & 28.264 & 0.1383 & \underline{0.8553} & 0.434 \\
\bottomrule
\end{tabular}
\caption{Ablation over resampling base $B$ for RFJS with DPS on box inpainting ($N{=}8$, $M{=}96$).
All values of $B$ substantially improve the perceptual FS metric over the DPS
baseline, and classical metrics remain in a consistently better range across all choices,
demonstrating low sensitivity to $B$.}
\label{tab:B_ablation}
\end{table}

\subsection{Evaluating Search Strategies in a 2D Setting}
\label{app:simulation}

To illustrate the effect of side information in inference-time search, we consider a simple 2D setup. The prior on $x_0$ is a mixture of Gaussians, shown in the leftmost image of Figure~\ref{fig:2d-mixture-of-gaussian}. A ground-truth sample $x_0$ is drawn from this prior, and a random forward operator $A$ is generated with fixed norm and random orientation. The measurement is then $y = A x_0 + n$. The corresponding posterior $p(x_0 \mid y)$, approximated by DPS, is also shown in Figure~\ref{fig:2d-mixture-of-gaussian}. Because this is an ill-posed inverse problem, the posterior is multimodal and DPS fails to reconstruct the correct solution.

We then add side information of the form $s = D x_0 + n_s$, where $D$ is chosen so that $D x_0$ is orthogonal to $A x_0$. The reward function is defined as $r(x_t, s) = - \lVert s - D \hat{x}_0(x_t) \rVert_2^2$, where $\hat{x}_0(x_t)$ is the Tweedie estimate of the clean signal from the noisy state $x_t$. Using this reward, RFJ Search produces posterior samples $p(x_0 \mid y, s)$ as shown in the three rightmost images of Figure~\ref{fig:2d-mixture-of-gaussian}. \rev{This is the Gaussian case of \cref{app:exactness}: because $\vs$ is generated as a noisy linear measurement of $x_0$ and the reward is the negative squared residual, the reward-tilted target of \cref{eq:proxy-posterior} coincides with $p(x_0 \mid y, s)$ at $\tau = 2\sigma_s^2$, so this setting is one of the few in which the particles can be compared against the distribution they are meant to represent.} As the resampling base $B$ decreases, the particles concentrate more tightly around the ground truth, highlighting how side information guides inference.

\begin{figure}[H]
    \centering
    \includegraphics[width=0.5\linewidth]{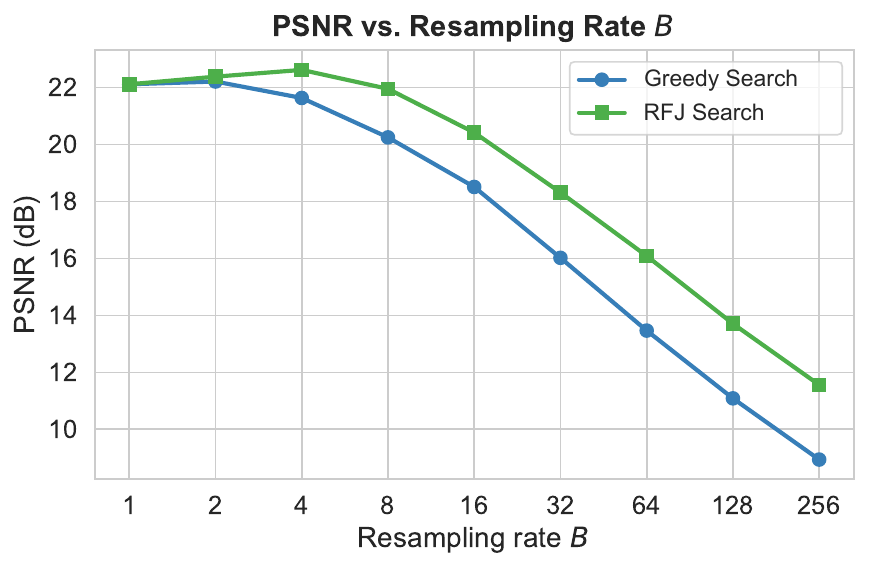}
    \caption{Comparison of performance of RFJ Search (RFJS) and Greedy Search (GS) as a function of $B$ for a randomly generated reward network $r_\theta$. RFJS outperforms GS across all values of $B$.}
    \label{fig:2d-rfjs-vs-gs}
\end{figure}

\begin{figure}[H]
    \centering
    \includegraphics[width=\linewidth]{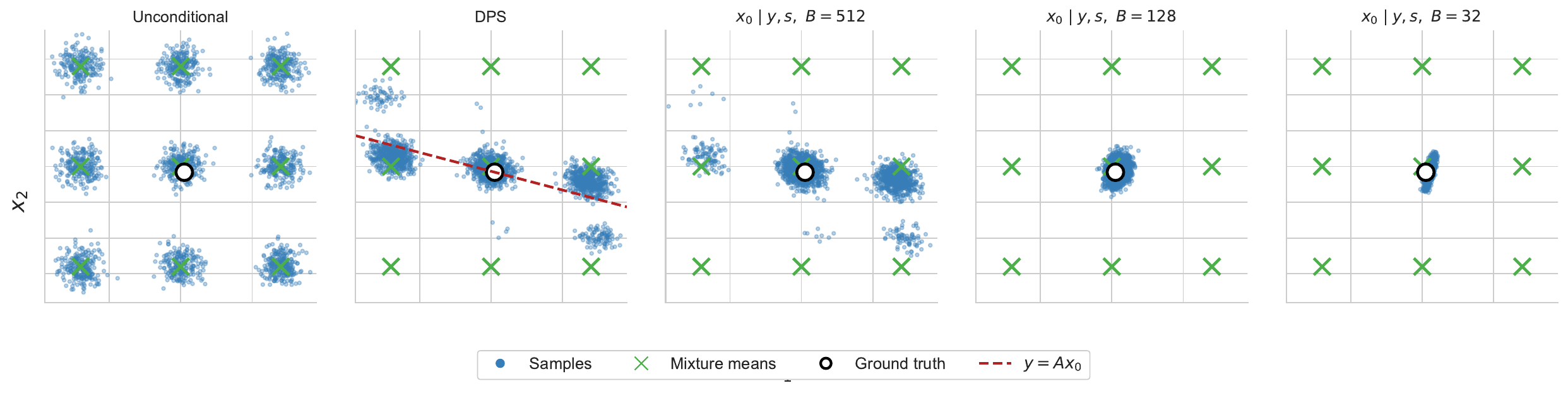}
    \caption{Illustration of the effect of $B$ in utilization of side information for the reconstruction in a linear inverse problem with a mixture of Gaussian prior.}
    \label{fig:2d-mixture-of-gaussian}
\end{figure}

To compare RFJ and Greedy Search, we next consider a more realistic scenario where side information is generated through a neural network reward model. For example, in face reconstruction tasks, side information may come from the embedding of an additional image, while for text-conditioned tasks, it may come from a text encoder. To mimic this, in each trial we sample a ground truth $x_0$, a forward operator $A$, and a reward network $r_\theta$ with randomly initialized weights. The side information is defined as $s = r_\theta(x_0) + n_s$, and the reward is the cosine similarity between $r_\theta(\hat{x}_0(x_t))$ and $s$. We repeat this experiment 16,000 times, and for each trial run both RFJ and Greedy Search 128 times with $N=8$ particles, varying the resampling base $B$ over powers of two from $1$ to $256$. The average PSNR values are reported in Figure~\ref{fig:2d-rfjs-vs-gs}. Across all settings, RFJ Search consistently outperforms Greedy Search, with the best performance observed at $B=4$.

\subsection{\rev{Alternative Resampling and SMC Strategies}}
\label{app:resampling_strategies}

\rev{\Cref{app: additional_related_work} places our algorithms on the two axes that specify an SMC method, weighting and resampling. The strategies compared here each vary one of them. \emph{Adaptive SMC} replaces the fixed resampling clock with a data-driven trigger, resampling whenever the effective sample size $1/\sum_i w_i^2$ falls below a threshold. \emph{Annealed SMC} tempers the weights over time through $\exp(\lambda_t r / \tau)$ with $\lambda_t$ increasing from $0$ to $1$, so that selection pressure is weak early in the trajectory and sharpens as the estimate of $\vx_0$ becomes reliable. \emph{Stratified resampling} reduces the variance of the survivor draw by partitioning $[0,1]$ into equal bins and drawing one index per bin; \emph{multinomial resampling} is the textbook alternative, included as its reference point. \emph{Adaptive group sizes} retains the structure of RFJS but sets the group size from a run-time statistic rather than a fixed schedule. \emph{Reward-guided beam search} maintains $M$ beams, expands each into $K$ children and deterministically keeps the top $M$ by reward.}

\paragraph{\rev{Setting.}}\rev{ All strategies solve the same box-inpainting task with a $96 \times 96$ mask and DPS as the base solver, use $N=8$ particles and the same AdaFace reward, and are evaluated on the same $16$ images; they differ only in the selection rule. The resampling cadence is $B=16$ for RFJS, adaptive group sizes and beam search, and $B=4$, which resamples the smallest group at every reverse step, for the SMC strategies, which is their textbook setting. Adaptive SMC uses systematic resampling whenever the effective sample size falls below $N/2$, with a temperature given by a robust (median-absolute-deviation) estimate of the spread of the reward increments; we swept this temperature over $1$, $2$, $4$ and $16\times$ and report $2\times$. Annealed SMC uses the same weights with a linear ramp $\lambda_t = t/T$ and resamples at every step. The stratified and multinomial rows both resample globally from weights proportional to $\exp(r/\tau)$ with the same robust $\tau$, and differ only in how the survivors are drawn. Adaptive group sizes uses the $B=16$ and the group range $2$--$8$ of RFJS, but sets $g_t$ to the nearest power of two to the run-time effective sample size. Beam search keeps $M=2$ beams of width $K=4$ and prunes deterministically at the same cadence.}

\begin{table}[H]
\centering
\scriptsize
\setlength{\tabcolsep}{6pt}
\rev{%
\begin{tabular}{lcccc}
\toprule
\textbf{Search algorithm} & \textbf{PSNR}$\uparrow$ & \textbf{LPIPS}$\downarrow$ & \textbf{SSIM}$\uparrow$ & \textbf{FS}$\downarrow$ \\
\midrule
RFJS (ours)               & \underline{28.23} & \textbf{0.128} & \textbf{0.877} & \textbf{0.196} \\
Adaptive SMC              & 28.01 & 0.142 & 0.857 & 0.254 \\
Annealed SMC              & 27.54 & 0.143 & 0.853 & 0.218 \\
Stratified resampling     & 27.83 & 0.143 & 0.850 & 0.209 \\
Multinomial resampling    & 27.62 & 0.156 & 0.842 & 0.213 \\
Adaptive group sizes      & 28.02 & \underline{0.139} & \underline{0.864} & 0.211 \\
Reward-guided beam search & \textbf{28.36} & \underline{0.139} & 0.863 & \underline{0.206} \\
\bottomrule
\end{tabular}}
\caption{\rev{Alternative weighting and resampling rules on box inpainting with a $96 \times 96$ mask, DPS as the base solver and $N=8$ particles, evaluated on $16$ images. Every strategy uses the side information through the same AdaFace reward and differs only in the selection rule (\textbf{best}, \underline{second best}). RFJS attains the best LPIPS, SSIM and FS, and the second-best PSNR.}}
\label{tab:smc_comparison}
\end{table}

\begin{table}[H]
\centering
\scriptsize
\setlength{\tabcolsep}{6pt}
\rev{%
\begin{tabular}{lcccc}
\toprule
\textbf{Method} & \textbf{PSNR}$\uparrow$ & \textbf{LPIPS}$\downarrow$ & \textbf{SSIM}$\uparrow$ & \textbf{FS}$\downarrow$ \\
\midrule
DPS (no resampling)                  & 25.72 & 0.1403 & 0.8437 & 1.2250 \\
Stochastic resampling, $\tau = 1/3$  & 25.97 & \textbf{0.1368} & 0.8456 & 0.5142 \\
Stochastic resampling, $\tau = 1/10$ & \textbf{26.67} & 0.1372 & \textbf{0.8467} & \textbf{0.2996} \\
\bottomrule
\end{tabular}}
\caption{\rev{Stochastic resampling at a positive temperature, for box inpainting with a $116 \times 116$ mask, DPS as the base solver, $N=8$ particles and $B=16$, evaluated on $8$ images. Positive-temperature resampling is a direct implementation of the reward-tilted formulation of \cref{prop:seq-diff-sampling}. Both temperatures improve FS substantially over the base sampler while also improving PSNR, SSIM and LPIPS, and identity preservation improves as $\tau$ decreases; the deterministic selection used throughout our experiments is the endpoint of that trend.}}
\label{tab:stochastic_resampling}
\end{table}

\subsection{\rev{Additional Baselines}}
\label{app:additional_baselines}

\paragraph{\rev{An SMC base solver: MCGdiff.}}\rev{ Because our framework acts on top of a base sampler, it can also be applied to a sampler that already maintains particles. We augment MCGdiff \citep{cardoso2024monte} by replacing its resampling rule every fourth step with resampling based on our side-information reward, and evaluate on $32\times$ super-resolution. \Cref{tab:mcgdiff_baseline} reports the result. Adding RFJS improves MCGdiff on all four metrics. The gain in FS is smaller than the corresponding gain for DPS, which we attribute to the fact that the particles of MCGdiff do not evolve independently between side-information resampling steps, so there is less trajectory diversity for the search to exploit. The table also contains the direct comparison between MCGdiff and DPS with RFJS.}

\begin{table}[H]
\centering
\scriptsize
\setlength{\tabcolsep}{6pt}
\rev{%
\begin{tabular}{lcccc}
\toprule
\textbf{Method} & \textbf{PSNR}$\uparrow$ & \textbf{LPIPS}$\downarrow$ & \textbf{SSIM}$\uparrow$ & \textbf{FS}$\downarrow$ \\
\midrule
MCGdiff ($N=64$)        & 17.980 & 0.3825 & 0.4868 & 1.3970 \\
MCGdiff + RFJS ($N=64$) & 18.207 & 0.3709 & \underline{0.4938} & \underline{0.9821} \\
DPS ($N=8$)             & 18.025 & \underline{0.3667} & 0.4831 & 1.3997 \\
DPS + RFJS ($N=8$)      & \textbf{18.426} & \textbf{0.3574} & \textbf{0.4853} & \textbf{0.3931} \\
\bottomrule
\end{tabular}}
\caption{\rev{RFJS applied on top of MCGdiff, an SMC solver, for $32\times$ super-resolution with face side information. The side-information reward replaces the resampling rule of MCGdiff every fourth step. Each method is run at its own particle budget, $N=64$ for MCGdiff and $N=8$ for DPS, and all four rows are evaluated on the same $16$ images.}}
\label{tab:mcgdiff_baseline}
\end{table}

\paragraph{\rev{A text-conditioned base solver: TReg.}}\rev{ \Cref{app:stable_diffusion} already evaluates our framework with Stable Diffusion as the base model. Here we instead use TReg \citep{kim2025regularization} as the base solver and apply our search on top of it, for the box-inpainting task with textual side information. Selection is Best-of-$N$ over $N=8$ particles, that is, the $B \geq T$ endpoint of Greedy Search, in which the particles evolve independently and the reward is applied once at the end of the trajectory. \Cref{tab:treg_baseline} shows that the search improves every reported metric relative to TReg alone.}

\begin{table}[H]
\centering
\scriptsize
\setlength{\tabcolsep}{6pt}
\rev{%
\begin{tabular}{lcccc}
\toprule
\textbf{Method} & \textbf{PSNR}$\uparrow$ & \textbf{LPIPS}$\downarrow$ & \textbf{SSIM}$\uparrow$ & \textbf{CS}$\uparrow$ \\
\midrule
TReg        & 15.25 & 0.393 & 0.542 & 0.814 \\
TReg + ours & \textbf{15.81} & \textbf{0.381} & \textbf{0.550} & \textbf{0.838} \\
\bottomrule
\end{tabular}}
\caption{\rev{Our search applied on top of TReg, a text-conditioned solver, for box inpainting with a $138 \times 138$ mask and textual side information. Selection is Best-of-$N$ over $N=8$ particles, evaluated on $24$ ImageNet images.}}
\label{tab:treg_baseline}
\end{table}

\paragraph{\rev{A trained reference-conditioned model: ReF-LDM.}}\rev{ ReF-LDM \citep{hsiao2024refldm} is a latent diffusion model trained for reference-based face restoration, and therefore provides a point of comparison in which the side information is used by a trained conditional model rather than at inference. We compare it against DPS with RFJS and $N=8$ particles on super-resolution at a mild and a severe degradation level, and on motion deblurring. Both methods receive the same degraded measurement and the same single reference image. \Cref{tab:refldm_baseline} reports the results. At the degradations for which ReF-LDM was trained it is better on LPIPS and SSIM, while our training-free search is better on PSNR and on FS. On motion deblurring, a degradation for which ReF-LDM was not trained, RFJS is better on every metric and by a large margin.}

\rev{These numbers should not be read as isolating the mechanism by which side information is used, because the two approaches differ in several components at once. ReF-LDM uses a stronger latent diffusion model with a learned VQ decoder, whereas our method uses an off-the-shelf pixel-space FFHQ prior, and the sharper outputs encouraged by the learned decoder may account for part of its advantage on LPIPS and SSIM. Our method incorporates the measurement through an inverse-problem solver at inference time, whereas ReF-LDM is trained on particular degradation processes, which is what the motion-deblurring row reflects. Finally, ReF-LDM learns to incorporate the reference during training, whereas our framework introduces it only at inference through a reward.}

\begin{table}[H]
\centering
\scriptsize
\setlength{\tabcolsep}{6pt}
\rev{%
\begin{tabular}{llcccc}
\toprule
\textbf{Task} & \textbf{Method} & \textbf{PSNR}$\uparrow$ & \textbf{LPIPS}$\downarrow$ & \textbf{SSIM}$\uparrow$ & \textbf{FS}$\downarrow$ \\
\midrule
\multirow{2}{*}{Super Resolution, $S{=}4$}
 & ReF-LDM (trained)            & 24.89 & \textbf{0.183} & \textbf{0.755} & 0.253 \\
 & RFJS (ours, training-free)   & \textbf{25.04} & 0.218 & 0.702 & \textbf{0.241} \\
\midrule
\multirow{2}{*}{Super Resolution, $S{=}12$}
 & ReF-LDM (trained)            & 19.94 & \textbf{0.267} & \textbf{0.592} & 0.445 \\
 & RFJS (ours, training-free)   & \textbf{20.74} & 0.294 & 0.566 & \textbf{0.327} \\
\midrule
\multirow{2}{*}{Motion Deblur}
 & ReF-LDM (trained)            & 22.25 & 0.204 & 0.688 & 0.292 \\
 & RFJS (ours, training-free)   & \textbf{26.48} & \textbf{0.188} & \textbf{0.752} & \textbf{0.208} \\
\bottomrule
\end{tabular}}
\caption{\rev{Comparison against ReF-LDM, a model trained for reference-based face restoration. Both methods receive the same measurement and the same single reference image. Motion deblurring is a degradation outside the training distribution of ReF-LDM. The RFJS rows use $N=8$ particles and $B=16$, and both methods are evaluated on the same $8$ images.}}
\label{tab:refldm_baseline}
\end{table}

\subsection{\rev{Reward Calibration and Ranking Fidelity}}
\label{app:reward_quality}

\rev{\Cref{app:exactness} records the condition under which the reward-tilted target of \cref{eq:proxy-posterior} coincides with the posterior: the reward must be an affine function of the log-likelihood, $r(\vx_0; \vs)/\tau = \log p_{S \mid 0}(\vs \mid \vx_0) + c(\vs)$ with $c$ free of $\vx_0$. This subsection examines how closely the reward we actually use satisfies that condition, and, since the main experiments operate in the $\tau \to 0$ limit in which only the ordering of the reward matters, how reliably that ordering agrees with an oracle that has access to the ground truth.}

\paragraph{\rev{Calibration.}}\rev{ Of the three rewards we use, AdaFace is the one whose training
objective leaves magnitudes free (\cref{app:exactness}), so it is the one worth measuring.
\Cref{eq:adaface-exact} is what makes the measurement possible. It writes the log-likelihood ratio
that \cref{eq:exactness} is a statement about, $\log \big( p_{S \mid 0}(\vs \mid \vx_0) / p_S(\vs)
\big)$, in closed form as $\log M + \log \langle \pi(\vx_0), \pi(\vs) \rangle$, so the quantity the
reward is meant to be affine in can be evaluated for any pair from the two embeddings alone, with
no density to estimate. We evaluate it at $\kappa = 8$ on the pairs the search actually forms,
which are pairs of a reconstruction with a photograph of the same person: $\vx_0$ ranges over the
$512$ reconstructions of the $\times 16$ super-resolution run of
\cref{tab:quantitative_results_dps} and each is paired with both photographs of its own identity,
giving $1{,}024$ pairs. On the same pairs we evaluate the reward we use,
$r(\vx_0; \vs) = -\lVert \Phi(\vx_0) - \Phi(\vs) \rVert_2$, and fit a straight line of the first
quantity on the second by least squares. \Cref{eq:exactness} is the statement that this line fits,
with slope $1/\tau$, and we measure how well it does by $R^2$: one minus the residual sum of
squares divided by the total sum of squares of the log-likelihood ratio about its mean, over the
same $1{,}024$ pairs. Two features of that figure fix how it should be read. It is a fit to the
individual pairs, with nothing averaged beforehand; grouping the pairs into bins by $r$ and fitting
the bin means instead discards the scatter within each bin and reports a markedly higher value on
the same data. And its denominator is the spread of the log-likelihood ratio over these pairs
alone, so it describes how well the line predicts a single pair drawn from the regime the search
operates in, rather than how well the reward separates photographs of different people, which lie
further apart and are correspondingly easier to tell apart. The line accounts for $80\%$ of that
variation, a correlation of $0.90$, at a slope corresponding to $\tau = 0.46$. The embedding
distance is therefore an approximately affine image of the exact quantity \cref{eq:adaface-exact}
supplies, informative about it rather than equivalent to it, which is what the $\tau \to 0$ limit
asks of it: the reward enters only through the ordering it induces on the particles of one image.}

\paragraph{\rev{Ranking fidelity.}}\rev{ In the $\tau \to 0$ limit the algorithm reads nothing from the reward but the ordering it induces: at a resampling step it holds $N = 8$ candidate reconstructions of the same image and keeps whichever the reward scores highest, so the only thing that matters is whether the reward orders those eight as an observer with access to the ground truth would. We measure that agreement directly. For each image we rank the eight particles by the reward $r(\vx_0; \vs)$, which sees only the side information, and rank them again by an oracle metric computed against the ground truth. Kendall's $\tau$ is the fraction of the $\binom{8}{2} = 28$ particle pairs that the two rankings place in the same relative order, minus the fraction they place in opposite orders: $\tau = 1$ means the two rankings are identical, $\tau = 0$ that they agree no more often than chance, and $\tau = -1$ that one is the reverse of the other. \Cref{tab:kendall} reports this averaged over the $64$ evaluation images, for four choices of oracle, with DPS as the base solver.}

\rev{The FS column is a control rather than a result: that oracle is the same embedding distance as the reward, measured against the ground truth instead of against $\vs$, and since the two images are photographs of one person the rankings ought to agree. They do, at $0.76$ to $0.88$, decreasing as the inverse problem becomes more ill-posed and more reconstructions become consistent with the measurement. The other three columns carry the content. Against the distortion oracles the agreement is near zero, so at the moment of selection the reward is almost uninformative about which particle will score better on PSNR, LPIPS or SSIM. This is the particle-level counterpart of the effect documented in \cref{app:importance_of_target_metric}: a search driven by a perceptual reward moves the perceptual metric decisively and leaves the distortion metrics in roughly the same range, because at the point where it chooses it cannot tell them apart.}

\begin{table}[H]
\centering
\scriptsize
\setlength{\tabcolsep}{6pt}
\rev{%
\begin{tabular}{lcccc}
\toprule
 & \multicolumn{4}{c}{\textbf{Kendall's} $\tau$ \textbf{against oracle}} \\
\cmidrule(lr){2-5}
\textbf{Task} & \textbf{FS} & \textbf{LPIPS} & \textbf{PSNR} & \textbf{SSIM} \\
\midrule
Box Inpainting, $M{=}96$      & 0.876 & 0.163 & 0.171 & 0.177 \\
Box Inpainting, $M{=}116$     & 0.874 & 0.076 & $-0.011$ & 0.122 \\
Box Inpainting, $M{=}135$     & 0.874 & 0.137 & 0.073 & 0.118 \\
Super Resolution, $S{=}4$     & 0.840 & $-0.022$ & 0.051 & 0.037 \\
Super Resolution, $S{=}12$    & 0.801 & 0.099 & 0.047 & 0.054 \\
Super Resolution, $S{=}16$    & 0.767 & 0.050 & 0.073 & 0.084 \\
Super Resolution, $S{=}32$    & 0.762 & 0.037 & 0.002 & 0.000 \\
Gaussian Deblur               & 0.857 & 0.041 & $-0.011$ & 0.022 \\
Motion Deblur, $K{=}61$       & 0.853 & 0.067 & 0.023 & 0.081 \\
Motion Deblur, $K{=}256$      & 0.830 & 0.040 & 0.102 & 0.086 \\
Nonlinear Deblur              & 0.848 & 0.098 & 0.048 & 0.084 \\
Blind Deblur                  & 0.798 & 0.182 & 0.176 & 0.198 \\
\bottomrule
\end{tabular}}
\caption{\rev{Agreement between the ranking of the $N=8$ particles induced by the AdaFace reward and the ranking induced by an oracle with access to the ground truth, averaged over the $64$ evaluation images, with DPS as the base solver. Kendall's $\tau$ is $1$ for identical rankings and $0$ for chance agreement. FS is a control, being the reward's own distance measured against the ground truth; the remaining columns show that the reward is nearly uninformative about the distortion metrics.}}
\label{tab:kendall}
\end{table}

\section{Hyperparameters}
\label{app:exp_settings}

In our implementation of \textsc{Resample} in Algorithm~\ref{alg:inference_search}, we perform a greedy resampling, i.e., we pick the best candidate within each group of size $g_t$ and replicate it $g_t$ times. Since we are using large enough $B$, this is justified and has similar effect as using smaller $B$ with moderate temperature, with the added advantage of utilizing less function calls to the reward network $r$. Thus, tuning the $B$ allows us to maintain balance without over-optimizing with respect to the reward.

For reproducibility, we provide detailed settings for each task, sampler, and search algorithm used in our experiments below.

\subsection{Face Identity Experiments}
For experiments with face identity as side information:
\begin{itemize}
    \item \textbf{DPS:} Box inpainting with box size $96$. All other parameters (downsampling rate, blur kernel, noise levels) are the same as the default DPS settings in the implementation. Search algorithms use $N=8$ particles and resampling base $B=16$ for both RFJ and Greedy search. Gradient guidance is applied with scale $0.5$.
    \item \textbf{DAPS:} Box inpainting with box size $96$, super-resolution with downsampling rate $10\times$. Noise levels unchanged from original DAPS defaults. Search algorithms use $N=8$ particles and resampling base $B=4$, with gradient guidance scale $13$.
    \item \textbf{MPGD:} Box inpainting with box size $64$, super-resolution with downsampling rate $6\times$, and Gaussian deblur with intensity of $5.0$. Search algorithms use $N=8$ particles and resampling base $B=8$. Gradient guidance scale is $0.5$ for box inpainting and $0.25$ for super-resolution and Gaussian deblur.
\end{itemize}

\subsection{Text Side Information Experiments}
\label{app: text_hyperparameters}
When text descriptions were used as side information, we made the degradation more severe so that the information in $\vs$ was not already present in the measurement $\vy$. Otherwise, side information would not provide meaningful guidance. For example, if the input image is sharp enough to identify the type of animal, then explicitly stating it in $\vs$ adds little value.

The settings for these tasks are:
\begin{itemize}
    \item \textbf{Box inpainting:} Box size $138$, measurement noise $\sigma=0.05$ (the default).
    \item \textbf{Super-resolution:} Downsampling rate $32\times$, measurement noise $\sigma=0.05$ (the default).
    \item \textbf{Motion/Gaussian deblur:} Kernel size $256$, \rev{intensity $12.0$ for the Gaussian kernel and $0.5$ for the motion kernel}, measurement noise $\sigma=0.1$.
    \item \textbf{Nonlinear deblur:} Kernel size unchanged, measurement noise $\sigma=0.5$.
    \item \rev{\textbf{Blind deblur:} Kernel size $64$ with intensity $0.5$, measurement noise $\sigma=0.02$. This task is run with BlindDPS, which estimates the kernel jointly with the image, and its absolute scores are therefore not comparable with the five DPS tasks above.}
\end{itemize}

For all tasks in this setting, search algorithms use $N=4$ particles and resampling base $B=100$.

\rev{Three further settings are shared across these tasks. Resampling is enabled only after the
first $400$ of the $1000$ diffusion steps, so that the particles diverge before any trajectory
is duplicated. The reward-gradient scale is zero throughout, so the text results isolate the
contribution of search and contain no reward gradient guidance. The DPS measurement-gradient
step size is $0.5$ for box inpainting, $0.9$ for super-resolution and $0.3$ for each of the
four deblurring tasks.}

These hyperparameters ensure that our framework is evaluated under severe degradations (heavy downsampling, blurring, or noise), while search and guidance settings remain consistent across samplers and modalities.

\subsection{MRI Experiments}
\label{app: mri_experiments}
We used the contrast-pairings among the files in the fastMRI dataset, provided by \cite{atalik2025mriwithsideinfo}. We collect the data from the (fastMRI) source and preprocess to be compatible with the inputs in ContextMRI. Specifically, the setup used in the data is multi-coil MRI acquisition, which requires us to estimate the coil sensitivity maps, and then a complex reconstruction from them. ContextMRI takes complex values as inputs and denoises to produce a complex-valued 2D image. We computed NMI with $64$ bins at each step of the diffusion process to balance complexity with performance. We use the defaults parameters as in ContextMRI, except for the acceleration factor, $16$ and the center fraction (ACS), $0.02$. We use a pair of anatomy which two contrasts, which has more 30 slices. We consider the slices $15$ to $28$ as these are more challenging and report the results by using one as the side information for the other.

\rev{The closest method to ours in this setting is TGVN \citep{atalik2025mriwithsideinfo}, which constrains the ambiguous subspaces of the solution using a second contrast through a learned unrolled network, and from which we take the contrast pairings above. We report no quantitative comparison against it: the released repository provides training code and data splits but no trained weights, so reproducing it means training from scratch, and the authors' configuration for a single experiment requests four A100 GPUs for up to fourteen days on fastMRI data that requires credentialed access.}

\begin{table}[t]
 \caption{Quantitative MRI reconstruction results (fastMRI knee, AF=16, ACS=2\%).}
    \label{tab:mri-quantitative-results}
    \centering
    \resizebox{0.5\linewidth}{!}{
    \begin{tabular}{l c c c c}
        \toprule
        & \multicolumn{4}{c}{~PDFS with PD} \\
        \midrule
        Algorithm & PSNR ($\uparrow$) & SSIM ($\uparrow$) & LPIPS ($\downarrow$) & NMI ($\uparrow$) \\
        \midrule
        RFJS & \textbf{25.85} & \textbf{0.801} & \textbf{0.375} & \textbf{0.457} \\
        GS & 25.33 & \underline{0.797} & \underline{0.375} & \underline{0.455} \\
        BON & \underline{25.47} & 0.797 & 0.376 & 0.454 \\
        ContextMRI & 25.39 & 0.795 & 0.383 & 0.451 \\
        \midrule
        & \multicolumn{4}{c}{~PD with PDFS} \\
        \midrule
        RFJS & \textbf{27.85} & \textbf{0.920} & \textbf{0.358} & \textbf{0.579} \\
        GS & \underline{27.80} & \underline{0.920} & \underline{0.360} & \underline{0.579} \\
        BON & 27.80 & 0.918 & 0.366 & 0.570 \\
        ContextMRI & 27.46 & 0.915 & 0.375 & 0.563 \\
        \bottomrule
    \end{tabular}%
    }
\end{table}

\section{Limitations}
\label{app:limitations}

Our proposed search algorithms lack formal optimality guarantees for exploration-exploitation, and we do not claim they are theoretically optimal. We expect that stronger algorithms are possible, potentially improving both sample efficiency and robustness. A central reason is the absence of a general mathematical framework for designing optimal exploration–exploitation strategies in diffusion-based inverse problems with side information, an open problem we highlight. Practically, this means our methods rely on principled heuristics and tuned schedules (e.g., reward scaling, resampling/branching rates) whose compute allocation is not provably optimal, suggesting a clear direction for future work.

\paragraph{\rev{Misleading side information.}}\rev{Our formulation assumes that $\vs$ comes from an external source that carries genuine information about $\vx_0$, such as a second photograph of the same person, a complementary MRI contrast, or a description written by someone who saw the scene. When that assumption fails, the method inherits the failure. The mechanism is a wrong reward ranking: in a severely ill-posed problem several semantically distinct reconstructions explain the same measurement, and a reward computed against misleading side information will rank the wrong one highest. We make no claim to eliminate hallucination. What the structure of the algorithm does bound is the extent of the failure, since the search only selects among candidates produced by the measurement-conditioned kernel of the base solver and never overrides $\vy$; \cref{tab:wrong_side_info} measures this, with identity fidelity degrading sharply while the distortion metrics stay close to their correct-reference values. The stakes are higher in safety-critical settings. In medical imaging in particular, a confidently wrong auxiliary contrast would produce a reconstruction that is consistent with the acquired data and still clinically misleading, which is the reason for treating the output as decision support rather than as a measurement.}

\paragraph{\rev{Compute for quality.}}\rev{Maintaining $N$ particles costs compute, and the quality we report is bought with it. The analytical cost is given in \cref{app:runtimes} and grows linearly in $N$, but the wall-clock cost grows considerably more slowly because the particles are processed in parallel: on box inpainting with DPS, one particle takes $55$ seconds per image and $N=2$, $4$ and $8$ take $75$, $131$ and $241$ seconds, so eight particles cost about $4.4\times$ the baseline rather than $8\times$. The trade-off is a knob rather than a fixed price. Setting $N=1$ recovers the base solver exactly, and the gains do not require the largest budget: $N=2$ already removes roughly $40\%$ of the FS error on box inpainting, from $0.74$ to $0.44$ (\cref{app:num_particles_effect}). Still, for applications where inference time is tightly constrained, the compute overhead is a genuine limitation of inference-time search as a family of methods.}

\paragraph{Broader impacts.} Our method can improve reconstruction quality in scientific imaging applications such as MRI, with positive impact on diagnostic imaging. Potential negative impacts include increased fidelity of identity-conditioned reconstructions, which raises privacy and consent considerations. In safety-critical domains such as medical imaging, our method should be treated as decision support only, with domain-expert review.

\section{Use of Large Language Models}
\label{app:llm}

Parts of this work were assisted by large language models (specifically GPT-5 from OpenAI). Their use was limited to improving clarity, grammar, and the presentation of experimental descriptions. All conceptual contributions, experimental design, analysis, and final decisions are solely the authors’ responsibility. The models were not used to generate new research ideas, design experiments, or make unverifiable scientific claims. Additionally, large language models were used to generate textual descriptions serving as side information for a specific set of experiments, as detailed in the main paper.

\end{document}